\def\isarxiv{1} 

\ifdefined\isarxiv
\documentclass[11pt]{article}

\usepackage[numbers]{natbib}
\usepackage{booktabs}

\else
\documentclass{article} 
\usepackage{iclr2025_conference,times}
\iclrfinalcopy

\usepackage{hyperref}
\usepackage{url}
\usepackage{booktabs}

\title{Efficient Alternating Minimization with Applications to Weighted Low Rank Approximation}

\author{Zhao Song \\
Simons Institute for the Theory of Computing, UC Berkeley \\
\texttt{magic.linuxkde@gmail.com} 
\And
Mingquan Ye \\
University of Illinois Chicago \\
\texttt{mye9@uic.edu} 
\And
Junze Yin \\
Rice University \\
\texttt{jy158@rice.edu}
\And
Lichen Zhang \\
MIT CSAIL \\
\texttt{lichenz@csail.mit.edu}
}

\fi

\usepackage{amsmath}
\usepackage{amsthm}
\usepackage{amssymb}
\usepackage{algorithm}
\usepackage{algpseudocode}
\usepackage{grffile}
\usepackage{wrapfig,epsfig}
\usepackage{url}
\usepackage{xcolor}
\usepackage{epstopdf}
\usepackage{accents}
\usepackage{enumerate}
\usepackage{enumitem}

\usepackage{bbm}
\usepackage{dsfont}

\allowdisplaybreaks

\ifdefined\isarxiv

\usepackage{tikz}
\usepackage{hyperref}  
\hypersetup{colorlinks=true,citecolor=blue,linkcolor=blue} 
\usetikzlibrary{arrows}
\usepackage[margin=1in]{geometry}

\else

\usepackage{microtype}
\usepackage{hyperref}
\definecolor{mydarkblue}{rgb}{0,0.08,0.45}
\hypersetup{colorlinks=true, citecolor=mydarkblue,linkcolor=mydarkblue}

\fi
\theoremstyle{plain}
\newtheorem{theorem}{Theorem}[section]
\newtheorem{lemma}[theorem]{Lemma}

\newtheorem{claim}[theorem]{Claim}

\newtheorem{problem}[theorem]{Problem}

\theoremstyle{definition}
\newtheorem{definition}[theorem]{Definition}
\newtheorem{assumption}[theorem]{Assumption}

\newtheorem{fact}[theorem]{Fact}
\newtheorem{remark}[theorem]{Remark}

\newcommand{\wh}{\widehat}
\newcommand{\wt}{\widetilde}
\newcommand{\ov}{\overline}
\newcommand{\N}{\mathcal{N}}
\newcommand{\R}{\mathbb{R}}

\renewcommand{\hat}{\wh}

\newcommand{\eps}{\epsilon}
\newcommand{\diag}{\mathrm{diag}}

\DeclareMathOperator*{\E}{{\mathbb{E}}}

\DeclareMathOperator{\OPT}{OPT}

\DeclareMathOperator{\poly}{poly}

\DeclareMathOperator{\nnz}{nnz}

\DeclareMathOperator{\rank}{rank}
\DeclareMathOperator{\dist}{dist}

\makeatletter
\newcommand*{\RN}[1]{\expandafter\@slowromancap\romannumeral #1@}
\makeatother

\usepackage{lineno}

\begin{document}

\ifdefined\isarxiv

\date{}

\title{Efficient Alternating Minimization with Applications to Weighted Low Rank Approximation}
\author{
Zhao Song\thanks{\texttt{magic.linuxkde@gmail.com}. Simons Institute for the Theory of Computing, UC Berkeley.}
\and 
Mingquan Ye\thanks{\texttt{mye9@uic.edu}. University of Illinois, Chicago.}
\and 
Junze Yin\thanks{\texttt{jy158@rice.edu}. Rice University.}
\and
Lichen Zhang\thanks{\texttt{lichenz@csail.mit.edu}. Massachusetts Institute of Technology.}
}

\else

\fi


\ifdefined\isarxiv
\begin{titlepage}
  \maketitle
  \begin{abstract}
Weighted low rank approximation is a fundamental problem in numerical linear algebra, and it has many applications in machine learning. Given a matrix $M \in \mathbb{R}^{n \times n}$, a non-negative weight matrix $W \in \mathbb{R}_{\geq 0}^{n \times n}$, a parameter $k$, the goal is to output two matrices $X,Y\in \mathbb{R}^{n \times k}$ such that $\| W \circ (M - X Y^\top) \|_F$ is minimized, where $\circ$ denotes the Hadamard product. It naturally generalizes the well-studied low rank matrix completion problem. Such a problem is known to be NP-hard and even hard to approximate assuming the Exponential Time Hypothesis~\citep{gg11,rsw16}. Meanwhile, alternating minimization is a good heuristic solution for weighted low rank approximation. In particular,~\cite{llr16} shows that, under mild assumptions, alternating minimization does provide provable guarantees. In this work, we develop an efficient and robust framework for alternating minimization that allows the alternating updates to be computed approximately. For weighted low rank approximation, this improves the cost of each iteration of~\cite{llr16} from $\|W\|_0k^2$ to $\|W\|_0 k$ where $\|W\|_0$ denotes the number of nonzero entries of the weight matrix. At the heart of our framework is a high-accuracy multiple response regression solver together with a robust analysis of alternating minimization. We also replace the random initialization of~\cite{llr16}, whose claimed guarantee fails, by subspace iteration from a Gaussian start, so our total running time of $\wt O((\|W\|_0 k+nk^3)\log(1/\epsilon))$ includes the initialization. In exchange, our guarantee needs the weight matrix to be within spectral distance $\gamma n$ of the all-ones matrix for a $\gamma$ that is polynomially small in $n$.

\end{abstract}
  \thispagestyle{empty}
\end{titlepage}

{\hypersetup{linkcolor=black}
\tableofcontents
}
\newpage

\else
\maketitle
\begin{abstract}
Weighted low rank approximation is a fundamental problem in numerical linear algebra, and it has many applications in machine learning. Given a matrix $M \in \mathbb{R}^{n \times n}$, a non-negative weight matrix $W \in \mathbb{R}_{\geq 0}^{n \times n}$, a parameter $k$, the goal is to output two matrices $X,Y\in \mathbb{R}^{n \times k}$ such that $\| W \circ (M - X Y^\top) \|_F$ is minimized, where $\circ$ denotes the Hadamard product. It naturally generalizes the well-studied low rank matrix completion problem. Such a problem is known to be NP-hard and even hard to approximate assuming the Exponential Time Hypothesis~\citep{gg11,rsw16}. Meanwhile, alternating minimization is a good heuristic solution for weighted low rank approximation. In particular,~\cite{llr16} shows that, under mild assumptions, alternating minimization does provide provable guarantees. In this work, we develop an efficient and robust framework for alternating minimization that allows the alternating updates to be computed approximately. For weighted low rank approximation, this improves the cost of each iteration of~\cite{llr16} from $\|W\|_0k^2$ to $\|W\|_0 k$ where $\|W\|_0$ denotes the number of nonzero entries of the weight matrix. At the heart of our framework is a high-accuracy multiple response regression solver together with a robust analysis of alternating minimization. We also replace the random initialization of~\cite{llr16}, whose claimed guarantee fails, by subspace iteration from a Gaussian start, so our total running time of $\wt O((\|W\|_0 k+nk^3)\log(1/\epsilon))$ includes the initialization. In exchange, our guarantee needs the weight matrix to be within spectral distance $\gamma n$ of the all-ones matrix for a $\gamma$ that is polynomially small in $n$.

\end{abstract}

\fi

\section{Introduction}

Given a matrix $M\in \R^{n\times n}$, the low rank approximation problem with rank $k$ asks us to find a pair of matrices $\wt X, \wt Y\in \R^{n\times k}$ such that $\|M-\wt X\wt Y^\top \|_F$ is minimized over all rank $k$ matrices $X$ and $Y$, where $\|\cdot \|_F$ is the Frobenius norm of a matrix. Finding a low rank approximation efficiently is a core algorithmic problem that is well studied in machine learning, numerical linear algebra, and theoretical computer science. The exact solution follows directly from singular value decomposition (SVD): let $M=U\Sigma V^\top$ and set $\wt X=U_k\sqrt{\Sigma_k}$, $\wt Y=V_k\sqrt{\Sigma_k}$, i.e., picking the space spanned by the top-$k$ singular values and corresponding singular vectors. Faster algorithms utilizing linear sketches can run in input sparsity time~\citep{cw13}. In addition to the standard model and Frobenius norm, low rank approximation has also been investigated in distributed setting~\citep{bwz16}, for entrywise $\ell_1$ norm~\citep{swz17} and for tensors~\citep{swz19}.

In practice, it is often the case that some entries of $M$ are more important than others and some entries can be completely ignored, so it's natural to look for a \emph{weighted} low rank approximation. More specifically, given a target matrix $M\in \R^{n\times n}$ and a non-negative weight matrix $W\in \R_{\geq 0}^{n\times n}$, the goal is to find $\wt X, \wt Y\in \R^{n\times k}$ with $\|W\circ (M-\wt X\wt Y^\top) \|_F$ minimized, where $\circ$ is the Hadamard product of two matrices. The formulation of weighted low rank approximation covers many interesting matrix problems, for example, the classic low rank approximation can be recovered by setting $W={\bf 1}_n{\bf 1}_n^\top$ and the matrix completion problem~\citep{jns13} is by observing a subset of entries of $M$, equivalent to picking $W$ as a Boolean matrix. In addition to its theoretical importance, weighted low rank approximation also has a significant practical impact in many fields, such as natural language processing~\citep{psm14,all+16,hhc+22}, collaborative filtering~\citep{sj03,kbv09,lkls13,clz+15}, ecology~\citep{rjms19,khwh22}, chromatin conformation reconstruction~\cite{ths22} and statistics~\citep{wahfk97,mrpkh06}.

Algorithmic study for weighted low rank approximation dates back to~\cite{y41}. On the computational hardness front,~\cite{gg11} has shown that the general weighted low rank approximation is NP-hard even for rank $k=1$. The hardness is further enhanced by~\cite{rsw16} by showing that assuming the Random Exponential Time Hypothesis, the problem is hard to approximate beyond a constant factor. Despite its hardness, many heuristic approaches have been proposed and witnessed many successes. For example,~\cite{s90} implements gradient-based algorithms, while~\cite{lpw97,la03} use the alternating minimization framework.~\cite{sj03} develops algorithm based on expectation-maximization (EM). Unfortunately, all these approaches are without provable guarantees.~\cite{rsw16} is the first to provide algorithms with theoretical guarantees. They propose algorithms with parameterized complexity on different parameters of $W$, such as the number of distinct columns or low rank. In general, these algorithms are not polynomial which is also indicated by their lower bound results.~\cite{bwz19_neurips} subsequently studies the weighted low rank approximation problem with regularization, and they manage to obtain an improved running time depending on the statistical dimension of the input, rather than the rank. When one relaxes to a bi-criteria solution with additive error guarantees,~\cite{brw21} provides a greedy algorithm. Whenever all entries of the weight matrix are nonzero, Dai shows that it is possible to convert the additive error to multiplicative~\citep{d23}.

\paragraph{Note on hardness assumptions.} On 26 October 2026, OpenAI~\citep{openai26eth} and Anthropic~\citep{anthropic26eth} independently announced deterministic algorithms, with Lean formalizations, that solve $k$-SAT in time $2^{o(n)}$ for every fixed $k$. As of 27 October 2026, neither proof had been peer reviewed. If the proofs are correct, they refute the Exponential Time Hypothesis (ETH), and hence also the Strong Exponential Time Hypothesis (SETH), which implies ETH~\citep{ipz98,ip01}. The inapproximability barrier of~\cite{rsw16} that this paper quotes, stated in the abstract under ETH, assumes the Random Exponential Time Hypothesis~\citep[Assumption~1.3]{rsw16}, an average-case form of ETH about random $4$-CNF formulas (see also~\cite{swz19}). An exact algorithm of this kind also decides random formulas, so this hypothesis would fail as well, and the barrier would no longer stand as stated. The NP-hardness proved by~\cite{gg11} does not depend on ETH, and neither do the results of this paper.

How to bypass the barrier of~\cite{rsw16} while still getting provable guarantees?~\cite{llr16} draws inspirations from matrix completion literature and views the problem as a low rank matrix recovery problem: suppose the matrix $M\in \R^{n\times n}$ is a noisy, full-rank observation that can be decomposed into $M=M^*+N$ where $M^*$ is the rank-$k$ ground truth and $N$ is the rank-$(n-k)$ noise matrix. They then analyze the performance of alternating minimization when 1).\ the ground truth is incoherent, 2).\ weight matrix has a spectral gap to all-$1$'s matrix, and 3).\ weight matrix is non-degenerate. Under these assumptions, they show that the alternating minimization algorithm provably finds a pair of matrices $\wt X, \wt Y\in \R^{n\times k}$ such that $\|M^*-\wt X\wt Y^\top\|\leq O(k\tau/\alpha)\cdot \|W\circ N\|+\epsilon$, where $\|\cdot \|$ is the spectral norm of a matrix, $\tau$ is the condition number of $M^*$ and $\alpha$ is the non-degeneracy parameter of the weight matrix (see Assumption~\ref{assump:weighted_low_rank}). This provides a solid theoretical ground on why alternating minimization works for weighted low rank approximation. 

While the~\cite{llr16} analysis provides a polynomial time algorithm for weighted low rank approximation under certain assumptions, the algorithm itself is still far from efficient. In particular, the alternating minimization framework requires one to solve $O(n)$ different linear regressions exactly per iteration. The overall runtime of their iterations is $O((\|W\|_0\cdot k^2+nk^3)\log(1/\epsilon))$ where $\|W\|_0$ denotes the number of nonzero entries in $W$, and the SVD start that their proved guarantee needs adds $O(n^3)$ time (see Section~\ref{sub:our_results:compare}), making it inefficient for practical deployment. Moreover, their analysis is non-robust, meaning that it cannot account for any error at each step. This is in drastic contrast with practice, where floating point errors and inexact solvers are used everywhere. In fact, there are good reasons for them to mandate exact regression solvers, as their algorithm only requires $\log(1/\epsilon)$ iterations to converge and any fast but approximate regression solver might break the nice convergence behavior of the algorithm. Hence, we ask the following question:

\begin{center}
    {\it Is it possible to obtain a faster and more robust alternating minimization-based algorithm with a similar convergence rate?}
\end{center}

In this paper, we provide a positive answer to this question. Specifically, we show that the alternating updates can be computed in \emph{nearly linear time} each iteration and \emph{errors of size $1/\poly(n)$} in each update can be tolerated. Both of these results rely on a fast, randomized and high-accuracy regression solver that uses sketching to compute a preconditioner. We summarize our main result in the following theorem:

\begin{theorem}[Informal version of Theorem~\ref{thm:main}]\label{thm:main_informal}
Under the assumptions of Theorem~\ref{thm:main}, there is an algorithm (see Algorithm~\ref{alg:main}, started from the fast SVD initialization of Algorithm~\ref{alg:fast_svd_init}) that runs in $\wt{O}((\|W\|_0\cdot k+nk^3) \log(1/\epsilon))$ time and, with high probability, outputs a matrix $\wt M$ of rank at most $k$ such that
\begin{align*}
    \|\wt M-M^*\| \leq & ~ O(\alpha^{-1}k\tau)\cdot \|W\circ N\|+\epsilon
\end{align*}
where $\tau$ is the condition number of $M^*$, $\alpha$ is the parameter of Assumption~\ref{assump:weighted_low_rank}, and $\wt O(\cdot)$ suppresses polylogarithmic factors in $n$ and $k$ and $\poly(\log\log(1/\epsilon))$ factors.
\end{theorem}

\begin{algorithm*}
    \caption{Main Algorithm. 
    The \textsc{Clip} procedure zeros out rows whose $\ell_2$ norm are large, and the \textsc{QR} procedure computes the QR decomposition of the matrix and outputs the orthonormal factor $Q$.}
    \label{alg:main}
    \begin{algorithmic}[1]
    \Procedure{FasterWeightLowRank}{$M \in \R^{n \times n}$, $W \in \R^{n \times n}$, $\epsilon$,$k$}  \Comment{Theorem~\ref{thm:main_informal}}
    \State $T \gets O(\log(1/\epsilon))$
    \State {\color{black} $\delta_{\mathrm{sk}} \gets 1/\poly(n, T)$}
    \State {\color{black} $\epsilon_{\mathrm{sk}} \gets 1/\poly(n, \tau)$} \Comment{$\tau$ is an estimate of the condition number of $M^*$.}
    \State $Y_0 \gets \textsc{FastSVDInit}(M, W, k)$ \Comment{Subspace iteration on $W\circ M$ from a Gaussian start, then clip and orthonormalize (Algorithm~\ref{alg:fast_svd_init}).}
    \label{line:init} 
    \For{$t=1$ to $T$}\label{line:for_loop}
            \State {\color{black} $\vec{X}_{t} \gets \textsc{FastMultipleRegression}(M^\top, Y_{t-1}, W^\top, \epsilon_{\mathrm{sk}}/\poly(n) , \delta_{\mathrm{sk}} )^\top$} \Comment{Solve $O(n)$ regressions using sparsity of $W$ and Algorithm~\ref{alg:iter_regression}.}\label{line:fast_mul_reg_y}
        \State $\wh{X}_{t} \gets \textsc{Clip}( \vec{X}_{t})$ \Comment{Clip rows with large $\ell_2$ norms.}
        
        \State $X_{t} \gets \textsc{QR}(\wh{X}_{t})$ 
        \State {\color{black} $\vec{Y}_{t} \gets \textsc{FastMultipleRegression}(M, X_{t}, W, \epsilon_{\mathrm{sk}}/\poly(n) ,\delta_{\mathrm{sk}})^\top$} \label{line:fast_mul_reg_x}
        \State $\wh{Y}_{t} \gets \textsc{Clip}(\vec{Y}_{t})$ 
        \State $Y_{t} \gets \textsc{QR}(\wh{Y}_{t})$
    \EndFor
    \State \Return $\wt{M} \gets \wh{X}_{T} Y_{T - 1}^\top$ 
    \EndProcedure 
    \end{algorithmic}
\end{algorithm*}

{\color{black}
\begin{remark}
    The general structure of our main algorithm (Algorithm~\ref{alg:main}) is based on the traditional alternating minimization method described in \cite{llr16}. We replace the exact update with an approximate update (lines~\ref{line:fast_mul_reg_y} and~\ref{line:fast_mul_reg_x}) based on Algorithm~\ref{alg:iter_regression}, which makes the overall algorithm both faster and more robust. We also replace the random initialization of \cite{llr16} by a fast approximate SVD (Algorithm~\ref{alg:fast_svd_init}), because the guarantee claimed for the random initialization fails (Remark~\ref{rem:random_init_gap}). The remainder of the paper is dedicated to presenting a theoretical guarantee for its efficiency and robustness. 
\end{remark}
}
\paragraph{Roadmap.}
In Section~\ref{sec:notation}, we introduce several basic notations and definitions which we will use throughout this paper. 
In Section~\ref{sec:technique_overview}, we give a brief overview of our techniques. In Section~\ref{sec:our_results}, we present our main result. In Section~\ref{sec:conclusion}, we give a conclusion for this paper.

\section{Preliminary}
\label{sec:notation}

In Section~\ref{sub:notation:notation}, we introduce the basic notation used in this paper. In Section~\ref{sub:notation:sketching}, we present the background of the sketching technique, including the $\mathsf{SRHT}$ matrix and oblivious subspace embedding. In Section~\ref{sub:notation:background}, we present the mathematical background and assumptions related to the weighted low rank approximation problem.


\subsection{Notation}
\label{sub:notation:notation}

Let $n, m$ be arbitrary positive integers. We define a set $[n]$ as $\{ 1, 2, \cdots, n\}$. We use $\R$, $\R^m$, $\R_{\geq 0}^m$, and $\R^{n \times m}$ to denote the sets containing all the real numbers, $m$-dimensional vectors with real entries, $m$-dimensional vectors with non-negative real entries, and $n \times m$ matrices with real entries. 

Let $x \in \R_{\geq 0}^m$ and $w \in \R_{\geq 0}^m$. Let $i \in [m]$. Let $x_i \in \R$ represent the $i$-th entry of $x$. We use $\sqrt{x} \in \R^{m}$ to represent a vector satisfying $(\sqrt{x})_i = \sqrt{x_i}$. We define $\| x\|_w := (\sum_{i=1}^m w_i x_i^2)^{1/2}$.

Let $A, W$ be two arbitrary matrices in $\R^{n \times m}$. Let $i \in [n]$ and $j \in [m]$. We use $A_{i, :} \in \R^m$ to represent a column vector that is equal to the $i$-th row of $A$ and $A_{:, j} \in \R^n$ represent a column vector that is equal to the $j$-th column of $A$. $A_{i, j} \in \R$ represents a entry of $A$, located at the $i$-th row and $j$-th column. $\diag ( x ) \in \R^{n \times n}$ represents the matrix satisfying $\diag ( x )_{i, j} = x_i$ if $i = j$ and $\diag ( x )_{i, j} = 0$ if $i \neq j$. $\nnz(A)$ represents the number of nonzero entries of $A$.

Suppose that $n \geq m$. We denote the spectral norm of $A$ as $\| A \| = \sup_{x \in \R^m} \| A x \|_2 / \| x \|_2$, denote the Frobenius norm of $A$ as $\| A \|_F$, which is equal to $(\sum_{i=1}^n \sum_{j=1}^m A_{i,j}^2 )^{1/2}$, and denote $\| A \|_{\infty,1}$ as $\max \{ \max_{i\in [n]} \| A_{i,:} \|_1, \max_{j\in [m]} \| A_{:,j} \|_1 \}$.

Further, let $U\Sigma V^\top$ be the singular value decomposition (SVD) of $A$. Then, we have $U \in \R^{n \times m}$ and $\Sigma, V \in \R^{m \times m}$, where $U, V$ have orthonormal columns and $\Sigma$ is a non-negative diagonal matrix. The Moore-Penrose pseudoinverse of a matrix $A$ is $A^\dagger = V \Sigma^{-1} U^\top$. If $\Sigma$ is a sorted diagonal matrix and $\sigma_1, \cdots, \sigma_m$ represent the diagonal entries of $\Sigma$, then we use $\sigma_i$ to represent the $i$-th singular value of $A$, namely $\sigma_i(A)$. We define $\sigma_{\min}(A) := \min_x \| A x \|_2 / \| x \|_2$ and $\sigma_{\max}(A) := \max_x \| A x \|_2 / \| x \|_2$. 

Now, we suppose that $m = n$, namely $A, W \in \R^{n \times n}$. We define $\| A \|_W := \sqrt{\sum_{i=1}^n \sum_{j=1}^n W_{i, j} A_{i, j}^2}$. $W \circ A$ is a matrix whose entries are defined as $(W \circ A)_{i,j} := W_{i,j} A_{i,j}$. We define $D_{W_i} := \diag( W_{:, i} )$ (the weights of the $i$-th column) and $D_{W^i} := \diag( W_{i, :} )$ (the weights of the $i$-th row). If $A$ is invertible, then the true inverse of $A$ is denoted as $A^{-1}$ and $\| A \| = 1/\sigma_{\min}(A^{-1})$. If $A$ is symmetric, then we define as $U \Lambda U^\top$ the eigenvalue decomposition of $A$, where $\Lambda$ is a diagonal matrix. Let $\lambda_1, \cdots, \lambda_n$ represent the entries on diagonal of $\Lambda \in \R^{n \times n}$. $\lambda_i$ is called the $i$-th eigenvalue, namely $\lambda_{i}(A)$. Furthermore, the eigenvalue and the singular value satisfy $\sigma_i^2(A) = \lambda_i (A^\top A)$. Given two $n\times n$ real symmetric matrices $A$ and $B$, we use $A\preceq B$ to denote the matrix $B-A$ is positive semidefinite, i.e., for any $x\in \R^n$, $x^\top (B-A)x\geq 0$. 

\subsection{Sketching}
\label{sub:notation:sketching}

An important algorithmic subroutine is the Subsampled Randomized Hadamard Transform {\sf SRHT}:

\begin{definition}[{\sf SRHT}~\citep{ldfu13}]
The {\sf SRHT} matrix of size $m\times n$ is the following matrix: $S=\frac{1}{\sqrt m} PHD$, where $D\in \R^{n\times n}$ is a diagonal matrix with diagonal being Rademacher random variables, $H\in \R^{n\times n}$ is the Hadamard matrix and $P\in \R^{m\times n}$ is a row sampling matrix that samples $m$ rows with replacement. 
\end{definition}

The key property we would like to leverage from {\sf SRHT} is the \emph{subspace embedding property}:

\begin{definition}[Oblivious subspace embedding~\citep{s06}]
Let $n, d$ be positive integers and $\epsilon,\delta\in (0,1)$ be parameters, we say a distribution $\Pi$ over $m\times n$ real matrices satisfy $(\epsilon,\delta,n,d)$-oblivious subspace embedding (OSE) if for any fixed matrix $A\in \R^{n\times d}$ and $S\sim \Pi$, with probability at least $1-\delta$, we have for any $x\in \R^d$,
\begin{align*}
    (1-\epsilon)\|Ax\|_2 \leq \|SAx\|_2 \leq (1+\epsilon)\|Ax\|_2.
\end{align*}
\end{definition}

Via standard matrix concentration inequalities such as matrix Chernoff bound (see e.g.~\cite{r99,aw02}), one can show {\sf SRHT} with $m=O(\epsilon^{-2} d\log^2(n/\delta))$ satisfying $(\epsilon,\delta,n,d)$-OSE. Moreover, since $H$ is a Hadamard matrix, applying $S$ to an $n$-dimensional vector can be done in $O(n\log n)$ using FFT. Thus, computing $SA$ takes $O(nd\log n)$ time.





\subsection{Background on Weighted Low Rank Approximation}
\label{sub:notation:background}

The weighted low rank approximation can be treated as a generalization of the noisy matrix completion problem, where the goal is to recover a target matrix $M\in \R^{n\times n}$ from a few observations (sublinear in $n^2$) where the weight is chosen as a Boolean matrix $P_\Omega\in \R^{n\times n}$. It is hence natural to impose and generalize assumptions from matrix completion if we would like to obtain any provable guarantees. Following~\cite{llr16}, we make three assumptions and we will justify them one by one.

\begin{assumption}
\label{assump:weighted_low_rank}
Given a noisy, possibly higher-rank observation $M\in \R^{n\times n}$ such that $M=M^*+N$, where $M^*$ is the rank-$k$ ground truth we want to recover and $N$ is the noise matrix. We assume:
\begin{enumerate}
    \item \label{def:mu_incoherent} $M^*$ is $\mu$-incoherent: Let $M^*=U\Sigma V^\top$ be its SVD, we assume
    \begin{align*}
        \max\{\|U_{i,:}\|_2^2, \|V_{i,:}\|_2^2 \}_{i=1}^n \leq & ~ \frac{\mu k}{n}.
    \end{align*}
    {\color{black} We use $\tau$ to denote the condition number of $M^*$: $\tau=\sigma_{\max}(M^*)/\sigma_{\min}(M^*)$.}
    \item \label{def:gamma_spectral_gap} Weight $W$ has a $\gamma$-spectral gap to all-1's matrix: 
    \begin{align*}
        \|W-{\bf 1}_n{\bf 1}_n^\top \| \leq & ~ \gamma n.
    \end{align*}
    \item \label{def:alpha_beta_bounded} Weight $W$ is $(\alpha,\beta)$-bounded: Let $M^*=U\Sigma V^\top$ be its SVD, we assume for any $i\in [n]$ and $0<\alpha\leq 1\leq \beta$,
    \begin{align*}
        \alpha I \preceq U^\top D_{W_i} U\preceq \beta I, \\
        \alpha I \preceq V^\top D_{W^i} V\preceq \beta I.
    \end{align*}
\end{enumerate}
\end{assumption}

Assumption~\ref{def:mu_incoherent} states that the largest row norms of the left and right singular factors should not be too far away from the average. Such matrix incoherence assumption has been very standard in the context of matrix completion~\citep{cr12} as it effectively eliminates the degenerate case where the ground truth $M^*$ has very weak signals. Consider the extreme case where $M^*=e_1e_1^\top$, in such a scenario, if the weight $W$ is rather uniform over all entries and $N$ is a dense noise matrix with its first entry has a small magnitude compared to other entries, then recovering $M^*$ will be next to impossible. The incoherence assumption makes sure that the row and column space of $M^*$ are spread over coordinates. Incoherence is also commonly observed in practice~\citep{mt11}.

Assumption~\ref{def:gamma_spectral_gap} is a natural generalization of the \emph{random sampling assumption} for matrix completion~\citep{jns13,h14}. In particular, if $W=\frac{n}{d}B$ for a Boolean matrix $B$ where each row has $d=\Omega(\log n)$ entries chosen uniformly at random, then with high probability $\gamma=O(\frac{1}{\sqrt{d}})$, e.g., $\gamma=O(\frac{1}{\sqrt{\log n}})$ for $d=\Theta(\log n)$. Generalize to a non-negative weight setting, it also bounds the largest possible magnitude of any entry in $W$ to avoid degeneracy.

Assumption~\ref{def:alpha_beta_bounded} is also best understood when $W$ is a Boolean matrix, so that $D_{W_i}$ and $D_{W^i}$ select subsets of rows of $U$ and $V$, {\color{black}and the condition essentially reduces to Assumption A2 of~\cite{bj14}. It is a strengthening and weighted generalization of the strong incoherence property as it directly implies the assumption in~\cite{ct10}, which is necessary for matrix completion.}

Having justified the assumptions we impose on the ground truth and the weight, we are in the position to state the weighted low rank approximation problem.

\begin{problem}\label{prob:weighted_low_rank}
Let $M\in \R^{n\times n}$ be a noisy, higher-rank matrix with $M=M^*+N$ where $M^*$ is the rank-$k$ ground truth and $N$ is a higher-rank noise matrix. Let $W\in \R^{n\times n}_{\geq 0}$ be a non-negative weight matrix. Suppose both $M^*$ and $W$ satisfy Assumption~\ref{assump:weighted_low_rank}. The goal is to find a rank-$k$ matrix $\wt M\in \R^{n\times n}$ such that
\begin{align*}
    \|\wt M-M^*\| \leq & ~ \delta\cdot \|W\circ N\|+\epsilon
\end{align*}
by observing the matrix $W\circ M$.
\end{problem}

When $W$ is a Boolean matrix, Problem~\ref{prob:weighted_low_rank} reduces to the noisy matrix completion problem where one needs to recover the rank-$k$ ground truth by observing a few entries of a higher-rank noisy matrix. 
\section{Technical Overview}
\label{sec:technique_overview}

In this section, we provide a preliminary overview of the techniques we use in this paper. 
Before diving into our algorithm and analysis, let us first review the algorithm of~\cite{llr16}. At each iteration, the algorithm alternates by solving two weighted multiple response regressions: starting with an initial matrix $Y$, it tries to find a matrix $X\in \R^{n\times k}$ that minimizes $\|M-XY^\top\|_W^2$, then they zero out the rows of $X$ with large $\ell_2$ norms and use the QR factor of $X$ to proceed. Then, they alternate and solve $\min_{Y\in \R^{n\times k}} \|M-XY^\top\|_W^2$ given the new $X$. Here $\|A\|_W^2=\sum_{i,j}W_{i,j}A_{i,j}^2$, so $\|M-XY^\top\|_W$ equals the objective $\|W\circ(M-XY^\top)\|_F$ of the introduction with $W$ replaced by its entrywise square root. The two objectives agree when $W$ is Boolean. After properly zeroing out large rows and QR, the algorithm proceeds to the next iteration. The main runtime bottleneck is to solve the weighted multiple response regression per iteration.

Following the trend of low rank approximation~\citep{cw13} and fixed parameter tractable algorithm for weighted low rank approximation~\citep{rsw16}, it is natural to consider using sketching to speed up the multiple response regression solves. Let us consider
\begin{align}
\label{eq:multiple_Y}
    \min_{Y\in \R^{n\times k}} \|M-XY^\top\|_W^2.
\end{align}
Let $D_{\sqrt{W_i}}$ denote the $n\times n$ diagonal matrix that puts $\sqrt{W_i}$ on the diagonal, where $W_i$ is the $i$-th column of $W$. It is not hard to verify that~\eqref{eq:multiple_Y} can be cast into $n$ linear regressions (see details in Claim~\ref{cla:X_Y_n_LR}), each of which is in the form of
\begin{align*}
    \min_{y\in \R^k} \|D_{\sqrt{W_i}}M_{:,i}-D_{\sqrt{W_i}}Xy \|_2^2.
\end{align*}
To solve these regressions faster, one can pick a random sketching matrix $S\in \R^{s\times n}$ where $s= O(\epsilon_0^{-2}k)$ and instead solve
\begin{align*}
    \min_{y\in \R^k} \|SD_{\sqrt{W_i}}M_{:,i}-SD_{\sqrt{W_i}} Xy\|_2^2.
\end{align*}
By picking a sparse sketching matrix $S$~\citep{nn13}, the above regression can be solved in $\wt O(\epsilon_0^{-1}\nnz(X)+\epsilon_0^{-2}k^3)$ time with high probability, and the output solution $y$ has cost at most $(1+\epsilon_0)\cdot {\rm OPT}$ where OPT is the optimal regression cost. Aggregate over $n$ regressions, this gives an $\wt O(\epsilon_0^{-1}n\cdot \nnz(X)+\epsilon_0^{-2}nk^3)$ time per iteration (see Lemma~\ref{lem:linear_regression}). 

This approach, however, has several drawbacks that make it infeasible for our application. The first is the error guarantee of such approximates regression solves. Essentially, we compute a matrix $\wt Y\in \R^{n\times k}$ such that 
\begin{align*}
 \|M-X\wt Y^\top\|_W^2 \leq & ~  (1+\epsilon_0)\cdot \min_{Y\in \R^{n\times k}}\|M-XY^\top \|_W^2,
\end{align*}
in other words, the approximate solution $\wt Y$ provides a relative \emph{forward error}. Unfortunately, the {\color{black} forward} error is much less helpful when we want to analyze how close $\wt Y$ is to the optimal solution $Y$, i.e., the \emph{backward error}. It is possible to convert forward error to backward error at the expense of dependence on other terms such as the cost of the regression and the spectral norm of $(D_{\sqrt{W_i}}X)^\dagger$, the pseudo-inverse of $D_{\sqrt{W_i}}X$. To cancel out the effect of these extra terms, we will have to set the error parameter $\epsilon_0$ to be very small, thus, a polynomial dependence on $\epsilon_0^{-1}$ in the running time is unacceptable.

This motivates us to design a fast and high precision regression solver whose $\epsilon_0$ dependence is $\log(1/\epsilon_0)$ (see Lemma~\ref{lem:alg_takes_time}). Given an algorithm that produces an $(1+\epsilon_0)$ relative forward error of regression in $\log(1/\epsilon_0)$ iterations, we can set $\epsilon_0$ to inverse proportionally to ${\rm OPT}\cdot \|(D_{\sqrt{W_i}} X)^\dagger \|$. Our analysis needs accurate solutions only for the regressions with $\sigma_{\min}(X^\top D_{W_i} X)>0.25\alpha$, and for these $\|(D_{\sqrt{W_i}} X)^\dagger\|<2/\sqrt{\alpha}$. Theorem~\ref{thm:main} assumes $\alpha\geq 1/\poly(n)$ and $\|N\|_\infty\leq \poly(n)$, and its spectral-gap assumption gives $\|W\|_\infty\leq \poly(n)$, so both factors are polynomially bounded. It also assumes $\|W\circ N\|\geq 1/\poly(n)$, so the accuracy that our analysis asks of each solution is $1/\poly(n)$ (Claim~\ref{cla:log_epsilon_sk}). Hence $\log(1/\epsilon_0)=O(\log n)$, and this incurs an extra $\log n$ term in the runtime. It remains to devise a regression solver with such runtime behavior. Our approach is to use the sketch as a preconditioner: we pick a dense sketching matrix $S\in \R^{s\times n}$ with $s=\wt O(k)$ rows such that $\|Sx\|_2 = (1\pm 0.01)\cdot \|x\|_2$ for every vector $x$ in the $k$-dimensional column space of $D_{\sqrt{W_i}}X$. We then apply $S$ to $D_{\sqrt{W_i}}X$ to form a small $s\times k$ matrix and compute the QR decomposition of this matrix. It turns out that the right QR factor of $SD_{\sqrt{W_i}}X$ is a good preconditioner to $D_{\sqrt{W_i}}X$. We then use $S$ to find a constant approximation to the regression problem and utilize it as a starting point. The algorithm then iteratively performs gradient descent to optimize towards an $\epsilon_0$-approximate solution. Overall, such an algorithm takes $\log(1/\epsilon_0)$ iterations to converge, and each iteration can be implemented in $\wt O(nk)$ time. Plus the extra $\wt O(nk+k^3)$ time to compute the initial solution, this yields an algorithm that runs in $\wt O((nk+k^3)\log(1/\epsilon_0))$ time to compute an $\epsilon_0$ forward error solution. Note here we sacrifice the input sparsity time in exchange of a sketching matrix that works with high probability. This also accounts for the fact that both $X$ and $Y$ are quantities changed across iterations and the sparsity cannot be controlled. 

The runtime can be further improved by leveraging the sparsity of the weight matrix $W$. Again, consider the regression $\min_{y\in \R^k}\|D_{\sqrt{W_i}} M_{:,i}-D_{\sqrt{W_i}} Xy\|_2^2$, if $W_i$ only has a few nonzero entries, then the diagonal matrix $D_{\sqrt{W_i}}$ will effectively zero out most rows of $X$ and entries of $M_{:,i}$, and we drop these zero rows before calling the solver (Algorithm~\ref{alg:multiple_regression}). This means that we are solving a regression of size $O(\|W_i\|_0 k)$ instead of $O(nk)$. As we iterate through all $n$ regressions, the total instance size is then $O(\sum_{i=1}^n \|W_i\|_0 k)=O(\|W\|_0 k)$, and we can effectively solve these regressions in an overall $\wt O((\|W\|_0 k+nk^3)\log(1/\epsilon_0))$ time. We note that in matrix completion, $\|W\|_0$ is oftentimes $\wt O(n\poly(k))$, making it much smaller than $O(n^2)$ and an algorithm that exploits its sparsity is therefore much more valuable. Our guarantee, however, does not cover uniformly sampled weights this sparse, because they do not meet the spectral-gap condition of Definition~\ref{def:gamma_upper_bound} (see Section~\ref{sub:our_results:compare}).

We want to remark that our high precision and dense regression solver not only works for weighted low rank approximation, but for any alternating minimization frameworks that require one to solve $O(1)$ multiple response regressions per iteration. Due to the good error dependence, the overall $\log(1/\epsilon)$ convergence is well-preserved, even though each iteration is only solved approximately. We believe this high precision solver will also find its use in problems like (low rank) matrix sensing and tasks in which backward error for multiple response regression is required.

In addition to our high-accuracy, high probability solver, we also devise a robust analytical framework for alternating minimization, which is the core to enable us with fast approximate solvers. In particular, we show that if we only output a matrix $\wt Y$ whose rows are close in $\ell_2$ norm to those of the exact regression solution $Y$ (only the rows of well-conditioned regressions need to be close), then the alternating minimization still converges to an error of $O(\alpha^{-1}k\tau)\cdot \|W\circ N\|$ with good speed. Our analysis uses a different strategy from~\cite{llr16} where they heavily rely on the closed-form of the regression solution. In contrast, we show that by a clever decomposition of errors, one can absorb the error caused by approximate solves into the noise term $O(\alpha^{-1}k)\cdot \|W\circ N\|$ of each update (Claim~\ref{cla:epsilon_sk_condition}) and thus polylogarithmically more rounds of the iterative solve suffices to give us good guarantees. We expect that our analysis can be adapted to noiseless matrix completion~\citep{gsyz23}, but we do not carry out this adaptation here. As noted above, Theorem~\ref{thm:main}, as proved, needs $\|W\circ N\|\geq 1/\poly(n)$ and a spectral gap that sparse sampled weights do not have.

Finally, the induction needs a starting matrix that is incoherent and already close to the right singular space $V$ of $M^*$. \cite{llr16} offers two starts, an exact SVD of $W\circ M$ and a random start. The exact SVD costs $O(n^3)$ time, and the claimed guarantee for the random start fails (Remark~\ref{rem:random_init_gap}). We instead run $O(\log n)$ rounds of subspace iteration on $W\circ M$ from a Gaussian start, and then clip and orthonormalize the result (Algorithm~\ref{alg:fast_svd_init}). Since $W\circ M$ is close to $M^*$ in spectral norm, its $k$-th singular value is larger than its $(k+1)$-th by a large constant factor. So each round shrinks the angle to the top-$k$ right singular space of $W\circ M$ by a constant factor, and $O(\log n)$ rounds give a start as good as the exact SVD (Lemma~\ref{lem:fast_svd_init}). This takes $O((\|W\|_0k+nk^2)\log n)$ time.

\section{Our Results}
\label{sec:our_results}

In Section~\ref{sub:our_results:weighted}, we analyze the weighted multiple response regression. In Section~\ref{sub:our_results:robust}, we show that the alternating minimization framework is robust, namely this alternating minimization framework can tolerate the error induced by the approximate solver and error conversion. In Section~\ref{sub:our_results:main_result}, we present the formal version of our main result. Finally, in Section~\ref{sub:our_results:compare}, we compare our results and contribution with those of prior works.

\subsection{Weighted Multiple Response Regression}
\label{sub:our_results:weighted}

One of our cornerstone results is a novel adaptation of a high-accuracy regression solver based on sketching. Its root can be perhaps traced back to~\cite{rt08}, and our two new insights are: 1).\ This type of high-accuracy regression solvers can also be generalized to \emph{weighted case}, where the design matrix and target vector are scaled by some non-negative weights. 2).\ We can convert the error on the \emph{cost} of the regression to the error on the \emph{solution}. This step is crucial, as to bridge the gap between our fast, approximate solves and the exact solutions used in~\cite{llr16}, it is essentially to quantify the difference between solutions.

\begin{lemma}
Let $A\in \R^{n\times d}$, $b\in \R^n$ and $w\in \R^n_{\geq 0}$. Let $\epsilon\in (0,0.1)$ be an accuracy parameter and $\delta\in (0,1)$ be the failure probability. Suppose ${\cal T}(n, d, \epsilon,\delta)$ is the runtime of a black-box regression solver that produces a vector $x'\in \R^d$ such that
\begin{align*}
    \|Ax'-b\|_2 \leq & ~ (1+\epsilon) \min_{x\in \R^d} \|Ax-b\|_2
\end{align*}
with probability at least $1-\delta$. Then, there exists an algorithm that runs in time
\begin{align*}
    O(\nnz(w)\cdot d)+{\cal T}(\nnz(w),d,\epsilon,\delta)
\end{align*}
and outputs a vector $x'\in \R^d$ such that with probability at least $1-\delta$,
\begin{align*}
    \|Ax'-b\|_w \leq & ~ (1+{\color{black} \epsilon}) \min_{x\in \R^d} \|Ax-b\|_w.
\end{align*}
\end{lemma}

The proof relies on a simple observation: the weights could be applied by scaling rows of $A$ and entries of $b$, which in turn could be implemented in nearly linear time. Rows with zero weight do not affect $\|Ax-b\|_w$, so they can be dropped. This simple reduction allows us to deploy a fast off-the-shelf regression solver for weighted regression. To facilitate the analysis, we also require a conversion from the regression cost to how close our approximate solution is to the optimal solution.


\begin{lemma}
Let $A\in \R^{n\times d}$ with $n\geq d$ and full rank, $b\in \R^n$ and let $x_{\OPT}$ be the exact solution to the regression problem $\min_{x\in \R^d} \|Ax-b\|_2$. Suppose there exists a vector $x'\in \R^d$ with 
\begin{align*}
    \|Ax'-b\|_2 \leq & ~ (1+\epsilon) \|Ax_{\OPT}-b\|_2,
\end{align*}
then we have
\begin{align*}
    \|x'-x_{\OPT}\|_2 \leq & ~ O(\sqrt{\epsilon}) \cdot \frac{1}{\sigma_{\min}(A)}\cdot \|Ax_{\OPT}-b\|_2.
\end{align*}
\end{lemma}
The conversion from forward to backward error is standard~\citep{psw17,gsyz23}, and it means that we will have to set $\epsilon$ to be polynomially small in $\sigma_{\min}(A)$ and the cost of the optimal solution. We combat this issue by employing a high-accuracy regression solver.

\begin{algorithm}[!ht]
\caption{High precision solver.}
\label{alg:iter_regression}
\begin{algorithmic}[1]
\Procedure{HighPrecisionReg}{$A\in \R^{n\times d}, b\in \R^n, \epsilon\in (0,1), \delta\in (0,1)$} \Comment{Lemma~\ref{lem:alg_takes_time}}
\State $\epsilon_1\gets 0.01$
\State $m\gets O(\epsilon_1^{-2}\cdot d\log^2(n/\delta))$
\State Let $S\in \R^{m\times n}$ be an {\sf SRHT} matrix
\State Compute QR decomposition of $SA=QR^{-1}$
\State $x_0\gets \arg\min_{x\in \R^d} \|S ARx-Sb\|_2$
\State $T\gets C\cdot \log(1/\epsilon)$ for sufficiently large constant $C$
\For{$t=0\to T$}
\State $x_{t+1}\gets x_t+R^\top A^\top (b-ARx_t)$
\EndFor
\State \Return $Rx_T$
\EndProcedure
\end{algorithmic}
\end{algorithm}

The rough idea behind Algorithm~\ref{alg:iter_regression} is to compute a quick preconditioner using sketching. Let $S\in \R^{m\times n}$ be an {\sf SRHT} matrix with $m=O(\epsilon_1^{-2} d\log^2(n/\delta))$ rows, it is an $(0.01,\delta,n,d)$-OSE, therefore with high probability, the singular values of $SA$ are close to $A$. The QR decomposition of $SA$ provides an orthonormal basis $Q$ and a non-singular upper triangular matrix $R^{-1}$ which serves as a good preconditioner for $A$. We can then proceed with preconditioned gradient descent using $R$. This procedure is particularly fast because the most time-consuming step is to compute the QR decomposition, but it is performed on an $m\times d$ matrix. Further, $SA$ can be carried out in nearly linear time, and all subsequent steps in gradient descent can be performed in a manner that takes nearly linear time. The property of {\sf SRHT} also ensures our initial point $x_0$ is a constant approximation of the optimal point, therefore the algorithm converges in $O(\log(1/\epsilon))$ iterations, as desired. For more details, we refer readers to Appendix~\ref{sec:app_sketch}.

\begin{lemma}
Given a matrix $A\in \R^{n\times d}$ with full column rank and a vector $b\in \R^n$, let $\epsilon \in (0, 0.1)$ and $\delta \in (0,0.1)$, there exists an algorithm that takes time 
\begin{align*}
    O((nd\log n+d^3\log^2(n/\delta))\log(1/\epsilon))
\end{align*}
and outputs $x' \in \R^d$ such that
\begin{align*}
\| A x' - b \|_2 \leq (1+\epsilon) \min_{x \in \R^d} \| A x - b \|_2
\end{align*}
holds with probability $1-\delta$.
\end{lemma}

\subsection{Robustness Analysis for Approximate Update}
\label{sub:our_results:robust}

Now that we have the regression solvers that can compute an approximate update in nearly linear time, we need to show that the alternating minimization framework is robust enough to tolerate the large error induced by the approximate solver. We introduce a generalized incoherence notion.

\begin{definition}
Let $A\in \R^{n\times k}$, we define the generalized incoherence of $A$ as $\rho(A) = \frac{n}{k}\cdot \max_{i\in [n]}\{\|A_{i,:}\|_2^2 \}$.
\end{definition}

As our analysis crucially exploits the interplay between exact and approximate updates, we summarize the notations in the following table to simplify the discussion.

\begin{table}[!ht]
    \centering
    \caption{Summarization of notations regarding exact and approximate regression solves. By ``clipped'', we mean zeroing out rows with large $\ell_2$ norms.}
    \begin{tabular}{|l|l|}
    \hline
    {\bf Notation} & {\bf Meaning} \\
    \hline
      $\wt X$ & Matrix for exact regression solve \\
      \hline
       $\ov X$  & Clipped matrix of $\wt X$ \\
       \hline
       $X$ & QR factor of $\wh X=XR$ \\
       \hline
       $\vec X$ & Matrix for approximate regression solve \\
       \hline
       $\wh X$ & Clipped matrix of $\vec X$ \\
       \hline
    \end{tabular}
    \label{tab:X_notation_main}
\end{table}


\begin{lemma}
Let $Y\in \R^{n\times k}$ be a matrix with orthonormal columns and $\xi$ and $\epsilon_{\rm sk}$ be parameters and $\Delta_u$ be defined as in Definition~\ref{def:Delta_u}. Suppose that $M^*$ and $W$ satisfy Assumption~\ref{assump:weighted_low_rank} with $\sigma_{\max}(M^*)=1$, that $\rho(Y)\leq 25\mu\tau^2$ and $\dist(Y,V)^2\leq \frac{\alpha^2 n}{400\mu k^2\|W\|_{\infty,1}}$ (which implies $\cos\theta(Y,V)\geq 1/2$), and that $\epsilon_{\rm sk}\leq 10^{-4}\Delta_f^2\|W\circ N\|^2$. Let $\wt X, \ov X, X, \vec X$ and $\wh X$ be defined as in Table~\ref{tab:X_notation_main} with the clipping threshold being $4\xi$. Moreover, we have $\|\vec X_{i,:}-\wt X_{i,:}\|_2^2 \leq \epsilon_{\rm sk}/n$ for every $i$ with $\sigma_{\min}(Y^\top D_{W^i} Y)>0.25\alpha$. Finally, let $M^*=U\Sigma V^\top$. 

Then, we have
\begin{itemize}
    \item $\|\wh X-M^* Y\|_F^2 \leq \Delta_u^2$;
    \item If $\Delta_u\leq 0.1\sigma_{\min}(M^*)$, then $\dist(U, X)\leq 8\Delta_u/\sigma_{\min}(M^*)$;
    \item If $\Delta_u\leq 0.1\sigma_{\min}(M^*)$, then $\rho(X)\leq 25\mu/\sigma_{\min}(M^*)^2$;
\end{itemize}
where $\dist(U, X)=\min_{Q\in O_{k\times k}} \| UQ - X \|$ where $O_{k\times k}$ is the set of all $k\times k$ orthogonal matrices.
\end{lemma}

Let us interpret the above lemma and explain why it's crucial to our final convergence analysis. For simplicity, suppose the noise $N=0$ and $M^*=X^* Y^\top$, since $Y$ has orthonormal columns, $M^* Y=X^*$ and the first part states that if we solve the regression approximately and clip rows with large norms, then the approximate clipped matrix $\wh X$ is close to $X^*$. The next two parts state that as long as $\wh X$ and $X^*$ are close enough, then two crucial properties are guaranteed: 1).\ the distance between the space spanned by left singular vectors and $X$, the QR factor of the clipped matrix $\wh X$, is small and 2).\ the generalized incoherence of $X$ is small. These guarantees lead to a natural inductive argument: suppose $\Delta_u$ is small enough, then by our algorithm, we know that $\wh X$ and $M^* Y$ are close and consequently $\dist(X, U)$ and $\rho(X)$ are small. These two conditions serve as a basis to prove that for the next iteration, we still have $\wh X$ and $M^* Y$ is small enough and the induction can proceed. 

We want to highlight the major challenges in proving these assertions. Note that the induction argument effectively provides bounds on both subspace distance and generalized incoherence, and both notions heavily rely on the conditioning of intermediate matrices. The original analysis of~\cite{llr16} gives quantitative bounds on condition numbers assuming the updates are computed exactly, but the picture becomes much less clear when the updates are only computed approximately. Nevertheless, we prove that when the approximate updates are close enough to the optimality, then these bounds still hold. To compute these updates to high-precision, we utilize the high-accuracy, weighted multiple response solver being developed. One could view our proof as a mixture of algorithm and analysis: our analysis mandates the algorithm to provide strong guarantees, and we in turn design algorithms to achieve these goals. For more details, we refer readers to Appendix~\ref{sec:app_key}.

\subsection{Main Result} 
\label{sub:our_results:main_result}

Our main theorem is as follows:

\begin{theorem}[Formal version of Theorem~\ref{thm:main_informal}]
\label{thm:main}
    Let $M\in \R^{n\times n}$ be a noisy, possibly higher-rank observation with $M=M^*+N$, where $M^*$ is the rank-$k$ ground truth with $\sigma_{\max}(M^*)=1$ and $N$ is the noise matrix, and let $W\in \R^{n\times n}_{\geq 0}$ be a weight matrix. Suppose that $M^*$ and $W$ satisfy Assumption~\ref{assump:weighted_low_rank}, that $\gamma$ satisfies the condition in Definition~\ref{def:gamma_upper_bound} (with $c_0\in[1/6,1/2]$), that
    \begin{align*}
        1/\poly(n) \leq & ~ \|W\circ N\| \leq c\,\alpha^2\sigma_{\min}(M^*)\,(n/\|W\|_{\infty,1})^{1/2}/(\sqrt{\mu}\, k^2), \\
        \Delta_1 \leq & ~ c\,\alpha\,(n/\|W\|_{\infty,1})^{1/2}/(\sqrt{\mu}\, k^{1.5}),
    \end{align*}
    for a sufficiently small absolute constant $c>0$, where $\Delta_1 = 10(\gamma\mu k+\|W\circ N\|)/\sigma_{\min}(M^*)$ (Definition~\ref{def:Delta_1}), and that $\alpha\geq 1/\poly(n)$ and $\|N\|_\infty\leq \poly(n)$. Then there is an algorithm (Algorithm~\ref{alg:main}, started from the fast SVD initialization of Algorithm~\ref{alg:fast_svd_init}) that runs in $O(\log(1/\eps))$ iterations and, with probability at least $1-O(1/n^2)$, generates a matrix $\wt{M}\in\R^{n\times n}$ of rank at most $k$ such that
    \begin{align*}
        \|\wt{M} - M^*\| \leq O( \alpha^{-1} k \tau ) \|W \circ N\| + \eps.
    \end{align*}
     The total running time is 
    \begin{align*}
        \wt{O}( (\|W\|_0\cdot  k + n k^3)\log(1/\epsilon)).
    \end{align*}
\end{theorem}
Due to space limitation, we delay the proof of Theorem~\ref{thm:main} to Appendix~\ref{sec:proof_thm_main}. We want to briefly remark that our algorithm can be run when both $W$ and $M$ are rectangular matrices of size $m \times n$. Our analysis, however, is written for square matrices only. Assumption~\ref{assump:weighted_low_rank}, the threshold $\xi=\mu k/n$ and Lemma~\ref{lem:W_infty_norm} all use $m=n$, so this extension is not proved here. We expect that one could replace the factor $n$ in our runtime by $\max\{m,n\}$ when dealing with rectangular weighted low rank approximation.

\subsection{Comparisons with Recent Works}
\label{sub:our_results:compare}


In this section, we provide a brief overview and comparison with other recent works, which could be classified into 3 categories: 1).\ slower, exact alternating minimization for weighted low rank approximation~\citep{llr16}; 2).\ faster, approximate alternating minimization for noiseless low rank matrix completion, a strictly simpler problem~\citep{gsyz23} and 3).\ new metrics for measuring the effectiveness of low rank matrix factorizations~\citep{yzls22,zyls24}.

 Compared to the result of~\cite{llr16}, we significantly improve the running time from $O((\|W\|_0 k^2+nk^3)\log(1/\epsilon))$ to $\wt O((\|W\|_0 k+nk^3)\log(1/\epsilon))$. Our bound includes the initialization. The bound of~\cite{llr16} quoted here is the cost of its iterations. Its random start avoids an $O(n^3)$ SVD, but its claimed guarantee for this start fails (Remark~\ref{rem:random_init_gap}). The proved guarantee of~\cite{llr16} needs the SVD start, which adds $O(n^3)$ time. We replace the random start with subspace iteration from a Gaussian start, which takes $O((\|W\|_0k+nk^2)\log n)$ time (Lemma~\ref{lem:fast_svd_init}). For moderately large $k$ (say $k=O(\sqrt n)$) and dense weight matrix (say $\|W\|_0=O(n^2)$), the~\cite{llr16} algorithm would take $O(n^3\log(1/\epsilon))$ time, while ours only takes $\wt O(n^{2.5}\log(1/\epsilon))$ time. In the noisy matrix completion setting~\cite{kllst23}, $W$ is a Boolean matrix with $\|W\|_0=\wt O(nk^{2+o(1)})$, and the running time bound of Theorem~\ref{thm:main} evaluates to $\wt O(nk^{3+o(1)}\log(1/\epsilon))$, which nearly matches the state-of-the-art~\citep{kllst23}. This is a comparison of running time bounds only, because a uniformly sampled $W$ rescaled to mean $1$ has $\gamma=\Theta(\sqrt{n/\|W\|_0})$, so Definition~\ref{def:gamma_upper_bound} requires $\|W\|_0\geq n^{1+2c_0}\cdot\poly(k,\tau,\mu,\alpha^{-1})$.
 In contrast, the running time bound of the~\cite{llr16} algorithm evaluates to $\wt O(nk^{4+o(1)}\log(1/\epsilon))$ in this setting (with the same caveat). Moreover, our analysis accounts for the approximated computation at each step, thus it opens up the gate for further speedup. This also better depicts the picture of \emph{practical} alternating minimization algorithms, where updates are computed approximately both due to floating point errors and efficiency concerns. We believe this also lays a foundation for theoretically verifying why alternating minimization with approximate updates has great empirical success.

Compared to the result of~\cite{gsyz23}, we note they show for the simpler problem of noiseless matrix completion, the alternating minimization procedure could be sped up and run in $\wt O(\|W\|_0\cdot k\log(1/\epsilon))$ time. Our model, even in the matrix completion setting, is more general than theirs, as they assume access to the entries of the ground truth $M^*$. Still, our theorem does not contain theirs (see the caveat above on sparse sampled weights and the noiseless case below). Our model can only access noisy entries $M=M^*+N$, thus our recovery result suffers an error in the form of $O(\alpha^{-1}k\tau)\cdot \|W\circ N\|$, which is 0 if $N={\bf 0}_{n\times n}$. However, Theorem~\ref{thm:main}, as proved, needs $\|W\circ N\|\geq 1/\poly(n)$, because it sets $\epsilon_{\mathrm{sk}}=\Theta(\Delta_f^2\|W\circ N\|^2)$. The noiseless case would need $\epsilon_{\mathrm{sk}}$ chosen in terms of $\epsilon$ instead, which we expect to cost an extra $\log(1/\epsilon)$ factor in the running time. We do not carry out this case here. We also provide spectral norm error guarantee rather than Frobenius norm error, which is the objective~\cite{gsyz23} obtains. The spectral norm is oftentimes considered more robust than the Frobenius norm. In terms of analysis, the proof approach of \cite{gsyz23} is particularly geared towards noiseless matrix completion, while our analysis is much more general, as it can account for noisy matrix completion and weighted low rank approximation. We believe the generality and simplicity of our framework could be further extended to analyze alternating minimization for other problems, such as robust PCA and multi-view learning.

Compared to the results of \cite{yzls22,zyls24}, we focus on providing theoretical guarantees on the algorithm's performance, whereas 
\cite{yzls22,zyls24} focus more on analyzing the optimization landscape and proposing complexity metrics for low rank matrix problems than developing specific algorithms for weighted low rank approximation. Specifically, the main contribution of \cite{zyls24} is developing a new complexity metric to characterize the difficulty of the nonconvex landscape arising from the Burer-Monteiro factorization. This metric aims to quantify when local search methods can successfully solve the factorized problem. The main contribution of \cite{yzls22} is constructing a class of low-complexity matrix completion problem instances that can be solved in polynomial time, but for which the popular Burer-Monteiro factorization approach fails. \cite{yzls22} also shows the existence of problem instances in this class that have exponentially many spurious local minima when using the Burer-Monteiro factorization, even though the original problem has a unique global solution. It would be interesting to study whether alternating minimization could also provide provable guarantees against these metrics and problems and in turn be accelerated.

\section{Conclusion}\label{sec:conclusion}

In this paper, we study the weighted low rank approximation problem and efficient algorithm to solve it under mild recovery assumptions. Alternating minimization has been shown to be a powerful algorithmic prototype for this problem~\citep{llr16}, and we provide a fast, approximate implementation together with a robust error analysis for the framework. To this end, we improve the running time of the iterations of~\cite{llr16} from $O((\|W\|_0 k^2+nk^3)\log(1/\epsilon))$ to $\wt O((\|W\|_0 k+nk^3)\log(1/\epsilon))$. Our error analysis also serves as a theoretical explanation of why alternating minimization works well in practice especially when these updates are computed approximately for better efficiency.

We would also like to point out that the runtime of our algorithm is nearly linear in terms of solution verification. Given the weight matrix $W$ and a pair of low rank factors $X$ and $Y$, it takes $O(k)$ to verify a single entry of $W\circ (XY^\top)$ and we would need to verify a total of $\|W\|_0$ entries. Our running time bound includes the initialization. An exact SVD of $W\circ M$ would take $O(n^3)$ time and dominate the bound. The random start of~\cite{llr16} avoids that cost, but its claimed guarantee fails on a simple example (Remark~\ref{rem:random_init_gap}). We instead use $O(\log n)$ rounds of subspace iteration from a Gaussian start (Algorithm~\ref{alg:fast_svd_init}), which give a start as good as the exact SVD in $O((\|W\|_0k+nk^2)\log n)$ time (Lemma~\ref{lem:fast_svd_init}). An interesting open problem is whether the spectral-gap condition of Definition~\ref{def:gamma_upper_bound}, which shrinks polynomially in $n$, can be relaxed to a gap that does not depend on $n$. The aim is to cover sparse random weights as in matrix completion. 





\section*{Acknowledgement}

This work was mostly done when Junze Yin was at Boston University. Junze Yin is supported by the Rice University graduate fellowship. Lichen Zhang is supported in part by NSF CCF-1955217 and NSF DMS-2022448. The corrections in this version were found and written with the help of Claude Opus 5.5. The authors have verified the corrections and take full responsibility for the content of this paper.
\ifdefined\isarxiv

\else
\bibliography{ref}
\bibliographystyle{iclr2025_conference}
\fi

\newpage
\onecolumn
\appendix
\section*{Appendix}
\paragraph{Roadmap.}

In Section~\ref{sec:app_basic}, we provide several basic definitions and tools. In Section~\ref{sec:app_related}, we discuss more related work. 
In Section~\ref{sec:app_sketch}, we describe the fast multiple response regression solver used to speed up the alternating minimization step. In Section~\ref{sec:app_key}, we provide our key lemmas for the update step. In Section~\ref{sec:app_induction}, we prove our induction lemma. In Section~\ref{sec:app_prev}, we state several tools from previous work. In Section~\ref{sec:app_init_svd}, we analyze the exact SVD initialization. In Section~\ref{sec:app_init_fast}, we analyze the fast SVD initialization used by Algorithm~\ref{alg:main}, and we explain why the random initialization of \cite{llr16} is not used. In Section~\ref{sec:app_final}, we show how to prove the final guarantee of our main Theorem. In Section~\ref{sec:proof_thm_main}, we present the complete proof of our main theorem.

\section{Basic Definitions and Algebra Tools}
\label{sec:app_basic}

 In Section~\ref{sub:app_basic:weighted_property}, we present the properties of the weight matrix. Moreover, we explain the algebra tools which are used for later proofs. 
In Section~\ref{sub:app_basic:inequalities}, we present some basic algebraic inequalities. In Section~\ref{sub:app_basic:property}, we state a few simple facts about the norm properties. 

\subsection{Properties of Weight Matrix}
\label{sub:app_basic:weighted_property}

Here, we present the properties of weighted matrices.

\begin{definition}
For weight matrix $W$, we define
\begin{align*}
    \| W \|_{\infty,1} := \max\{ \max_{i \in [n]} \| W_{i,:} \|_1, \max_{j \in [n] } \| W_{:,j} \|_1 \} 
\end{align*}
and 
\begin{align*}
    \| W \|_{\infty,2} := \max\{ \max_{i \in [n]} \| W_{i,:} \|_2, \max_{j \in [n] } \| W_{:,j} \|_2 \} .
\end{align*}
\end{definition}

\begin{lemma}\label{lem:W_infty_norm}
Let $\gamma > 0$, if $\| W - {\bf 1}_n {\bf 1}_n^\top \| \leq \gamma n$, then we have
\begin{itemize}
    \item Part 1. $\| W - {\bf 1}_n {\bf 1}_n^\top \|_F \leq n^{1.5} \gamma$
    \begin{itemize}
        \item Further $\| W \|_F \leq n^{1.5} \gamma + n$
    \end{itemize}
    \item Part 2. $\| W - {\bf 1}_n {\bf 1}_n^\top \|_{\infty,1} \leq n^{1.5} \gamma$
    \begin{itemize}
        \item Further $\| W \|_{\infty,1} \leq n^{1.5} \gamma + n$
    \end{itemize}
\end{itemize}
\end{lemma}
\begin{proof}

{\bf Proof of Part 1.}
We have
\begin{align}\label{eq:fnorm_bound}
\| W - {\bf 1}_n {\bf 1}_n^\top \|_F^2 
\leq & ~ n \| W - {\bf 1}_n {\bf 1}_n^\top \|^2 \notag\\
\leq & ~ n \cdot (\gamma n)^2 \notag\\
\leq & ~ n^3 \gamma^2,
\end{align}
where the first step follows from {\bf Part 4} of Fact~\ref{fac:norm}, the second step follows from the assumption from the lemma statement, and the last step follows from simple algebra.

Moreover, by the triangle inequality, we have
\begin{align*}
    \| W \|_F
    = & ~ \| W - {\bf 1}_n {\bf 1}_n^\top + {\bf 1}_n {\bf 1}_n^\top \|_F\\
    \leq & ~ \| W - {\bf 1}_n {\bf 1}_n^\top \|_F + \|{\bf 1}_n {\bf 1}_n^\top \|_F\\
    \leq & ~ n^{1.5} \gamma + \|{\bf 1}_n {\bf 1}_n^\top \|_F\\
    = & ~ n^{1.5} \gamma + n,
\end{align*}
where the first step follows from simple algebra, the second step follows from the triangle inequality, the third step follows from Eq.~\eqref{eq:fnorm_bound}, and the last step follows from the definition of the Frobenius norm.


{\bf Proof of Part 2.} Note that 
\begin{align*}
\| W - {\bf 1}_n {\bf 1}_n^\top \|_{\infty,1} 
\leq & ~ 
\sqrt{n} \cdot \| W - {\bf 1}_n {\bf 1}_n^\top  \|_{\infty,2} \\
\leq & ~ \sqrt{n} \cdot \| W - {\bf 1}_n {\bf 1}_n^\top  \| \\
\leq & ~ n^{1.5} \gamma
\end{align*}
By the triangle inequality, we have
\begin{align*}
\| W \|_{\infty,1}
\leq & ~
\| W - {\bf 1}_n {\bf 1}_n^\top \|_{\infty, 1} + \| {\bf 1}_n {\bf 1}_n^\top \|_{\infty,1} \\
\leq & ~ n^{1.5} \gamma + n.
\end{align*}
\end{proof}

\begin{lemma}\label{lem:gamma_vs_n}
Suppose that $\|W-{\bf 1}_n{\bf 1}_n^\top\|\leq \gamma n$. Bounds on $\gamma$ lead to bounds on $\|W\|_{\infty, 1}$. Specifically,
\begin{itemize}
\item 
Part 1. If $\gamma < 1/(10 n^{1/6})$, then we have
\begin{align*}
\gamma \cdot (\| W \|_{\infty,1} / n)^{1/2} < 1
\end{align*}
\item 
Part 2. If $\gamma < 1/(10 n^{1/2})$, then we have 
\begin{align*}
\gamma \cdot (\| W \|_{\infty,1} )^{1/2} < 1
\end{align*}
\end{itemize}
\begin{remark}
In previous work \citep{llr16}, they wrote the final bound as $\gamma < f/(\| W \|_{\infty,1} / n)^{1/2} $ where $f$ are factors not depending on $\gamma$. For example, $f = \poly(\alpha^{-1}, k, \tau, \mu)$. Their bound technically is not complete, because $\| W \|_{\infty,1}$ is also function of $\gamma$. So, in our work, our Lemma~\ref{lem:gamma_vs_n} further calculates the actual condition required by \cite{llr16} and hence completes their correctness proof.
\end{remark}
\end{lemma}
\begin{proof}

{\bf Proof of Part 1.}
We need that
\begin{align*}
\gamma \cdot (\| W \|_{\infty,1} / n)^{1/2} < 1
\end{align*}
It suffices to show that
\begin{align*}
\gamma \cdot ( (n^{1.5} \gamma + n) / n)^{1/2} < 1
\end{align*}
The above equation is equivalent to
\begin{align*}
\gamma \cdot (n^{0.5} \gamma + 1)^{1/2} < 1
\end{align*}
It is sufficient to show that
\begin{align*}
\gamma^{1.5} n^{0.25} + \gamma< 1
\end{align*}
Thus, as long as
\begin{align*}
\gamma < 1/(10 n^{1/6})
\end{align*}
the promised bound is held.

{\bf Proof of Part 2.} We need that
\begin{align*}
\gamma \cdot (\| W \|_{\infty,1} )^{1/2} < 1
\end{align*}
It suffices to show that
\begin{align*}
\gamma \cdot ( n^{1.5} \gamma + n )^{1/2} < 1
\end{align*}
It is sufficient to show that
\begin{align*}
\gamma^{1.5} n^{0.75} + \gamma n^{0.5} < 1
\end{align*}
Thus, as long as
\begin{align*}
\gamma <  1/(10 n^{1/2})
\end{align*}
we have the desired result.
\end{proof}

\subsection{Basic Algebraic Inequalities}
\label{sub:app_basic:inequalities}

In this section, we introduce some basic inequalities.

\begin{fact}\label{fac:x+y_square_lower_bound}
For any $x, y$ and $\epsilon \in (0,1)$, we have
\begin{align*}
(x+y)^2 \geq (1-\epsilon) x^2 - \epsilon^{-1} y^2
\end{align*}
\end{fact}
\begin{proof}
It suffices to show
\begin{align*}
x^2 + 2 xy +y^2 \geq (1-\epsilon) x^2 - \epsilon^{-1} y^2.
\end{align*}
Re-organizing the above terms, we have
\begin{align*}
\epsilon x^2 + 2 xy + (1+\epsilon^{-1}) y^2 \geq 0
\end{align*}
Thus it suffices to show that
\begin{align*}
\epsilon x^2 + 2 xy + \epsilon^{-1} y^2 \geq 0.
\end{align*}
It is obvious  that
\begin{align*}
\epsilon x^2 +\epsilon^{-1} y^2\geq 2 |x y |. 
\end{align*}
Thus, we can complete the proof.
\end{proof}

\begin{fact}\label{fac:inequality}
Let $n$ be an arbitrary positive integer. Let $a_i \geq 0$ and $b_i \geq 0$ for all $i \in [n]$. Then, the following two inequalities hold
    \begin{align*}
        \min_{i \in [n]}\{a_i\} \sum_{i \in [n]} b_i 
        \leq & ~ \sum_{i \in [n]} a_i b_i \leq \max_{i \in [n]} \{a_i\}  \sum_{i \in [n]} b_i\\
        \min_{i \in [n]} \{b_i\} \sum_{i \in [n]} a_i 
        \leq & ~ \sum_{i \in [n]} a_i b_i \leq \max_{i \in [n]} \{b_i\}  \sum_{i \in [n]} a_i.
    \end{align*}
\end{fact}

\subsection{Properties of Norms}
\label{sub:app_basic:property}

We state some standard facts about norms without providing proofs.
\begin{fact}\label{fac:norm}
We have the following facts about norms:
\begin{itemize}
    \item Part 1. For any matrix $A \in \R^{n \times n}$, let $A_j$ denote the $j$-th column of $A$. Then we have $\sum_{j=1}^n \| A_j \|_2^2 = \| A \|_F^2$.
    \item Part 2. For any psd matrix $A$, for any vector $x$, $x^\top A x \geq \sigma_{\min}(A)\|x\|_2^2$.
    \item Part 3. Let $U \in \R^{n \times k}$ denote an orthonormal basis. Then for any $k \times k$ matrix $B$, we have $\| U B \| = \| B \|$.
    \item Part 4. For any matrix $A \in \R^{n \times k}$, we have $\| A \| \leq \| A \|_F \leq \sqrt{k} \| A \|$.
    \item Part 5. For any matrix $A$ and $B$, $\sigma_{\min}(A) \geq \sigma_{\min}(B) - \| A - B \|$
    \item Part 6. For any matrix $A \in \R^{n \times k}$ and any orthonormal basis $Q \in \R^{k \times k}$. $\sigma_{\min}(A) = \sigma_{\min}(AQ)$.
    \item Part 7. For any vector $x \in \R^{k}$ and for any orthornomal basis $Q \in \R^{k \times k}$, we have $\| x \|_2 = \| Q x \|_2$.
\end{itemize}
\end{fact}

\subsection{Generalized Matrix Incoherence}

In this section, we provide a generalized notion of matrix incoherence, denoted by $\rho$.
\begin{definition}\label{def:rho}
Let $A \in \R^{n \times k}$. The generalized incoherence of $A$ is denoted as $\rho(A)$, i.e.,
\begin{align}
    \rho (A) := \frac{n}{k} \cdot \max_{i \in [n]} \{ \| A_{i, :} \|_2^2 \}. 
\end{align}
\end{definition}

\begin{claim}\label{cla:rho}
When $A \in \R^{n \times k}$ has orthonormal columns, $1 \leq \rho( A ) \leq \frac{n}{k}$. 
\end{claim}
\begin{proof}
Since $A$ is an orthogonal matrix, $\| A_{i, :} \|_2^2 \leq 1$ for all $i \in [n]$, and thus $\rho( A ) \leq \frac{n}{k}$. In addition, $\sum_{i = 1}^{n} \| A_{i, :} \|_2^2 = k$, we have $\max_{i \in [n]} \{ \| A_{i, :} \|_2^2 \} \geq \frac{k}{n}$ and then $\rho( A ) \geq 1$. 
\end{proof}

\subsection{Angles and Distances Between Subspaces}

An important metric we use in this paper to quantify the progress of our algorithm is the distance between subspaces. We illustrate these definitions below.

\begin{definition}\label{def:angle_and_distance}
Let $X, Y$ be $n\times k$ matrices with orthonormal columns, i.e., $X^\top X = I_k$ and $Y^\top Y = I_k$.

We define $\tan \theta(Y,X)$ to be equal to
\begin{align*}
    \| Y_{\bot}^\top X ( Y^\top X )^{-1} \|.
\end{align*}

We define $\cos \theta (Y,X)$ to be equal to
\begin{align*}
    \sigma_{\min} (Y^\top X);
\end{align*}
we define $\sin \theta(Y,X)$ to be equal to
\begin{align*}
    \| (I - Y Y^\top) X \|;
\end{align*}
Let $O_k$ be a set containing all $k\times k$ orthogonal matrices. We define $\dist(Y,X)$ to be equal to
\begin{align*}
    \min_{Q \in O_k} \| YQ - X \|.
\end{align*}

Note that by their definitions, we can get
\begin{itemize}
    \item $\cos \theta(Y,X) = 1/ \| (Y^\top X)^{-1} \|$,
    \item $\cos\theta(Y,X) \leq 1$,
    \item $\sin \theta(Y,X) = \| Y_{\bot} Y_{\bot}^\top X \| = \| Y_{\bot}^\top X \|$, and
    \item $\sin \theta(Y,X) \leq 1$.
\end{itemize}
\end{definition}

\begin{lemma}[Structural lemma for orthonormal columns, Lemma A.5 of \cite{gsyz23}]
\label{lem:trig_structural}

We let $X$ and $Y$ to be arbitrary matrices in $\R^{n\times k}$ and both are orthogonal. Then, we can get
\begin{align*}
    (Y^\top X)_\bot = & ~ Y_\bot^\top X.
\end{align*}
\end{lemma}

\begin{lemma}[Lemma A.7 of \cite{gsyz23}]\label{lem:tan_is_sin_cos}
We let $X$ and $Y$ be two matrices in $\R^{n\times k}$ and both have orthonormal columns. Then, we have 
\begin{align*}
    \tan\theta(Y, X) = & ~ \frac{\sin \theta(Y,X)}{\cos \theta(Y,X)}.
\end{align*}
\end{lemma}

\begin{lemma}[Lemma A.8 of \cite{gsyz23}]
\label{lem:sin^2_and_cos^2_is_1}
Let $X, Y\in \R^{n\times k}$ be orthogonal matrices. Then, we can get 
\begin{align*}
    \sin^2\theta(Y, X) + \cos^2\theta(Y,X) =1.
\end{align*}
\end{lemma}

\begin{lemma}[Lemma A.9 of \cite{gsyz23}]
\label{lem:more_properties_angle_distance}

Let $X$ and $Y$ be two matrices in $\R^{n \times k}$ with orthonormal columns, then, we can get
    \begin{itemize}
    \item  $\tan \theta (Y, X) \geq \sin \theta (Y, X)$ 
    \item  $\tan \theta (Y, X) \geq \frac{1 - \cos \theta ( Y, X)}{\cos \theta (Y, X)}$
    \item  $\dist ( Y, X ) \geq \sin \theta (Y, X)$
    \item  $\sin \theta ( Y, X ) + \frac{1 - \cos \theta ( Y, X )}{\cos \theta ( Y, X )} \geq \dist ( Y, X )$
    \item  $2 \tan \theta ( Y, X ) \geq \dist ( Y, X )$
    \end{itemize}
\end{lemma}
\section{More Related Work}\label{sec:app_related}

\paragraph{Sketching}

To achieve the crucial speedup, we utilize sketching-based preconditioners and we therefore provide an overview of the sketching literature. Roughly speaking, given a tall dense matrix $A$, the goal of sketching is to design a family of random matrices $\Pi$ such that, if we randomly sample $S\sim \Pi$, we have
\begin{itemize}
    \item $S$ has much smaller number of rows than $A$ (thus the matrix $SA$ is close to a square matrix rather than rectangular);
    \item $SA$ preserves singular values of $A$ with high probability;
    \item $S$ can be quickly applied to $A$.
\end{itemize}

Given such a family $\Pi$, it is natural to apply an $S$ to $A$ then solve the smaller problem directly. This is the so-called \emph{sketch-and-solve} paradigm. Sketch-and-solve has led to the development of fast algorithms for many problems, such as linear regression \citep{cw13,nn13,syz23_quantum,syyz23_ellinf}, linear and kernel SVMs \citep{gsz25}, low rank approximation with Frobenius norm \citep{cw13,nn13}, matrix CUR decomposition \citep{bw14,swz17,swz19}, weighted low rank approximation \citep{rsw16}, entrywise $\ell_1$ norm low rank approximation \citep{swz17,swz19_neurips_l1}, tensor regression \citep{swyz21,rsz22,dssw18,djs+19}, tensor low rank approximation \citep{swz19}, tensor power method \citep{dsy23}, and general norm column subset selection \citep{swz19_neurips_general}. 

As modern machine learning centers around algorithms that are iterative in nature. Sketching can also be adapted to an iterative process to reduce the cost of iteration. This is the so-called \emph{Iterate-and-sketch} approach and it has led to fast algorithms for many fundamental problems, such as linear programming \citep{cls19,sy21,jswz21}, empirical risk minimization \citep{lsz19,qszz23}, semi-definite programming \citep{gs22,syz23_sp}, John Ellipsoid computation \citep{syyz22}, Frank-Wolfe algorithm \citep{xss21,sxyz22}, hamming estimation \citep{hll+24}, reinforcement learning \citep{ssx23}, $k$ means clustering \citep{lss+22}, online weighted matching problem \citep{swyy23}, barrier functions \citep{gkm+24}, softmax-inspired regression \citep{dls23_softmax,gsy23_hyper,lsz23,ssz23,lswy23,swy23,syz23}, leverage score inspired regression \citep{lsxy24}, federated learning \citep{swyz23,bsy23}, discrepancy problem \citep{dsw22,sxz22}, non-convex optimization \citep{syz21,szz21,als+22,z22}, and attention approximation \citep{gswy23,gsy23_coin}.

\paragraph{(Weighted) low rank approximation}

Low rank approximation has emerged as a crucial technique in machine learning and numerical linear algebra, enabling the extraction of essential structures from high-dimensional data while reducing computational costs. The goal is to find $X, Y \in \mathbb{R}^{n \times k}$ which minimizes $\|M - XY^\top\|_F$. It has been applied to numerous fields, including training deep neural networks \citep{szz21}, approximating attention mechanisms \citep{as23,as24,chl+24}, maintaining dynamic Kronecker products \citep{syyz23}, and tensor product regression \citep{rsz22}. 
In many practical scenarios, certain entries of \( M \) hold greater significance than others, giving rise to weighted low-rank approximation, where the objective is to minimize \( \|W \circ (M - XY^\top)\|_F \) for some weight matrix \( W \in \mathbb{R}^{n \times n}_{\geq 0} \) \citep{llr16,rsw16,gsyz23,lls+24}.

\paragraph{Subsequent work.} After our work,~\cite{llss25} studied weighted low-rank approximation when both $M$ and $W$ are dense. Let $r$ be the number of distinct rows and columns of $W$. Suppose that $W\circ M$ has at most $r\cdot n^{o(1)}$ distinct rows and columns, that $k^2r=O(\log n/\log\log n)$, and that every entry takes $n^{o(1)}$ bits. Then for every constant $\epsilon>0$, their algorithm runs in $n^{1+o(1)}$ time and, with probability at least $0.99$, outputs a rank-$k$ matrix $\widehat M$ with $\|W\circ(M-\widehat M)\|_F^2\leq(1+\epsilon)\min_{\operatorname{rank}(M')=k}\|W\circ(M-M')\|_F^2$. Their time bound is below the number $n^2$ of entries of a dense input, so it does not count reading $W$ and $M$ entry by entry. Their proof works with one representative of each distinct row and column of $W$ and $W\circ M$, together with the groups of equal rows and columns, and their paper does not say how the input is given or how these groups are found. If $W$ and $W\circ M$ are given in this form, their algorithm is faster than ours by a factor of $n^{1-o(1)}$ in this regime, since our algorithm in Theorem~\ref{thm:main} reads the nonzero entries of $W$ and $M$ and runs in time $\widetilde O(n^2k\log(1/\epsilon))$ for dense $W$. The guarantees are different. We recover the ground truth $M^*$ in spectral norm up to the error $O(\alpha^{-1}k\tau)\|W\circ N\|+\epsilon$, under Assumption~\ref{assump:weighted_low_rank} and the other hypotheses of Theorem~\ref{thm:main}, for $\epsilon$ as small as $1/\operatorname{poly}(n)$, while their condition on the distinct rows and columns of $W\circ M$ need not hold for a noisy observation $M=M^*+N$. With the same input format, their algorithm also improves upon the $n^{2+o(1)}$ time that they state for~\cite{rsw16} on dense inputs, which counts reading the input.

\section{Weighted Multiple Response Regression Solvers}
\label{sec:app_sketch}

In this section, we show how to solve weighted multiple response regression by solving standard linear regressions. We present randomized and fast regression solvers based on sketching and preconditioning.

\subsection{Generic Reduction and Error Conversion}

In this section, we present a generic framework to reduce the weighted multiple response regression problem to solving $O(n)$ ordinary least-square regressions. This simple and efficient reduction enables us to deploy fast regression solvers to handle the approximate updates in alternating minimization. We also present a tool that converts the relative error on the regression cost to the quality of approximate solution.

The first lemma states that the cost of a weighted multiple response regression can be decomposed into a summation of $n$ weighted linear regressions.

\begin{claim}\label{cla:X_Y_n_LR}
 
Given matrices $M, W \in \R^{n \times n}$ and $X, Y \in \R^{n \times k}$, we have 
\begin{align*}
 ~ \min_{X \in \R^{n \times k}} \| M - XY^\top \|_W^2 
= ~ \sum_{i = 1}^{n} \min_{X_{i, :} \in \R^{k}} \| D_{\sqrt{W^i}} Y X_{i, :} - D_{\sqrt{W^i}} M_{i, :} \|_2^2,  
\end{align*}
and
\begin{align*}
 ~ \min_{Y \in \R^{n \times k}} \| M - XY^\top \|_W^2 
=  ~ \sum_{i = 1}^{n} \min_{Y_{i, :} \in \R^k} \| D_{\sqrt{W_i}} X Y_{i, :} - D_{\sqrt{W_i}} M_{:, i} \|_2^2.  
\end{align*}
\end{claim}

\begin{proof}
Since the two equations can be proved in a similar way, we only prove the first one. 
\begin{align*}
     \min_{X \in \R^{n \times k}} \| M - XY^\top \|_W^2 
    = & ~ \min_{X \in \R^{n \times k}} \sum_{i, j} W_{i, j} (XY^\top - M)_{i, j}^2 \\
    = & ~ \min_{X \in \R^{n \times k}} \sum_{i=1}^{n} \| D_{\sqrt{W^i}} (Y X_{i, :} - M_{i, :}) \|_2^2 \\ 
    = & ~ \min_{X \in \R^{n \times k}} \sum_{i = 1}^{n} \| D_{\sqrt{W^i}} Y X_{i, :} - D_{\sqrt{W^i}} M_{i, :} \|_2^2 \\
    = & ~ \sum_{i = 1}^{n} \min_{X_{i, :} \in \R^{k}} \| D_{\sqrt{W^i}} Y X_{i, :} - D_{\sqrt{W^i}} M_{i, :} \|_2^2, 
\end{align*} 
where the 1st step is due to $\| A \|_W^2$'s definition, the 2nd step is by rewriting each row as an independent regression problem (row $i$ carries the weights $W_{i,:}$ of row $i$ of $W$), the 3rd step follows from simple algebra, and the last step follows from the fact that the $i$-th summand of $\sum_{i = 1}^{n} \| D_{\sqrt{W^i}} Y X_{i, :} - D_{\sqrt{W^i}} M_{i, :} \|_2^2$ depends on $X$ only through $X_{i, :}$. Thus, we complete the proof.
\end{proof}

The next lemma provides a simple conversion of weighted linear regression to ordinary least-squares, via a scaling trick.

\begin{lemma}[A variation of Lemma B.6 of~\cite{gsyz23}]
\label{lem:weighted_lr_meta}

Let $A$ be a real $n\times d$ matrix with $n\geq d$, $b$ be an $n$-dimensional real vector and $w$ be a non-negative $n$-dimensional vector (weight). Let $\epsilon_0\in (0,0.1)$ be accuracy parameter and $\delta_0\in (0,0.1)$ controls failure probability. Suppose that ${\cal T}(n,d,\epsilon_0,\delta_0)$ is the running time of a regression solver, and $x'\in \R^d$ is the output of the regression solver satisfying
\begin{align*}
    \|Ax'-b\|_2 \leq & ~ (1+\epsilon_0) \min_{x\in \R^d} \|Ax-b\|_2
\end{align*}
with probability at least $1-\delta_0$.

Then, there exists an algorithm whose running time is 
\begin{align*}
O(\nnz(w)\cdot d)+{\cal T}(\nnz(w), d, \epsilon_0, \delta_0)
\end{align*} 
and outputs a vector $x' \in \R^d$, which satisfy
\begin{align*}
\| A x' - b \|_w \leq (1+\epsilon_0) \min_{x \in \R^d} \| A x - b \|_w
\end{align*}
with probability at least $1-\delta_0$.
\end{lemma}
\begin{proof}
Let $S := \{ j \in [n] : w_j \neq 0 \}$. For every $x \in \R^d$, we have $\| A x - b \|_w = \| (D_{\sqrt{w}} A)_{S,:} x - (D_{\sqrt{w}} b)_S \|_2$, where $D_{\sqrt{w}} := \diag(\sqrt{w})$. So it suffices to run the regression solver on the $\nnz(w) \times d$ matrix $(D_{\sqrt{w}} A)_{S,:}$ and the vector $(D_{\sqrt{w}} b)_S$, which can be formed in $O(\nnz(w) \cdot d)$ time.
\end{proof}

One of the main reasons~\cite{llr16} resorts to exact weighted multiple response regression is that most approximate solvers provide backward error guarantees on the cost of regression. On the other hand, we would like the \emph{approximate solution} of the regression to be close to the exact solution. The following lemma converts the backward error on the cost, to the forward error on the solution.

\begin{lemma}[Backward error, Lemma B.5 in \cite{gsyz23}]
\label{lem:distance_app_opt}

Let $A$ be a real $n\times d$ matrix with $n\geq d$ and full column rank, $b$ be an $n$-dimensional real vector. Let $x_{\OPT}$ be the exact solution to the regression problem

\begin{align*}
    \min_x \|Ax-b\|_2.
\end{align*}

Suppose that there exists a vector $x'\in \R^d$, satisfying
\begin{align*}
\| A x' - b \|_2 \leq (1+\epsilon) \min_{x \in \R^d} \| A x - b \|_2. 
\end{align*}

Then, we have
\begin{align*}
    \|x' - x_{\rm OPT}\|_2 \leq & ~ O(\sqrt{\epsilon}) \cdot \frac{1}{\sigma_{\min}(A)} \cdot \|Ax_{\rm OPT}-b\|_2.
\end{align*}
\end{lemma}

Before wrapping up this section, we present a meta algorithm for solving weighted multiple response regression.

\begin{algorithm}[!ht]\caption{Fast, high precision solver for weighted multiple response regression}\label{alg:multiple_regression}
\begin{algorithmic}[1]
\Procedure{MultipleRegression}{$A \in \R^{n \times n}, B \in \R^{n \times k},W \in \R^{n \times n}$}
    \State \Comment{$A_i$ is the $i$-th column of $A$}
    \State \Comment{$W_i$ is the $i$-th column of $W$}
    \State \Comment{$D_{\sqrt{W_i}}$ is a diagonal matrix where we put $\sqrt{W_i}$ on diagonal, other locations are zero}
    \State $X_i \gets \arg\min_{x \in \R^k} \| D_{\sqrt{W_i}} B x - D_{\sqrt{W_i}} A_i \|_2$ for each $i\in[n]$
    \State \Return $X$ \Comment{$X \in \R^{k \times n}$}
\EndProcedure
\State
\Procedure{FastMultipleRegression}{$A \in \R^{n \times n}, B \in \R^{n \times k},W \in \R^{n \times n}, \epsilon\in(0,0.1), \delta\in(0,0.1)$}
    \State \Comment{$A_i$ is the $i$-th column of $A$}
    \State \Comment{$W_i$ is the $i$-th column of $W$}
    \State \Comment{$D_{\sqrt{W_i}}$ is a diagonal matrix where we put $\sqrt{W_i}$ on diagonal, other locations are zero}
    \State $X_i \gets \textsc{HighPrecisionReg}((D_{\sqrt{W_i}} B)_{S_i,:}, (D_{\sqrt{W_i}} A_i)_{S_i}, \epsilon,\delta)$ for each $i\in[n]$, where $S_i:=\{j\in[n]: W_{j,i}\neq 0\}$ \Comment{Algorithm~\ref{alg:iter_regression}. If $(D_{\sqrt{W_i}} B)_{S_i,:}$ does not have full column rank, set $X_i\gets 0$.}
    \State \Return $X$ \Comment{$X \in \R^{k \times n}$}
\EndProcedure
\end{algorithmic}
\end{algorithm}

\subsection{Low Accuracy Solver}

We provide an algorithm that uses a sparse sketching matrix to obtain a low accuracy solution (inverse polynomial dependence on accuracy parameter $\epsilon$).

\begin{definition}[OSNAP matrix,~\citep{nn13}]
\label{def:sparse_sketch}
For every sparsity parameter $s$, target dimension $m$, and positive integer $d$, the OSNAP matrix with sparsity $s$ is defined as
\begin{align*}
    S_{r,j} = & ~ \frac{1}{\sqrt s}\cdot \delta_{r,j}\cdot \sigma_{r,j},
\end{align*}
for all $r\in [m], j\in [n]$, where $\sigma_{r,j}$ are independent Rademacher random variables and $\delta_{r,j}$ are Bernoulli random variables with
\begin{itemize}
    \item For every $i\in [n]$, $\sum_{r\in [m]} \delta_{r,i}=s$, which means each column of $S$ contains exactly $s$ nonzero entries.
    \item For all $r\in [m]$ and $i\in [n]$, $\E[\delta_{r,i}]=s/m$.
    \item $\forall T\subseteq [m]\times [n]$, $\E[\prod_{(r,i)\in T} \delta_{r,i}]\leq \prod_{(r,i)\in T}\E[\delta_{r,i}]=(s/m)^{|T|}$, i.e., $\delta_{r,i}$ are negatively correlated.
\end{itemize}
\end{definition}

Crucially, the OSNAP matrix produces a subspace embedding with nearly linear in $d$ row count.

\begin{lemma}[\citep{c16}]
\label{lem:osnap_ose}
Let $S\in \R^{m\times n}$ be an OSNAP matrix as in Def.~\ref{def:sparse_sketch}. 

Let $\epsilon,\delta\in (0,1)$ be parameters. 

For any integer $d\leq n$, if 
\begin{itemize}
    \item $m=O(\epsilon^{-2}d\log(d/\delta))$;
    \item $s=O(\epsilon^{-1}\log(d/\delta))$,
\end{itemize}
then an $s$-sparse OSNAP matrix $S$ is an $(\epsilon,\delta)$ oblivious subspace embedding, i.e., for any fixed orthonormal basis $U\in \R^{n\times d}$ with probability at least $1-\delta$, and the singular values of $SU$ lie in $[1-\epsilon,1+\epsilon]$.
\end{lemma}

To distinguish with $\epsilon,\delta$ for our final algorithm, here we use $\epsilon_0, \delta_0$ for the subroutine  (approximate linear regression).

\begin{lemma}[Input sparsity and low accuracy regression]\label{lem:linear_regression}
Given a matrix $A \in \R^{n \times d}$ and a vector $b \in \R^n$, let $\epsilon_0 \in (0,0.1)$ and $\delta_0 \in (0,0.1)$, there exists an algorithm that takes time  
\begin{align*}
 O((\epsilon_0^{-1} \nnz(A)+\epsilon_0^{-2}d^3)\cdot \log(d/\delta_0))
\end{align*}
and outputs $x' \in \R^d$ such that
\begin{align*}
\| A x' - b \|_2 \leq (1+\epsilon_0) \min_{x \in \R^d} \| A x - b \|_2
\end{align*}
holds with probability $1-\delta_0$.
\end{lemma}

\begin{proof}

To obtain desired accuracy and probability guarantee, we pick $S$ to be an OSNAP (Definition~\ref{def:sparse_sketch}) with 
\begin{align*}
    m=O(\epsilon_0^{-2}d\log(d/\delta_0))
\end{align*}
and 
\begin{align*}
    s=O(\epsilon_0^{-1}\log(d/\delta_0)).
\end{align*}

We simply apply $S$ to $A$ then solve the sketched regression $\min_{x\in \R^d} \|SAx-Sb\|_2$. With $\epsilon_0/3$ in place of $\epsilon_0$ in $m$ and $s$, Lemma~\ref{lem:osnap_ose} applied to the $(d+1)$-dimensional column span of $[A,\ b]$ shows that, with probability at least $1-\delta_0$, the sketched minimizer $x'$ and the true minimizer $x^*$ satisfy $\|Ax'-b\|_2\leq \frac{1}{1-\epsilon_0/3}\|SAx'-Sb\|_2\leq \frac{1}{1-\epsilon_0/3}\|SAx^*-Sb\|_2\leq \frac{1+\epsilon_0/3}{1-\epsilon_0/3}\|Ax^*-b\|_2\leq (1+\epsilon_0)\|Ax^*-b\|_2$. 
\begin{itemize}
    \item As $S$ is a matrix where each column only has $s$ nonzero entries, the time to compute $SA$ is 
    \begin{align*}
        O(s\nnz(A))=O(\epsilon_0^{-1}\nnz(A)\log(d/\delta_0)).
    \end{align*}
    \item The regression can then be solved via normal equation, i.e., 
    \begin{align*}
        (A^\top S^\top SA)^{\dagger} A^\top S^\top b.
    \end{align*}
    The time to form the Gram matrix is 
    \begin{align*}
        O(md^2),
    \end{align*}
    computing the $d\times d$ inversion takes $O(d^3)$ time, and forming the final solution takes another
    \begin{align*}
        O(md^{2})
    \end{align*}
    time. Overall, this gives a runtime of
    \begin{align*}
        O(\epsilon_0^{-2}d^3 \log(d/\delta_0)).
    \end{align*}
\end{itemize}
Thus, the overall runtime is 
\begin{align*}
    O((\epsilon_0^{-1} \nnz(A)+\epsilon_0^{-2}d^3)\cdot \log(d/\delta_0)). 
\end{align*}
\end{proof}

\subsection{High Accuracy Solver}

Our key algorithmic ingredient is a high accuracy, iterative and sketching-based solver for regression. The sketching matrix we will be using is the dense subsampled randomized Hadamard transform {\sf SRHT} due to loss of structure in the iterative process.

\begin{definition}[Subsampled randomized Hadamard transform ({\sf SRHT}),~\citep{ldfu13}]
\label{def:srht}
Let $P$ be a random sampling matrix in $\{0, 1\}^{m \times n}$ and for each row of $P$, there exists a $1$ at a uniformly random position.

Let $H \in \{-1, 1\}^{n \times n}$ be the Hadamard matrix.

Let $D \in \R^{n \times n}$ be a diagonal matrix, whose diagonal entries are all in $\{-1, +1\}$ with the same probability.

We define the \textsf{SRHT} matrix $S\in\R^{m \times n}$ as
\begin{align*}
    S:=\frac{1}{\sqrt{m}} P H D.
\end{align*}
\end{definition}

\begin{remark}\label{rem:SA_compute_time}
For a real $n\times d$ matrix $A$ it takes $O(nd\log n)$ time to apply $S$ to $A$.
\end{remark}

\begin{lemma}
\label{lem:iter_regression}
Let $S\in \R^{m\times n}$ be an {\sf SRHT} matrix (see Definition~\ref{def:srht}), $\epsilon,\delta\in (0,1)$ be parameters. Let $d$ be an arbitrary integer, which is less than or equal to $n$. Suppose $m=O(\epsilon^{-2}d\log^2(n/\delta))$. Let $U\in \R^{n\times d}$ be a fixed orthonormal basis.

Then $S$ is an $(\epsilon,\delta)$-oblivious subspace embedding: with probability at least $1-\delta$, the singular values of the matrix $SU$ are in the interval $[1-\epsilon, 1+\epsilon]$. 

\end{lemma}

\begin{lemma}[Dense and high accuracy regression, Lemma B.1 in \cite{gsyz23}]\label{lem:alg_takes_time}
Let $A$ be a real $n\times d$ matrix with full column rank, $b$ be a real $n$-dimensional vector, $\epsilon\in (0,0.1)$ be an accuracy parameter and $\delta\in (0,0.1)$ be the failure probability. Then, there exists an algorithm that takes time 
\begin{align*}
    O((nd\log n+d^3\log^2(n/\delta))\log(1/\epsilon))
\end{align*}
and outputs a vector $x' \in \R^d$ satisfying
\begin{align*}
\| A x' - b \|_2 \leq (1+\epsilon) \min_{x \in \R^d} \| A x - b \|_2
\end{align*}
 with probability at least $1-\delta$.
\end{lemma}

\section{Key Property for Robust Update}\label{sec:app_key}

In this section, we prove crucial properties of the algorithm that enable the approximate updates. In Section~\ref{sub:app_key:update}, we formally define several necessary notations needed to analyze our robust updated step. In Section~\ref{sub:app_key:update_lemma}, we analyze the key properties of our robust update step. 

\subsection{Definitions for Update Step}
\label{sub:app_key:update}

We present a closed-form solution for linear regression via normal equation.
\begin{fact}\label{fac:exact_LR}
Define $x := \arg\min_{x} \| A x - b \|_2$, then we have
\begin{align*}
x = (A^\top A)^{-1} A^\top b.
\end{align*}

Similarly, for weighted regression, we define 
\begin{align*}
    x := \arg\min_x \|  D_{\sqrt{W_i}} A x - D_{\sqrt{W_i}} b\|_2.
\end{align*}

Then, we have
\begin{align*}
x = (A^\top D_{W_i} A)^{-1} A^\top D_{W_i} b
\end{align*}
\end{fact}

\begin{definition}\label{def:xi}
We define $\xi$ as 
\begin{align*}
\xi : =  \mu k / n.
\end{align*} 
 $\xi$ captures the maximum incoherence of the ground truth.
\end{definition}

\begin{definition}\label{def:eta}
An earlier version of this paper carried a parameter $\eta \geq 1$ that distinguished a random initialization ($\eta = \mu k$) from the SVD initialization ($\eta = 1$). This paper uses two initializations, the exact SVD initialization and the fast SVD initialization of Algorithm~\ref{alg:fast_svd_init}. Both satisfy the guarantees of the SVD case, so we set
\begin{align*}
\eta := 1
\end{align*}
throughout. We keep the symbol $\eta$ in the formulas below so that they can be compared with the earlier version. Remark~\ref{rem:random_init_gap} explains why the random initialization is not used.
\end{definition}

We next define the value choice of $\gamma$, which controls how far away the weight matrix $W$ can be from the all-1's matrix.

\begin{definition}\label{def:gamma_upper_bound}
Let 
    \begin{align*}
    \gamma \leq \frac{1}{100} \cdot \frac{\alpha^{1.5}}{ \poly(k,\tau, \mu) \cdot n^{c_0} } 
    \end{align*}
    where $c_0$ is a fixed constant in $[1/6,1/2]$.
\end{definition}

For the convenience of analysis, we define the following threshold parameters:
\begin{definition}\label{def:Delta_d_Delta_f}
Let $C \geq 10^5$ denote a sufficiently large constant.
We define
\begin{align*}
\Delta_d := & ~ C \alpha^{-1.5}  \mu^{1.5} k^{2} \gamma (\| W \|_{\infty,1}/n) ^{1/2}  + C \alpha^{-1} \eta \mu k^2 \tau  \gamma    \\
\Delta_f:= & ~ C \alpha^{-1} \eta k.
\end{align*}
The choice of $\Delta_d$ is decided in Eq.~\eqref{eq:Delta_d_condition_1} and Eq.~\eqref{eq:Delta_d_condition_2}. The choice of $\Delta_f$ is decided in Eq.~\eqref{eq:Delta_f_condition}.
\end{definition}

The next two definitions capture the error gap of our algorithm.
\begin{definition}\label{def:Delta_u}
Let $\Delta_d$ and $\Delta_f$ be defined as Definition~\ref{def:Delta_d_Delta_f}. 
We define
\begin{align*}
\Delta_u
:= & ~ \Delta_d \cdot \dist( Y, V )
+ \Delta_f \cdot \| W \circ N \| . 
\end{align*}
and
\begin{align*}
\Delta_g := & ~ 0.01 \Delta_d \cdot \dist( Y, V )
+ 0.01 \Delta_f \cdot \| W \circ N \| + 2 \sqrt{\epsilon_{\mathrm{sk}}}
\end{align*}

\end{definition}

By properly controlling the error $\epsilon_{\rm sk}$, we can show that $\Delta_g$ is a constant factor smaller than $\Delta_u$.
\begin{claim}\label{cla:epsilon_sk_condition}
If the following condition holds
\begin{align*}
\epsilon_{\mathrm{sk}} \leq 10^{-4} \Delta_f^2 \cdot \| W \circ N \|^2,
\end{align*}
then we have
\begin{align*} 
\Delta_g \leq 0.1 \Delta_u.
\end{align*}
\end{claim}
\begin{proof}
We have
\begin{align*}
\Delta_g 
= & ~ 0.01 \Delta_d \cdot \dist( Y, V )
+ 0.01 \Delta_f \cdot \| W \circ N \| + 2 \sqrt{\epsilon_{\mathrm{sk}}} \\
\leq & ~ 0.01 \Delta_d \cdot \dist( Y, V )
+ 0.01 \Delta_f \cdot \| W \circ N \| + 0.02 \Delta_f \cdot \| W \circ N \| \\
\leq & ~ 0.1 \Delta_u
\end{align*}
where the second step follows from condition on $\epsilon_{\mathrm{sk}}$, and the last step follows from the definition of $\Delta_u$.
\end{proof}

By setting the error and failure probability appropriately, we can show the extra blowups in our algorithm are of the order $\poly \log n$.

\begin{claim}\label{cla:log_epsilon_sk}
By the choice of $\epsilon_{\mathrm{sk}}$, we have
\begin{align*}
\log(n/\epsilon_{\mathrm{sk}}) = O(\log(n))
\end{align*}
By choice of failure probability ($\delta_0$) of sketch, 
\begin{align*}
\log(n/\delta_0) = \log(n \log(1/\epsilon))
\end{align*}
\end{claim}
\begin{proof}
By Lemma~\ref{lem:W_infty_norm}, we can see that 
\begin{align*}
    \| W \|_{\infty} \leq \| W \|_{\infty,1} \leq n^{1.5}\gamma + n \leq \poly(n).
\end{align*}

Together with the hypotheses $\| N \|_{\infty} \leq \poly(n)$ and $\| W\circ N \| \geq 1/\poly(n)$ of Theorem~\ref{thm:main} (the lower bound does not follow from a lower bound on $\| N \|_\infty$ alone, since $W$ may vanish on the support of $N$), we know 
\begin{align}\label{eq:1_over_poly_leq_W_circ_N}
    1/\poly(n) \leq \| W \circ N \| \leq \| W \circ N \|_F \leq \poly(n).
\end{align}

We also know that $k \leq n$.

Now, we can compute
\begin{align*}
\log(n/\epsilon_{\mathrm{sk}}) 
\leq & ~ O( \log( n  / ( \Delta_f^2 \| W \circ N \|_F^2 ) ) ) \\
\leq & ~ O( \log( n / \| W \circ N \|_F^2 ) ) \\
\leq & ~ O(\log(n)),
\end{align*}
where the first step holds for the choice $\epsilon_{\mathrm{sk}} = 10^{-4} \Delta_f^2 \| W \circ N \|^2 \geq 10^{-4} \Delta_f^2 \| W \circ N \|_F^2 / n$, the second step follows from $\Delta_f \geq 1$, the third step follows from $\| W \circ N \|_F^2 \geq 1/\poly(n)$ (see Eq.~\eqref{eq:1_over_poly_leq_W_circ_N}). The same bound holds for any choice with $10^{-4}n^{-2c_1} \leq \epsilon_{\mathrm{sk}} \leq 10^{-4}\Delta_f^2\|W\circ N\|^2$, where $\|W\circ N\| \geq n^{-c_1}$, because then $\log(n/\epsilon_{\mathrm{sk}}) \leq \log(10^4 n^{1+2c_1}) = O(\log n)$. Lemma~\ref{lem:main_correct} uses one such choice, $\epsilon_{\mathrm{sk}} = 10^{-4}n^{-2c_1}$.

Sum over all the $T = O(\log(1/\epsilon))$ iterations, so
\begin{align*}
\log(n/\delta_0) = O( \log(n \log(1/\epsilon) )).
\end{align*}
\end{proof}

\subsection{Key Lemma for Robust Update Step}
\label{sub:app_key:update_lemma}

\begin{table}[!ht]
    \centering
    \caption{For convenience, we provide a table to summarize the notations in Lemma~\ref{lem:robust_update_lemma}.}
    \begin{tabular}{|l|l|l|}
    \hline
      {\bf Notation}   & {\bf Meaning} \\ \hline
        $\wt X$ & Optimal matrix for exact regression  \\ \hline
        $\ov X$ & Clipped matrix of $\wt X$ \\ \hline
        $\vec X$ & Optimal matrix for sketched regression   \\ \hline
        $\wh X$ & Clipped matrix of $\vec X$  \\ \hline
    \end{tabular}
    \label{tab:X_notations}
\end{table}

\begin{algorithm}[!ht]
    \caption{Clipping rows whose norms are larger than a constant factor of $\xi$.}
    \label{alg:clip}
    \begin{algorithmic}[1]
    \Procedure{Clip}{$\wt{X} \in \R^{n \times k}$}
    \State $\xi \gets \frac{ \mu k}{n}$
    \For{$i=1$ to $n$}
        \If{$ \|\wt{X}_{i, :}\|_2^2 \leq 4\xi $}
            \State $\ov{X}_{i, :} \gets \wt{X}_{i, :}$
        \Else 
            \State $\ov{X}_{i, :} \gets 0$
        \EndIf 
    \EndFor
    \State \Return $\ov{X}$
    \EndProcedure
    \end{algorithmic}
\end{algorithm}

Here, we analyze the key properties of the robust update step. 

\begin{lemma}[Key lemma for update step]\label{lem:robust_update_lemma}
    Suppose that the ground truth $M^*$ and the weight matrix $W$ satisfy Assumption~\ref{assump:weighted_low_rank} and that $\sigma_{\max}(M^*)=1$, and let $\epsilon_{\mathrm{sk}}$ satisfy the condition of Claim~\ref{cla:epsilon_sk_condition}. Let $Y \in \R^{n \times k}$ be a (column) orthogonal matrix with $\rho(Y) \leq 25\mu\tau^2$ and $\dist(Y,V)^2 \leq \alpha^2 n / (400 \mu k^2 \|W\|_{\infty,1})$ (this implies $\cos\theta(Y,V) \geq 1/2$). Let $\xi$ be defined as Definition~\ref{def:xi}. Let $\Delta_u$ be defined as Definition~\ref{def:Delta_u}.

    We define matrix $\wt{X} \in \R^{n \times k}$ as follows:
    \begin{align*}
    \wt{X} := \arg \min_{X \in \R^{n \times k}} \| M - X Y^\top \|_W, 
    \end{align*} 
    
    We define matrix $\ov{X} \in \R^{n \times k}$ as follows
    \begin{align*}
        \ov{X}_{i, :} := \begin{cases}
            \wt{X}_{i, :} & \text{if $\| \wt{X}_{i, :} \|^2 \leq 4\xi$}\\
            0 & \text{otherwise}
        \end{cases}
    \end{align*}
    We define $\vec{X} \in \R^{n \times k}$ to be the sketch solution such that for all $i \in [n] \setminus T$, where $T:=\{ i \in [n] ~|~ \sigma_{\min} (Y^\top D_{W^i} Y) \leq 0.25 \alpha /  \eta \}$ (for $i \in T$, the row $\vec{X}_{i,:}$ may be arbitrary),
    \begin{align*}
        \| \vec{X}_{i,:} - \wt{X}_{i,:} \|^2 \leq \epsilon_{\mathrm{sk}} / n.
    \end{align*}
    We define $\wh{X} \in \R^{n \times k}$ to denote the clip of the sketched solution. We define $X \in \R^{n \times k},R \in \R^{k \times k}$ to be the factors of the QR decomposition of $\wh{X}$, i.e. $\wh{X} = X R$. Recall that $M^*=U\Sigma V^\top$. Then, 
    \begin{itemize}
        \item Part 1.
        \begin{align*}
            \| \wh{X} - U\Sigma V^\top Y \|_F^2 \leq \Delta_u^2; 
        \end{align*}
        \item Part 2. If $\Delta_u < 0.1 \sigma_{\min}( M^* )$, then 
        \begin{align*}
            \dist ( U, X ) \leq 8 \Delta_u / \sigma_{\min}(M^*)  
        \end{align*}
        \item Part 3. If $\Delta_u < 0.1 \sigma_{\min}( M^* )$, then 
        \begin{align*}
             \rho( X ) \leq 25 \mu / \sigma_{\min} (M^*)^2  
        \end{align*}
    \end{itemize}
\end{lemma}

\begin{proof}

 {\bf Proof of Part 1.}
Recall the weighted multiple response regression
 \begin{align*}
 \min_{X \in \R^{n \times k} } \| M - X Y^\top \|_W^2,
 \end{align*}

 The above problem can be written as $n$ different regression problems. The $i$-th linear regression has the formulation 
 \begin{align*}
 \min_{X_{i, :} \in \R^{k}} \| D_{ \sqrt{W^i} } Y X_{i, :} - D_{ \sqrt{W^i} } M_{i, :} \|^2, \quad \text{where } D_{W^i} := \diag(W_{i,:}).
 \end{align*}

We have
        \begin{align}\label{eq:wt_Xi_2_terms}
            \wt{X}_{i, :}^\top 
            = & ~ M_{i, :}^\top \cdot D_{W^i} Y ( Y^\top D_{W^i} Y )^{-1} \notag\\
            = & ~ ((M^*)_{i, :}^\top + N_{i, :}^\top) \cdot D_{W^i} Y ( Y^\top D_{W^i} Y )^{-1} \notag\\
            = & ~ (M^*)_{i, :}^\top \cdot D_{W^i} Y ( Y^\top D_{W^i} Y )^{-1} 
            +  N_{i, :}^\top \cdot D_{W^i} Y ( Y^\top D_{W^i} Y )^{-1},
        \end{align}
        where the first step follows from the Fact~\ref{fac:exact_LR} (applied with the weights $W_{i,:}$ of row $i$), and the second step follows from $M_{i, :}^\top = (M^*)_{i, :}^\top + N_{i, :}^\top$ (because $M = M^* + N$), and the third step follows from simple algebra.
        
        Given $M^* = U \Sigma V^\top$, the first term in Eq.~\eqref{eq:wt_Xi_2_terms} can be rewritten as follows:
        \begin{align}\label{eq:wt_Xi_2_terms_first}
            & ~ (M^*)_{i, :}^\top \cdot D_{W^i} Y ( Y^\top D_{W^i} Y )^{-1} \notag\\
            = & ~ U_{i, :}^\top \cdot \Sigma V^\top D_{W^i} Y ( Y^\top D_{W^i} Y )^{-1} \notag \\
            = & ~ U_{i, :}^\top \cdot \Sigma V^\top ( Y Y^\top + Y_{\bot} Y_{\bot}^\top ) D_{W^i} Y ( Y^\top D_{W^i} Y )^{-1} \notag \\
            = & ~ U_{i, :}^\top \cdot \Sigma V^\top Y Y^\top D_{W^i} Y (Y^\top D_{W^i} Y)^{-1} 
            +  U_{i, :}^\top \Sigma V^\top Y_{\bot} Y_{\bot}^\top D_{W^i} Y ( Y^\top D_{W^i} Y )^{-1} \notag \\
             = & ~ U_{i, :}^\top \cdot \Sigma V^\top Y  + U_{i, :}^\top \Sigma V^\top Y_{\bot} Y_{\bot}^\top D_{W^i} Y ( Y^\top D_{W^i} Y )^{-1},  
        \end{align}
        where the first step follows from the fact that $(M^*)_{i, :}^\top = U_{i, :}^\top \Sigma V^\top$, the second step follows from $I = Y Y^\top + Y_{\bot} Y_{\bot}^\top$, the third step follows from simple algebra, and the last step follows from $A A^{-1} = I$.
        
        Combining Eq.~\eqref{eq:wt_Xi_2_terms} and Eq.~\eqref{eq:wt_Xi_2_terms_first} we have
        \begin{align}\label{eq:wt_Xi_minus_U_Sigma_V_Y}
              \wt{X}_{i, :}^\top - U_{i, :}^\top \Sigma V^\top Y 
             = & ~ U_{i, :}^\top \Sigma V^\top Y_{\bot} Y_{\bot}^\top D_{W^i} Y ( Y^\top D_{W^i} Y )^{-1} 
             +  N_{i, :}^\top D_{W^i} Y ( Y^\top D_{W^i} Y )^{-1} 
        \end{align}

Recall the set $T \subset [n]$ from the statement:
\begin{align}\label{eq:T_set_upper_bound_sigma}
T:=\{ i \in [n] ~|~ \sigma_{\min} (Y^\top D_{W^i} Y) \leq 0.25 \alpha /  \eta \}.
\end{align}

We upper bound $|T|$.

Recall that $\eta = 1$ (Definition~\ref{def:eta}). Using Lemma~\ref{lem:lemma_10_in_llr16} with $\epsilon =3/4$ and with $Q \in O_k$ such that $\| VQ - Y \| = \dist(Y,V)$ (it satisfies the hypothesis of that lemma, because $\| VQ - Y \|_F^2 \leq k \dist(Y,V)^2 \leq \frac{\alpha^2 n}{400 \mu k \|W\|_{\infty,1}} \leq \frac{\epsilon^3 \alpha^2 n}{160 \mu k \| W \|_{\infty,1}}$), we have
\begin{align*}
|T| \leq & ~ 10^5 \cdot \alpha^{-3}  \mu^2 k^3 \gamma^2  \cdot \| W \|_{\infty,1} \cdot  \dist(V,Y)^2 \\
= & ~ 10^5 \cdot \alpha^{-3} \mu^2 k^3 \gamma^2 \cdot \mu k \cdot (\| W \|_{\infty,1} / n)  \cdot \dist(V,Y)^2  /\xi \\
\leq & ~ 10^{-5} \Delta_d^2 \cdot \dist(V,Y)^2 / \xi \\
\leq & ~ \Delta_g^2 / \xi
\end{align*}
where the second step follows from $\xi = \mu k / n$, the third step follows from Definition of $\Delta_d$ (Definition~\ref{def:Delta_d_Delta_f}) and $10^{-5} C^2 \geq 10^5$, and the last step follows from $\Delta_g \geq 0.01 \Delta_d \cdot \dist(Y,V)$. 

(In particular, the third step requires 
 \begin{align}\label{eq:Delta_d_condition_1}
 \Delta_d \geq \Omega(\alpha^{-1.5} \mu^{1.5} k^{2} \gamma \cdot ( \| W \|_{\infty,1} / n)^{1/2})
 \end{align}
 )

For each $i \in [n] \backslash T$, we have
        \begin{align}\label{eq:wt_Xi_minus_U_Sigma_V_Y_norm}
              \| \wt{X}_{i, :}^\top - U_{i, :}^\top \Sigma V^\top Y \|_2^2 
            = & ~  \| ( U_{i, :}^\top \Sigma V^\top Y_{\bot} Y_{\bot}^\top D_{W^i} Y + N_{i, :}^\top D_{W^i} Y ) ( Y^\top D_{W^i} Y )^{-1} \|_2^2 \notag \\
            \leq & ~  \| ( U_{i, :}^\top \Sigma V^\top Y_{\bot} Y_{\bot}^\top D_{W^i} Y + N_{i, :}^\top D_{W^i} Y ) \|_2^2 
            \cdot  \| ( Y^\top D_{W^i} Y )^{-1} \|^2 \notag \\
            \leq & ~ 20 \alpha^{-2} \eta^2 \cdot \| ( U_{i, :}^\top \Sigma V^\top Y_{\bot} Y_{\bot}^\top D_{W^i} Y + N_{i, :}^\top D_{W^i} Y ) \|_2^2 \notag \\
            \leq & ~ 20 \alpha^{-2} \eta^2 
            \cdot  2 ( \| U_{i, :}^\top \Sigma V^\top Y_{\bot} Y_{\bot}^\top D_{W^i} Y \|_2^2 + \| N_{i, :}^\top D_{W^i} Y \|_2^2 )  \notag \\
            \leq & ~ 40 \alpha^{-2} \eta^2 
            \cdot  ( \| U_{i, :}^\top \|_2^2 \| \Sigma \|^2 \| V^\top Y_{\bot} Y_{\bot}^\top D_{W^i} Y \|^2 + \| N_{i, :}^\top D_{W^i} Y \|_2^2 ) \notag \\
            \leq & ~ 40 \alpha^{-2} \eta^2  
            \cdot  ( \frac{\mu k}{n} \| \Sigma\|^2   \| V^\top Y_{\bot} Y_{\bot}^\top D_{W^i} Y \|^2 +   \| N_{i, :}^\top D_{W^i} Y \|_2^2 )  
        \end{align}
        where 
        the first step follows from Eq.~\eqref{eq:wt_Xi_minus_U_Sigma_V_Y}, the second step follows from $\| Ax \|_2 \leq \| A \| \cdot \| x \|_2$, the third step follows from $\sigma_{\min}( Y^\top D_{W^i} Y ) \geq 0.25 \alpha/\eta$ (for all $i \in [n]\backslash T$, see Eq.~\eqref{eq:T_set_upper_bound_sigma}), the fourth step follows from $(a+b)^2 \leq 2a^2 + 2b^2$, the fifth step follows from $\| A x \|_2 \leq \| A \| \cdot \| x \|_2$, the sixth step follows from $\| U_{i,:}^\top \|^2 \leq \mu k/n$.

Taking the summation over $i \in [n]\backslash T$ coordinates (for Eq.~\eqref{eq:wt_Xi_minus_U_Sigma_V_Y_norm}), we have
\begin{align}\label{eq:sum_i_not_in_T}
  \sum_{i \in [n] \backslash T} \| \wt{X}_{i, :}^\top - U_{i, :}^\top \Sigma V^\top Y \|_2^2 
 \leq & ~ \sum_{i \in [n] \backslash T} 40 \alpha^{-2}\eta^2 \cdot ( \frac{\mu k}{n} \| \Sigma \|^2 \| V^\top Y_{\bot} Y_{\bot}^\top D_{W^i} Y \|^2 \notag \\
 & ~ +   \| N_{i,:}^\top D_{W^i} Y \|_2^2 )  
\end{align}

 For the first term in Eq.~\eqref{eq:sum_i_not_in_T} (ignore coefficients $40\alpha^{-2}\eta^2$ and $\frac{\mu k}{n} \| \Sigma \|^2$), we have
\begin{align}\label{eq:apply_lemma_11_in_llr16}
\sum_{i\in [n] \backslash T} \| V^\top Y_{\bot} Y_{\bot}^\top D_{W^i} Y \|^2 
\leq & ~ n \gamma^2 \rho(Y)  k^3 \dist(Y,V)^2 \notag \\
\leq  & ~ 25 n \gamma^2 \mu \sigma_{\min}^{-2}(\Sigma)  k^3 \dist(Y,V)^2 
\end{align}
where the first step follows from Lemma~\ref{lem:lemma_11_in_llr16}, the last step follows from $\rho(Y) \leq 25\mu\tau^2 = 25\mu/\sigma_{\min}^2(\Sigma)$ (as $\sigma_{\max}(\Sigma)=1$).

For the second term in Eq.~\eqref{eq:sum_i_not_in_T} (ignore coefficients $40\alpha^{-2}\eta^2$), we have

\begin{align}\label{eq:get_W_circ_N_spectral}
    \sum_{i \in [n] \backslash T } \| N_{i, :}^\top D_{W^i} Y \|_2^2 
    \leq & ~  \sum_{i \in [n] } \| N_{i, :}^\top D_{W^i} Y \|_2^2 \notag \\
    = & ~ \| (W \circ N) Y \|_F^2 \notag \\
    \leq & ~ k \| ( W \circ N ) Y \|^2.
\end{align}
where the last step follows from Fact~\ref{fac:norm}.

Loading Eq.~\eqref{eq:apply_lemma_11_in_llr16} and Eq.~\eqref{eq:get_W_circ_N_spectral} into Eq.~\eqref{eq:sum_i_not_in_T}, we have
\begin{align}\label{eq:sum_i_not_in_T_next}
 \sum_{i \in [n] \backslash T} \| \wt{X}_{i, :}^\top - U_{i, :}^\top \Sigma V^\top Y \|_2^2 
 \leq & ~ 40 \alpha^{-2} \eta^2 \cdot ( \frac{\mu k}{n} \| \Sigma \|^2 \cdot 25 n \gamma^2 \mu \sigma_{\min}^{-2}(\Sigma)  k^3 \dist(Y,V)^2 \notag \\
 & ~ + k \| W\circ N \|^2 ) \notag \\
 \leq & ~ 40 \alpha^{-2} \eta^2 \cdot ( 25 \mu^2 k^4  \tau^2 \gamma^2 ) \cdot \dist (Y,V)^2 + 40 \alpha^{-2} \eta^2 k \| W \circ N \|^2
\end{align}
where the last step follows from $\|\Sigma \| =1$ and $\tau =\sigma_{\max}(\Sigma) / \sigma_{\min} (\Sigma)$.

 Thus, we have
 \begin{align*}
\sum_{i \in [n] \backslash T} \| \wt{X}_{i, :}^\top - U_{i, :}^\top \Sigma V^\top Y \|_2^2
\leq & ~ 40 \alpha^{-2} \eta^2 \cdot ( 25 \mu^2 k^4  \tau^2 \gamma^2 ) \cdot \dist (Y,V)^2 + 40 \alpha^{-2} \eta^2 k \| W \circ N \|^2 \\
\leq & ~ 10^{-5} \Delta_d^2 \cdot \dist(Y,V)^2 + 10^{-5} \Delta_f^2 \cdot \| W \circ N \|^2 \\
\leq & ~ 0.1 \Delta_g^2
 \end{align*}
 where the first step follows from Eq.~\eqref{eq:sum_i_not_in_T_next}, the second step follows from Definition~\ref{def:Delta_d_Delta_f} and $10^{-5} C^2 \geq 10^5$, and the last step follows from Definition~\ref{def:Delta_u}, which gives $\Delta_g^2 \geq 10^{-4} ( \Delta_d \cdot \dist(Y,V) + \Delta_f \cdot \| W \circ N \| )^2$.
 
 (In particular, the second step above requires 
 \begin{align}\label{eq:Delta_d_condition_2}
 \Delta_d \geq \Omega(\alpha^{-1} \eta \mu k^2  \tau \gamma)
 \end{align}
 and 
 \begin{align}\label{eq:Delta_f_condition}
\Delta_f \geq \Omega(\alpha^{-1} \eta k )
 \end{align}
 )

By Definition~\ref{def:Delta_u} and choosing $\epsilon_{\mathrm{sk}}$ to be sufficiently small as Claim~\ref{cla:epsilon_sk_condition}, we know that
\begin{align*}
\Delta_g^2 \leq 0.01 \Delta_u^2
\end{align*}

Then, we can show
        \begin{align*}
            \sum_{i \in [n] \backslash T} \| \vec{X}_{i, :}^\top - U_{i, :}^\top \Sigma V^\top Y \|_2^2 
            \leq & ~ 2 \sum_{i \in [n] \backslash T} \| \wt{X}_{i, :}^\top - U_{i, :}^\top \Sigma V^\top Y \|_2^2 + 2 \sum_{i \in [n] \backslash T} \| \vec{X}_{i, :}^\top - \wt{X}_{i,:}^\top \|_2^2 \\
            \leq & ~ \Delta_g^2,
        \end{align*}

Note that
\begin{align*}
\| U_{i,:}^\top \Sigma V^\top Y \|_2^2 \leq \mu k / n = \xi.
\end{align*}
If $\|\vec{X}_{i, :}^\top \|_2^2 \geq 4\xi $, then 
\begin{align*}
\| \vec{X}_{i, :}^\top - U_{i,:}^\top \Sigma V^\top Y \|_2 \geq 2\sqrt{\xi} - \sqrt{\xi} \geq  \sqrt{\xi}
\end{align*}
which implies that
\begin{align}\label{eq:vec_X_minus_U_Sigma_V_Y}
\| \vec{X}_{i, :}^\top - U_{i,:}^\top \Sigma V^\top Y \|_2^2 \geq \xi.
\end{align}

        We define set $S \subset [n]$ as follows
        \begin{align*}
        S:= \{ i \in [n] \backslash T ~|~ \| \vec{X}_{i, :}^\top \|_2^2 \geq 4\xi \}.
        \end{align*}

        Then we have 
        \begin{align*}
           |S|
            = & ~ | \{ i \in [n] \backslash T \mid \| \vec{X}_{i, :}^\top \|_2^2 \geq 4 \xi \} | \\
            \leq & ~ | \{ i \in [n] \backslash T \mid \| \vec{X}_{i, :}^\top - U_{i, :}^\top \Sigma V^\top Y \|_2^2 \geq \xi \} | \\
            \leq & ~ \Delta_g^2/\xi. 
        \end{align*}
        where the first step follows from the definition of $S$, the second step follows from Eq.~\eqref{eq:vec_X_minus_U_Sigma_V_Y}, the third step follows from $\sum_{i \in [n] \backslash T} \| \vec{X}_{i, :}^\top - U_{i, :}^\top \Sigma V^\top Y \|^2 \leq \Delta_g^2$.

        We can show
        \begin{align*}
            \| \wh{X} - U \Sigma V^\top Y \|_F^2 
            = & ~ \sum_{i = 1}^n \| \wh{X}_{i, :}^\top - U_{i, :}^\top \Sigma V^\top Y \|_2^2 \\
            = & ~ \sum_{i \in T \cup S} \| \wh{X}_{i, :}^\top - U_{i, :}^\top \Sigma V^\top Y \|_2^2 + \sum_{i \notin T \cup S} \| \wh{X}_{i, :}^\top - U_{i, :}^\top \Sigma V^\top Y \|_2^2 \\
            = & ~ \sum_{i \in T \cup S} \| \wh{X}_{i, :}^\top - U_{i, :}^\top \Sigma V^\top Y \|_2^2 + \sum_{i \notin T \cup S} \| \vec{X}_{i, :}^\top - U_{i, :}^\top \Sigma V^\top Y \|_2^2 \\
            \leq & ~ \sum_{i \in T \cup S} 2 ( \| \wh{X}_{i, :}^\top \|_2^2 + \| U_{i, :}^\top \Sigma V^\top Y \|_2^2 ) 
            + \sum_{i \notin T \cup S} \| \vec{X}_{i, :}^\top - U_{i, :}^\top \Sigma V^\top Y \|_2^2 \\
            \leq & ~ |T \cup S| \cdot 2 \cdot (4 \xi + \xi) + \Delta_g^2 \\
            = & ~ |T \cup S| \cdot 10 \xi + \Delta_g^2 \\
            \leq & ~ 50 \Delta_g^2 \notag \\
            \leq & ~ \Delta_u^2
        \end{align*}
            where the first step follows from the definition of $\| \wh{X} - U \Sigma V^\top Y \|_F^2$, the second step follows from $S\subseteq [n]$, the third step follows from $\wh{X}_{i, :}^\top = \vec{X}_{i, :}^\top$ when $i \notin T \cup S$, the fourth step follows from the triangle inequality, the fifth step follows from $\| \wh{X}_{i,:}^\top \|^2 \leq 4 \xi$ and $ \| U_{i, :}^\top \Sigma V^\top Y \|^2 \leq \xi $, and the sixth step follows from $| T \cup S | \leq  2 \Delta_g^2/\xi$, the last step follows from Claim~\ref{cla:epsilon_sk_condition}.

{\bf Proof of Part 2.}  We let $B = \Sigma V^\top Y$ and have 
        \begin{align}\label{eq:sin_theta_u_x}
            \sin \theta(U, X) 
            = & ~ \|U_\bot^\top X\| \notag \\
            = & ~ \|U_\bot^\top (\wh{X} - UB) R^{-1} \| \notag \\
            \leq & ~ \|(\wh{X} - UB)\| \cdot \|R^{-1} \| \notag \\
            = & ~ \frac{\|(\wh{X} - UB)\|}{\sigma_{\min} ( \wh{X} )} \notag \\
            \leq & ~ \frac{\Delta_u}{\sigma_{\min} ( \wh{X} )},
\end{align}
where the first step follows from the definition of 
\begin{align*}
    \sin \theta(U, X)
\end{align*}
(see Definition~\ref{def:angle_and_distance}), the second step follows from 
\begin{align*}
    U_\bot^\top X = U_\bot^\top \wh{X} R^{-1} = U_\bot^\top (\wh{X} - UB) R^{-1},
\end{align*}
the 3rd step is due to the Cauchy-Schwarz inequality, and the 4th step is because of 
\begin{align*}
    \|R^{-1} \| = \frac{1}{\sigma_{\min} (\wh{X}) },
\end{align*}
and the 5th step follows from 
\begin{align*}
    \|(\wh{X} - U\Sigma V^\top Y)\|^2 \leq \|(\wh{X} - U\Sigma V^\top Y)\|_F^2 \leq \Delta_u^2,
\end{align*}
which was proved in part 1, and it infers $\|(\wh{X} - UB)\| \leq \Delta_u$ for $B = \Sigma V^\top Y$. 

Using $\|(\wh{X} - UB)\| \leq \Delta_u$, we have
\begin{align}\label{eq:sigma_min_ov}
            \sigma_{\min} ( \wh{X} ) 
            \geq & ~ \sigma_{\min} ( UB ) - \Delta_u \notag \\
             = & ~ \sigma_{\min} (U \Sigma V^\top Y ) - \Delta_u \notag \\
            = & ~ \sigma_{\min} ( \Sigma V^\top Y ) - \Delta_u \notag \\
            \geq & ~ \sigma_{\min} ( M^* ) \cos \theta (Y, V) - \Delta_u \notag\\
            \geq & ~\sigma_{\min} ( M^* )/2- \Delta_u \notag \\
            \geq & ~ 0.4\,\sigma_{\min}(M^*),
\end{align}
where the second step follows from how we defined $B$, the third step follows from $U$ has orthonormal columns, the third step follows from 
\begin{align*}
    \sigma_{\min} ( \Sigma V^\top Y ) \geq \sigma_{\min} ( M^* ) \cos \theta (Y, V),
\end{align*}
where the forth step follows from $\cos \theta(Y,V) \geq 1/2$ (a consequence of the hypothesis on $\dist(Y,V)$), and the last step follows from $\Delta_u \leq \sigma_{\min}(M^*) /10$.

Then, by combining Eq.~\eqref{eq:sin_theta_u_x} and Eq.~\eqref{eq:sigma_min_ov}, we have 
        \begin{align}\label{eq:upper_bound_sin_U_X_by_Delta_u}
            \sin \theta (U, X) 
            \leq & ~ \frac{1}{\sigma_{\min} ( \wh{X} )} \Delta_u \notag \\
            \leq & ~ 2.5 \Delta_u / \sigma_{\min}(M^*).
        \end{align}
 In particular $\sin\theta(U,X) \leq 0.25$, so $\cos\theta(U,X) \geq 0.96$ by Lemma~\ref{lem:sin^2_and_cos^2_is_1}.

 Therefore,    
        \begin{align*}
            \dist( U, X ) 
            \leq & ~ 2 \tan \theta(U, X) \\
            = & ~ 2 \sin \theta(U, X) / \cos \theta(U, X) \\
            \leq & ~ 5.3 \Delta_u / \sigma_{\min}(M^*) \\
            \leq & ~ 8 \Delta_u / \sigma_{\min}(M^*). 
        \end{align*}
        where the first step follows from Part 5 of Lemma~\ref{lem:more_properties_angle_distance}, the second step follows from Lemma~\ref{lem:tan_is_sin_cos}, and the third step follows from Eq.~\eqref{eq:upper_bound_sin_U_X_by_Delta_u} and $\cos\theta(U,X) \geq 0.96$.

{\bf Proof Part 3.} Given $\wh{X} = X R$, we have $\wh{X}_{i, :}^\top = X_{i, :}^\top R$ and  
        \begin{align}\label{eq:xi_smaller_than_four}
            \| X_{i, :} \|_2^2 
            \leq & ~ \| \wh{X}_{i, :} \|_2^2 \| R^{-1} \|^2 \notag \\
            \leq & ~ \frac{4\xi}{\sigma_{\min}( \wh{X} )^2} \notag \\
            \leq & ~ \frac{4\xi}{( 0.4\,\sigma_{\min}( M^* ) )^2} \notag \\
            = & ~ 25 \xi / \sigma_{\min}(M^*)^2,
        \end{align}
        where the first step follows from $\wh{X}_{i, :}^\top = X_{i, :}^\top R$ and Cauchy-Schwarz inequality, the second step follows from $\| \wh{X}_{i, :} \|^2 \leq 4\xi$ (by clipping) and $\| R^{-1} \| = \frac{1}{\sigma_{\min}( \wh{X} )}$, the third step follows from Eq.~\eqref{eq:sigma_min_ov}, and the last step follows from simple algebra.
        
        To see this, we have 
        \begin{align*}
        \| R^{-1} \|^2 
        = & ~ \lambda_{\max} (R^{-1} ( R^{-1} )^\top ) \\
        = & ~ \lambda_{\max} ( (\wh{X}^\top \wh{X})^{-1} ) \\
        = & ~ \frac{1}{\lambda_{\min}(\wh{X}^\top \wh{X})} \\
        = & ~ \frac{1}{\sigma_{\min}^2( \wh{X} )}, 
        \end{align*}
        where the first step follows from $\| A \|^2 = \lambda_{\max}(AA^\top)$, the second step follows from  $\wh{X}^\top \wh{X} = R^\top R$, the third step follows from $\lambda_{\max} ( A^{-1} ) = \lambda_{\min} ( A )^{-1}$, and  the last step follows from $\lambda(A^\top A) = \sigma^2(A)$.  
        
        Then, 
        \begin{align*}
            \rho( X ) 
            = & ~ \max_{i \in [n]} \frac{n}{k} \| X_{i, :} \|_2^2 \\
            = & ~\frac{n}{k} \max_{i \in [n]}  \| X_{i, :} \|_2^2 \\
            \leq & ~ \frac{n}{k} \frac{25 \xi}{ \sigma_{\min}(M^*)^2} \\
            = & ~ \frac{25 \mu}{ \sigma_{\min}( M^* )^2},
        \end{align*}
    where the first step follows from the definition of $\rho( X ) $, the second step follows from simple algebra, the third step follows from Eq.~\eqref{eq:xi_smaller_than_four}, and the last step follows from $\frac{\xi n}{k} = \mu$.
\end{proof}
\section{The Analysis of the Induction Lemma}
\label{sec:app_induction}
The goal of this section is to prove induction.

\begin{lemma}[Induction]\label{lem:induction}
Let $\Delta_d$ and $\Delta_f$ be as in Definition~\ref{def:Delta_d_Delta_f}, with $\eta = 1$ (Definition~\ref{def:eta}), so that $\Delta_f = C\alpha^{-1}k$, and let
\begin{align*}
D_* := \Big( \frac{\alpha^2 n}{400 \mu k^2 \| W \|_{\infty,1}} \Big)^{1/2}, \qquad \Delta_F := 100\, \tau\, \Delta_f = 100\, \sigma_{\min}(M^*)^{-1} \Delta_f .
\end{align*}
Suppose that $M^*$ and $W$ satisfy Assumption~\ref{assump:weighted_low_rank} with $\sigma_{\max}(M^*)=1$, that $\epsilon_{\mathrm{sk}}$ satisfies the condition of Claim~\ref{cla:epsilon_sk_condition}, and that
\begin{align*}
\Delta_d \leq \frac{1}{100} \sigma_{\min} (M^*), \qquad \Delta_F \| W \circ N \| \leq D_*, \qquad \dist(Y_0, V) \leq D_*, \qquad \rho(Y_0) \leq 25 \mu \tau^2,
\end{align*}
where $Y_0$ is the initial matrix of Algorithm~\ref{alg:main}. Suppose also that every regression step of Algorithm~\ref{alg:main} meets the accuracy hypothesis of Lemma~\ref{lem:robust_update_lemma}. That is, in the step that computes $\vec X_t$ from $Y = Y_{t-1}$, every row $i$ with $\sigma_{\min}(Y^\top D_{W^i} Y) > 0.25\alpha$ satisfies $\| (\vec X_t)_{i,:} - (\wt X_t)_{i,:} \|_2^2 \leq \epsilon_{\mathrm{sk}}/n$, where $\wt X_t$ is the exact minimizer, and the same holds symmetrically for $\vec Y_t$. Then $\dist(Y_t, V) \leq D_*$ and $\rho(Y_t) \leq 25\mu\tau^2$ for $0 \leq t \leq T$, and $\dist(X_t, U) \leq D_*$ and $\rho(X_t) \leq 25\mu\tau^2$ for $1 \leq t \leq T$. Moreover, for $1 \leq t \leq T$,
 \begin{align}\label{eq:dist_x_y}
        \dist( X_t, U ) \leq \frac{1}{2^t} + \Delta_F \cdot \| W \circ N \|, \qquad
        \dist( Y_t, V ) \leq \frac{1}{2^t} + \Delta_F \cdot \| W \circ N \|,
    \end{align}
and the bound on $\dist(Y_t, V)$ also holds for $t = 0$.
\end{lemma}
\begin{proof}
Write $\sigma := \sigma_{\min}(M^*) = 1/\tau$ and $\delta := \| W \circ N \|$. First, we show that $D_* \leq 1/20$. For every $i \in [n]$ and every unit vector $a \in \R^k$, Assumption~\ref{def:alpha_beta_bounded} and the incoherence of $V$ give
\begin{align*}
\alpha \leq a^\top V^\top D_{W^i} V a = \sum_{j \in [n]} W_{i,j} \langle V_{j,:}, a \rangle^2 \leq \frac{\mu k}{n} \sum_{j \in [n]} W_{i,j},
\end{align*}
so $\| W \|_{\infty,1} \geq \alpha n/(\mu k)$ and $D_*^2 \leq \alpha/(400 k) \leq 1/400$.

\textbf{One half step.} Let $Y \in \R^{n\times k}$ have orthonormal columns with $\dist(Y, V) \leq D_*$ and $\rho(Y) \leq 25\mu\tau^2$, and let $X$ be computed from $Y$ by one regression, clipping and QR step of Algorithm~\ref{alg:main} that meets the accuracy hypothesis above. Then the hypotheses of Lemma~\ref{lem:robust_update_lemma} hold, and its $\Delta_u = \Delta_d \dist(Y,V) + \Delta_f \delta$ satisfies
\begin{align*}
\Delta_u \leq \frac{\sigma}{100} D_* + \frac{\sigma}{100} D_* \leq 0.02\, \sigma D_* < 0.1\, \sigma,
\end{align*}
because $\Delta_f \delta = \Delta_F \delta/(100\tau) \leq \sigma D_*/100$. So Parts 2 and 3 of Lemma~\ref{lem:robust_update_lemma} apply. They give $\rho(X) \leq 25\mu/\sigma^2 = 25\mu\tau^2$ and
\begin{align*}
\dist(X, U) \leq \frac{8}{\sigma} \Delta_u \leq 0.08 \dist(Y, V) + 8\tau \Delta_f \delta \leq 0.08 D_* + 0.08 D_* \leq D_*,
\end{align*}
where we used $\Delta_d \leq \sigma/100$ and $8\tau\Delta_f\delta = 0.08\,\Delta_F\delta \leq 0.08 D_*$. Part 2 bounds $\dist(U,X)$, which equals $\dist(X,U)$ because $\|UQ - X\| = \|U - XQ^\top\|$ for every $Q \in O_k$. The step from $X$ to the next $Y$ is the same with the roles of $U$ and $V$, of $M$ and $M^\top$, and of $W$ and $W^\top$ exchanged. Assumption~\ref{assump:weighted_low_rank}, $\|W\|_{\infty,1}$ and $\| W\circ N\|$ do not change under this exchange, so it gives the same bounds with the roles of $X$ and $Y$ swapped.

\textbf{Induction.} Let $a_0 := \dist(Y_0, V)$, and for $t \geq 1$ let $a_{2t-1} := \dist(X_t, U)$ and $a_{2t} := \dist(Y_t, V)$. The matrix $Y_0$ satisfies the hypotheses of a half step. Applying the half step alternately to $Y_0, X_1, Y_1, X_2, \ldots$ shows, by induction on the number of half steps, that every iterate satisfies the hypotheses of the next half step. This gives the bounds $\dist(\cdot,\cdot) \leq D_*$ and $\rho(\cdot) \leq 25\mu\tau^2$ of the statement, and $a_{j+1} \leq 0.08\, a_j + e$ for every $j \geq 0$, where $e := 8 \tau \Delta_f \delta$. Unrolling the recursion,
\begin{align*}
a_j \leq 0.08^j a_0 + e \sum_{i \geq 0} 0.08^i \leq 0.08^j D_* + 1.1\, e .
\end{align*}
For $j \in \{2t-1, 2t\}$ with $t \geq 1$ we have $0.08^j D_* \leq 0.08^{2t-1}/20 \leq 2^{-t}$, and $1.1\, e \leq 100\tau\Delta_f\delta = \Delta_F \delta$. This proves Eq.~\eqref{eq:dist_x_y}. For $t = 0$, $a_0 \leq D_* \leq 1 = 2^{0}$.
\end{proof}


\section{Tools From Previous Work}\label{sec:app_prev}

In this section, we state several tools from previous work. 
In Section~\ref{sub:app_prev:eigenvalue}, we introduce a varied version of a Lemma from \cite{llr16} to bound the eigenvalues. In Section~\ref{sub:app_prev:eq1}, we bound $\| V^\top Y_{\bot} (Y_{\bot})^\top D_{W^i} Y \|^2$. In Section~\ref{sub:app_prev:summary}, we summarize the base case lemma.

\subsection{Bounding Eigenvalues}
\label{sub:app_prev:eigenvalue}

Now, in this section, we start to bound the eigenvalues.

\begin{lemma}[A variation of Lemma 10 in \cite{llr16}]\label{lem:lemma_10_in_llr16}

Let $Y$ be a (column) orthogonal matrix in $\R^{n \times k}$. Let $\epsilon \in (0,1)$. Let $Q \in O_k$ and suppose $\| VQ - Y \|_F^2 \leq \frac{\epsilon^3 \alpha^2 n}{160 \mu k \| W \|_{\infty,1}}$. Then the bound below holds with $V$ replaced by $VQ$. In particular, when the minimizing $Q$ qualifies, $\| V - Y \|$ may be replaced by $\dist(Y,V)$. We have 
\begin{align*}
| \{ i \in [n] ~|~ \sigma_{\min} (Y^\top D_{W^i} Y) \leq (1-\epsilon) \alpha \} | \leq 10^4 \cdot \frac{\mu^2 k^3 \gamma^2}{\epsilon^4 \alpha^3} \cdot \| W \|_{\infty,1} \cdot \| V  - Y \|^2.
\end{align*}
\end{lemma}
\begin{proof}
The proof uses only the row norms $\| V_j \|_2^2 \leq \mu k/n$ and the bound $V^\top D_{W^i} V \succeq \alpha I$, and both are unchanged when $V$ is replaced by $VQ$. So we write $V$ for $VQ$ throughout the proof.

Let $j$ be an arbitrary integer in $[n]$. Let $g$ be greater than $0$. $j$ is called ``good" if
\begin{align*}
    \|Y_j - V_j\|_2^2 \leq g^2.
\end{align*}

We define $S_g \subset [n]$ as follows
\begin{align*}
S_g := \{ j \in [n] ~|~  \| Y_j - V_j \|_2^2 \leq g^2 \}.
\end{align*}

For convenience, we define $\ov{S}_g \subset [n] $ as follows
\begin{align*}
\ov{S}_g : = [n] \backslash S_g.
\end{align*}
We choose $g$ to satisfy the following condition
\begin{align}\label{eq:choice_g^2}
    g^2 = \frac{\epsilon^2 \alpha}{20 \| W \|_{\infty,1}}.
\end{align}
    Let $a$ be an arbitrary unit vector in $\R^k$. Thus, we have
\begin{align}\label{a^top_Y^top_D_W_i_Y_a}
    a^\top Y^\top D_{W^i} Y a 
    = & ~ \sum_{j \in [n]} (D_{W^i})_j \langle a, Y_j \rangle^2 \notag \\
    \geq & ~ \sum_{j \in S_g} (D_{W^i})_j \langle a, Y_j \rangle^2 \notag\\
    = & ~ \sum_{j \in S_g} (D_{W^i})_j (\langle a, V_j \rangle + \langle a, Y_j - V_j \rangle)^2 \notag\\
    \geq & ~ (1 -  \epsilon / 4 ) \sum_{j \in S_g} (D_{W^i})_j \langle a, V_j \rangle^2 - 4 \epsilon^{-1}   \sum_{j \in S_g} (D_{W^i})_j \langle a, Y_j - V_j \rangle^2 \notag\\
    \geq & ~ (1 - \epsilon / 4 ) \sum_{j \in S_g} (D_{W^i})_j \langle a, V_j \rangle^2 - 4 \epsilon^{-1}   g^2 \sum_{j \in S_g} (D_{W^i})_j \notag \\
    \geq & ~ (1 - \epsilon / 4 ) \sum_{j \in S_g} (D_{W^i})_j \langle a, V_j \rangle^2 - 4 \epsilon^{-1}   g^2 \sum_{j \in [n]} (D_{W^i})_j \notag \\
    \geq & ~ (1 - \epsilon / 4 ) \sum_{j \in [n]} (D_{W^i})_j \langle a, V_j \rangle^2 - \sum_{j\in \ov{S}_g} (D_{W^i})_j \langle a , V_j \rangle^2 - 4 \epsilon^{-1}  g^2 \sum_{j \in [n]} (D_{W^i})_j \notag \\
    \geq & ~ (1 - \epsilon / 4 ) \sum_{j \in [n]} (D_{W^i})_j \langle a, V_j \rangle^2 - \frac{\mu k}{n} \sum_{j\in \ov{S}_g} (D_{W^i})_j - 4 \epsilon^{-1}  g^2 \sum_{j \in [n]} (D_{W^i})_j,
\end{align}
where the first step follows from simple algebra, the second step follows from $S_g \subset [n]$, the third step follows from the property of the inner product, the fourth step follows from Fact~\ref{fac:x+y_square_lower_bound},
the fifth step follows from the definition of $S_g$, the sixth step follows from $(D_{W^i})_j \geq 0$, the seventh step follows from $1-\epsilon/4 \leq 1$, and the last step follows from the property of $V$ (e.g. $\langle a , V_j\rangle^2 \leq \| a \|_2^2 \cdot \| V_j \|_2^2 \leq \| V_j \|_2^2 \leq \xi \leq \mu k/n$).

We can show that
\begin{align}\label{eq:D_W_i_j}
    \sum_{j \in [n]} (D_{W^i})_j \langle a, V_j \rangle^2
    = & ~ a^\top V^\top D_{W^i} V a \notag\\
    \geq & ~ \sigma_{\min} (V^\top D_{W^i} V) \notag\\
    \geq & ~ \alpha,
\end{align}
where the second step follows from Fact~\ref{fac:norm} and the third step follows from definition of $\alpha$ (see Definition~\ref{def:alpha_beta_bounded}) and $\sigma_{\min}(A) \leq \sigma_{\min}(B)$ if $A \preceq B$.

Moreover, recall 
\begin{align}\label{eq:W_infty_1}
    \| W \|_{\infty,1} \geq \max_{i \in [n]} \sum_{j \in [n]} | (D_{W^i})_j |,
\end{align}

We can show that
\begin{align}\label{eq:4_epsilon^-1_g^2}
    4 \epsilon^{-1} g^2 \sum_{j \in [n]} (D_{W^i})_j
    \leq & ~ 4 \epsilon^{-1} g^2 \| W \|_{\infty,1} \notag\\
    = & ~ 4 \epsilon^{-1} \frac{\epsilon^2 \alpha}{20 \| W \|_{\infty,1}} \| W \|_{\infty,1} \notag \\
    \leq & ~  \frac{\epsilon \alpha}{4},
\end{align}
where the first step follows from the definition of $\| W \|_{\infty,1}$ (see Eq.~\eqref{eq:W_infty_1}), the second step follows from the definition of $g^2$ (see Eq.~\eqref{eq:choice_g^2}), and the last step follows from simple algebra.

We define
\begin{align*}
T: = \{ i \in [n] ~|~ \sigma_{\min} (Y^\top D_{W^i} Y) \leq (1-\epsilon) \alpha \} .
\end{align*}

Let us consider 
\begin{align*}
    \sum_{j\in \ov{S}_g} (D_{W^i})_j.
\end{align*}

We define 
\begin{align}\label{eq:S}
    S := \{ i\in [n] \mid \frac{\mu k}{n} \sum_{j\in \ov{S}_g} (D_{W^i})_j \geq \frac{\epsilon \alpha}{4} \}.
\end{align}

If $i \notin S$, then we have
\begin{align*}
a^\top Y^\top D_{W^i} Y a 
\geq & ~ (1 - \epsilon / 4 ) \sum_{j \in [n]} (D_{W^i})_j \langle a, V_j \rangle^2 - \frac{\mu k}{n} \sum_{j\in \ov{S}_g} (D_{W^i})_j - 4 \epsilon^{-1}  g^2 \sum_{j \in [n]} (D_{W^i})_j\\
\geq & ~ (1-\epsilon/4) \alpha - \frac{\mu k}{n} \sum_{j\in \ov{S}_g} (D_{W^i})_j - 4 \epsilon^{-1}  g^2 \sum_{j \in [n]} (D_{W^i})_j\\
\geq & ~ (1-\epsilon/4) \alpha - \epsilon \alpha / 4 - 4 \epsilon^{-1}  g^2 \sum_{j \in [n]} (D_{W^i})_j\\
\geq & ~ (1-\epsilon/4) \alpha -  \epsilon \alpha / 4 - \epsilon \alpha /4 \\
\geq & ~ (1 - \epsilon )\alpha,
\end{align*}
where the first step follows from Eq.~\eqref{a^top_Y^top_D_W_i_Y_a}, the second step follows Eq.~\eqref{eq:D_W_i_j}, the third step follows from the Definition of $S$ (see Eq.~\eqref{eq:S}), the fourth step follows from Eq.~\eqref{eq:4_epsilon^-1_g^2}, and the last step follows from simple algebra.

In summary, we know that if $i \notin S$, then $i \notin T$. By taking its contraposition, we have that if $i \in T$, then $i \in S$.

Thus, we can show that
\begin{align*}
|T| \leq |S|
\end{align*}

In the next a few paragraphs, we will explain how to upper bound $|S|$.

Using Fact~\ref{fac:norm}
\begin{align*}
    \sum_{j \in [n]} \|V_j - Y_j\|_2^2 = \|V - Y\|_F^2,
\end{align*}

By simple counting argument (you have $n$ positive values, their summation is $\| V- Y\|_F^2$, you can't have more than ${\|V - Y\|_F^2}/{g^2}$ of them that are bigger than $g^2$), we have
\begin{align}\label{eq:[n]_S_g}
    |\ov{S}_g |
    \leq & ~ {\|V - Y\|_F^2}/{g^2}.
\end{align}

Let $u_S \in \R^n$ be the indicator vector of $S$, i.e.,
\begin{align*}
\forall i \in [n], ~~~ (u_S)_i = \begin{cases}
1 & \text{~if~} i \in S; \\
0 & \text{~otherwise~} i \notin S.
\end{cases}
\end{align*}

Let $u_g \in \R^n$ be the indicator vector of $\ov{S}_g$, i.e.,
\begin{align*}
\forall i \in [n], ~~~
(u_g)_i = \begin{cases}
1 & \text{~if~} i \in \ov{S}_g; \\
0 & \text{~otherwise~} i \notin \ov{S}_g. 
\end{cases}
\end{align*}

Then, we know that
\begin{align}\label{eq:u_Stop_W_u_g_geq}
    u_S^\top W u_g
    = & ~ \sum_{i \in S} \sum_{j \in \ov{S}_g} (D_{W^i})_j \notag \\
    \geq & ~ |S| \cdot \min_{i \in S} \sum_{j \in \ov{S}_g} (D_{W^i})_j \notag \\
    \geq & ~ |S| \cdot \frac{\epsilon \alpha n}{4 \mu k} ,
\end{align}
where the first step follows from simple algebra, and the second step follows from simple algebra, the third step follows from Definition~\eqref{eq:S}. 

On the other hand,
\begin{align}\label{eq:u_Stop_W_u_g_leq}
    u_S^\top W u_g
    = & ~ u_S^\top {\bf 1}_n {\bf 1}_n^\top u_g + u_S^\top ( W - {\bf 1}_n {\bf 1}_n^\top ) u_g \notag \\
    = & ~ \| u_S \|_1 \cdot \| u_g \|_1 + u_S^\top ( W - {\bf 1}_n {\bf 1}_n^\top ) u_g \notag \\
    \leq & ~ \| u_S \|_1 \cdot \| u_g \|_1 + \| u_S \|_2 \cdot \| W - {\bf 1 }_n {\bf 1}_n^\top \| \cdot \| u_g \|_2 \notag \\
    \leq &~  \| u_S \|_1 \cdot \| u_g \|_1 + \gamma n \cdot \| u_S \|_2  \cdot \| u_g \|_2 \notag \\
    \leq & ~ |S| |\ov{S}_g| + \gamma n \sqrt{|S| |\ov{S}_g|},
\end{align}
where the first step follows from simple algebra, the second step follows from simple algebra, the third step follows from $x^\top A y \leq \| x \|_2 \| A \| \| y\|_2$, the fourth step follows from Definition~\ref{def:gamma_spectral_gap}, and the last step follows from $u_S$ and $u_g$ are indicator vectors.

By combining Eq.~\eqref{eq:u_Stop_W_u_g_geq} and Eq.~\eqref{eq:u_Stop_W_u_g_leq}, we have 
\begin{align}\label{eq:ov_S_g_relation}
    |\ov{S}_g| + \gamma n \cdot \sqrt{ |\ov{S}_g| /|S| } \geq \frac{\epsilon \alpha n}{4 \mu k}.
\end{align}

Note that if $A + B \geq C$. Then if $A \leq C/2$, then $B \geq C/2$.

For the terms in Eq.~\eqref{eq:ov_S_g_relation}, we define $A, B$ and $C$ as follows
\begin{align*}
A: = & ~ |\ov{S}_g| \\
B: = & ~ \gamma n \cdot \sqrt{ |\ov{S}_g| /|S| } \\
C: = & ~ \frac{\epsilon \alpha n}{4 \mu k}
\end{align*}

By Eq.~\eqref{eq:[n]_S_g}, Eq.~\eqref{eq:choice_g^2} and the hypothesis on $\| V - Y \|_F^2$ (recall that $V$ stands for $VQ$), we have
\begin{align*}
    |\ov{S}_g| \leq \| V - Y \|_F^2 \cdot \frac{20 \| W \|_{\infty,1}}{\epsilon^2 \alpha} \leq \frac{\epsilon \alpha n }{8 \mu k},
\end{align*}
so $A \leq C/2$ and therefore $B \geq C/2$, that is,
\begin{align*}
\gamma n \cdot \sqrt{ |\ov{S}_g| /|S| } \geq \frac{\epsilon \alpha n }{8 \mu k}. 
\end{align*}

The above equation implies 
\begin{align}\label{eq:|S|}
    |S|
    \leq & ~ 500 \cdot \frac{ \mu^2 k^2 \gamma^2  }{\epsilon^2 \alpha^2} \cdot | \ov{S}_g | \notag\\
    \leq & ~ 500 \cdot \frac{ \mu^2 k^2 \gamma^2 }{\epsilon^2 \alpha^2  } \cdot ( \| V - Y \|_F^2 / g^2 ), \notag\\
    \leq & ~   500 \cdot \frac{ \mu^2 k^2 \gamma^2 }{\epsilon^2 \alpha^2  } \cdot \| V - Y \|_F^2\cdot \frac{20 \| W \|_{\infty,1}}{\epsilon^2 \alpha}
\end{align}
where the second step follows from Eq.~\eqref{eq:[n]_S_g} and last step follows from Eq.~\eqref{eq:choice_g^2}.
 
We have
\begin{align*}
    | \{ i \in [n] ~|~ \sigma_{\min} (Y^\top D_{W^i} Y) \leq (1-\epsilon) \alpha \} | 
    \leq & ~ |S|\\
    \leq & ~ 10^4 \cdot \frac{\mu^2k^2 \gamma^2}{\epsilon^4 \alpha^3} \cdot \| W \|_{\infty,1} \cdot \| V  - Y \|_F^2 \\
    \leq & ~  10^4 \cdot \frac{\mu^2k^2 \gamma^2}{\epsilon^4 \alpha^3} \cdot \| W \|_{\infty,1} \cdot k \cdot \| V - Y \|^2 
\end{align*}
where the first step follows from the definition of $S$ and the second step follows from combining Eq.~\eqref{eq:|S|} and Eq.~\eqref{eq:choice_g^2}, and the last step follows from Fact~\ref{fac:norm}. This completes the proof. 
\end{proof}

\subsection{Bounding \texorpdfstring{$\| V^\top Y_{\bot} (Y_{\bot})^\top D_{W^i} Y \|^2$}{}}
\label{sub:app_prev:eq1}

In this section, we bound $\| V^\top Y_{\bot} (Y_{\bot})^\top D_{W^i} Y \|^2$ by a multiplicative factor times $\sin^2\theta(Y,V) \leq \dist(Y,V)^2$.

\begin{lemma}[A variation of Lemma 11 in \cite{llr16}]\label{lem:lemma_11_in_llr16}
 
    Let $Y$ be a (column) orthogonal matrix in $\R^{n \times k}$. Let $i \in [n]$. Then we have
    \begin{align*}
        \sum_{i\in [n]} \|V^\top Y_\bot (Y_\bot)^\top D_{W^i} Y \|^2 \leq \gamma^2 \rho (Y)nk^3 \sin^2\theta(Y,V) \leq \gamma^2 \rho (Y)nk^3 \dist(Y,V)^2
    \end{align*}
\end{lemma}

For the completeness, we still provide the proof.
\begin{proof}
    Let $j', j$ be two positive integers in $[k]$. $Y_j$ represents the matrix $Y$'s $j$-th column. $\wt{V}_j$ represents the matrix $Y_\bot Y_\bot^\top V$'s $j$-th column. We define $x^{j,j'} \in \R^n$ as
    \begin{align*}
        x_{i}^{j,j'} = (\wt{V}_j)_i(Y_{j'})_i.
    \end{align*}
    We need to show that the spectral norm of
    \begin{align*}
        V^\top Y_\bot Y_\bot^\top D_{W^i} Y,
    \end{align*}
    is bounded. Note that $\langle \wt{V}_j, Y_{j'} \rangle = 0$, which implies that 
    \begin{align*}
        \sum_{i \in [n]} x_i^{j,j'} = 0.
    \end{align*}

    For $V_j^\top Y_\bot Y_\bot^\top D_{W^i} Y_{j'}$, 
    \begin{align}\label{eq:vjtop_ybot}
        V_j^\top Y_\bot Y_\bot^\top D_{W^i} Y_{j'} 
        = & ~ \sum_{s\in [n]} (D_{W^i})_s (\wt{V}_j)_s (Y_{j'})_s \notag \\
        = & ~ \sum_{s\in [n]} (D_{W^i})_s x_s^{j,j'},
    \end{align}
where the first step follows from the definition of 
\begin{align*}
    V_j^\top Y_\bot Y_\bot^\top D_{W^i} Y_{j'}
\end{align*}
and the second step follows from $(\wt{V}_j)_s (Y_{j'})_s = x_s^{j,j'}$.

It implies that
\begin{align*}
    \sum_{i\in [n]} (\sum_{s\in [n]} (D_{W^i})_s x_s^{j,j'})^2
    = & ~ \|Wx^{j,j'}\|_2^2 \\
    = & ~ \|(W - {\bf 1}_n {\bf 1}_n^\top)x^{j,j'}\|_2^2 \\
    \leq & ~ \|W - {\bf 1}_n {\bf 1}_n^\top\|^2 \|x^{j,j'}\|_2^2 \\
    \leq & ~ \gamma^2n^2\|x^{j,j'}\|_2^2,
\end{align*}
where the first step follows from the definition of $\|Wx^{j,j'}\|_2^2$, the second step follows from ${\bf 1}_n {\bf 1}_n^\top x^{j,j'} = 0$, the third step follows from $\| A x \|_2 \leq \| A \| \| x \|_2$, and the last step follows from $\| W - {\bf 1}_n {\bf 1}_n^\top \| \leq \gamma n$ (see Definition~\ref{def:gamma_spectral_gap}).

Observe that 
\begin{align*}
    \|x^{j,j'}\|_2^2 
    = & ~ \sum_{i\in [n]} (x^{j,j'})^2\\
    = & ~ \sum_{i\in [n]} (\wt{V}_j)_i^2 (Y_{j'})_i^2\\
    \leq & ~ \frac{\rho(Y)k}{n}\sum_{i\in [n]}(\wt{V}_j)_i^2\\
    = & ~ \frac{\rho(Y)k}{n}\|\wt{V}_j\|_2^2\\
    \leq & ~ \frac{\rho(Y)k}{n}\|Y_\bot Y_\bot^\top V\|^2\\
    = & ~ \frac{\rho(Y)k}{n} \sin^2\theta(Y,V)
\end{align*}

where the first step follows from the definition of $\|x^{j,j'}\|_2^2$, the second step follows from $(x^{j,j'})^2 = (\wt{V}_j)_i^2 (Y_{j'})_i^2$, the third step follows from definition of $\rho$, the fourth step follows from the definition of $\|\cdot\|_2^2$, the fifth step follows from the fact that $\wt{V}_j$ is defined to be the $j$-th column of $Y_\bot Y_\bot^\top V$, and the last step follows from Definition~\ref{def:angle_and_distance} ($\sin\theta(Y,V) = \| Y_\bot Y_\bot^\top V \|$). 

It implies that 
\begin{align}\label{eq:dwi}
    \sum_{i\in [n]} (\sum_{s\in [n]} (D_{W^i})_s x_s^{j,j'})^2 \leq \gamma^2 \rho (Y) nk \sin^2\theta(Y,V).
\end{align}

Now we are ready to bound $V^\top Y_\bot Y_\bot^\top D_{W^i} Y$. Note that
\begin{align}\label{eq:Vtop_Y}
     \|V^\top Y_\bot Y_\bot^\top D_{W^i} Y\|^2 
    \leq & ~ \|V^\top Y_\bot Y_\bot^\top D_{W^i} Y\|_F^2 \notag \\
    \leq & ~ \sum_{j,j'\in [k]} (V_j^\top Y_\bot Y_\bot^\top D_{W^i} Y_{j'})^2 \notag \\
    = & ~ \sum_{j,j'\in [k]} (\sum_{s\in [n]} (D_{W^i})_s x_s^{j,j'})^2,
\end{align}
where the first step follows from $\| A\| \leq \| A \|_F$ for all matrix $A$, the second step follows from definition of $\| \cdot \|_F$, and the third step follows from Eq.~\eqref{eq:vjtop_ybot}.

This implies that
\begin{align*}
     \sum_{i\in [n]} \|V^\top Y_\bot Y_\bot^\top D_{W^i} Y\|^2 
    \leq & ~ \sum_{i\in [n]} \sum_{j,j'\in [k]} (\sum_{s\in [n]} (D_{W^i})_s x_s^{j,j'})^2\\
    \leq & ~ \sum_{j,j'\in [k]} \gamma^2 \rho (Y) nk \sin^2\theta(Y,V)\\
    = & ~ \sum_{j \in [k]} \sum_{j'\in [k]} \gamma^2 \rho (Y) nk \sin^2\theta(Y,V)\\
    = & ~ kk \gamma^2 \rho (Y) nk \sin^2\theta(Y,V)\\
    = & ~ \gamma^2 \rho (Y)nk^3 \sin^2\theta(Y,V) \\
    \leq & ~ \gamma^2 \rho (Y)nk^3 \dist(Y,V)^2,
\end{align*}
where the first step follows from Eq.~\eqref{eq:Vtop_Y}, the second step follows from Eq.~\eqref{eq:dwi}, the third step follows from simple algebra, the fourth step follows from the property of $\sum$ (e.g. $\sum_{i = 1}^n a = an$), the fifth step follows from simple algebra, and the last step follows from Lemma~\ref{lem:more_properties_angle_distance}. This completes the proof.
\end{proof}

\subsection{Summary of Base Case Lemma}
\label{sub:app_prev:summary}

We state a base case lemma that covers the two initializations used in this paper, the exact SVD initialization of \cite{llr16} and the fast SVD initialization of Algorithm~\ref{alg:fast_svd_init}.
\begin{lemma}[Base case]\label{lem:base}
Suppose that $M^*$ and $W$ satisfy Assumption~\ref{assump:weighted_low_rank} with $\sigma_{\max}(M^*) = 1$, and that $\Delta_1$ (Definition~\ref{def:Delta_1}, with $\delta = \|W\circ N\|$) satisfies the assumption of Lemma~\ref{lem:svd_init}. Let $D_*$ be as in Lemma~\ref{lem:induction}, and let $Y_0$ be the initial matrix of Algorithm~\ref{alg:main}.
\begin{itemize}
    \item For the exact SVD initialization, $Y_0$ is the matrix $Y_1$ of Lemma~\ref{lem:svd_init}, and $\dist(Y_0,V) \leq 5\sqrt{k}\,\Delta_1 \leq D_*$ and $\rho(Y_0) \leq 16\mu$.
    \item For the fast SVD initialization, suppose moreover that $\|W\circ N\| \geq n^{-c_1}$ for a constant $c_1 \geq 0$ and that the number of rounds of Algorithm~\ref{alg:fast_svd_init} is $q \geq (c_1+6)\ln n/12$. Then, with probability at least $1-2/n^2$, $\dist(Y_0,V) \leq 5\sqrt{k}\,\Delta_1 \leq D_*$ and $\rho(Y_0) \leq 5\mu$.
\end{itemize}
In both cases $Y_0$ satisfies the hypotheses $\dist(Y_0,V)\leq D_*$ and $\rho(Y_0) \leq 25\mu\tau^2$ of Lemma~\ref{lem:induction}.
\end{lemma}
The first part is Lemma~\ref{lem:svd_init}, together with $5\sqrt{k}\,\Delta_1 \leq D_*$, which is the assumption on $\Delta_1$ rewritten with $D_* = \alpha (n/\|W\|_{\infty,1})^{1/2}/(20\sqrt{\mu}\,k)$. The second part is Lemma~\ref{lem:fast_svd_init}. Both are proved below, in Section~\ref{sec:app_init_svd} and Section~\ref{sec:app_init_fast}. The last sentence holds because $16\mu \leq 25\mu\tau^2$.

\section{SVD Initialization and Main Result}\label{sec:app_init_svd}

In Section~\ref{sub:app_init_svd:assumption}, we introduce our assumption on $\delta$. In Section~\ref{sub:app_init_svd:eq1}, we bound $\|(W - \mathbf{1}_n \mathbf{1}_n^\top) \circ H\|$. In Section~\ref{sub:app_init_svd:property}, we analyze the property of the $\rank$-$k$ SVD. In Section~\ref{sub:app_init_svd:init_property}, we analyze the properties of $\dist$ and $\rho$. In Section~\ref{sub:app_init_svd:main_result}, we present our main result. 

\subsection{Assumption}
\label{sub:app_init_svd:assumption}

Here, we set the parameter $\delta$.

\begin{definition}\label{def:delta}
We assume that 
\begin{align*}
    \delta := \|W \circ N\| \leq  0.001 \cdot \alpha \sigma_{\min} (M^*) / k. 
\end{align*}
 
\end{definition}

\subsection{Bounding \texorpdfstring{$\|(W - \mathbf{1}_n \mathbf{1}_n^\top) \circ H\|$}{}}
\label{sub:app_init_svd:eq1}

In this section, we bound $\|(W - \mathbf{1}_n \mathbf{1}_n^\top) \circ H\|$ in terms $\gamma$, the rank $k$ and the top singular value $\sigma_1$.

\begin{lemma}[Spectral lemma, Lemma 5 in \cite{llr16}]\label{lem:lemma_5_in_llr16}
    Let $K$ and $J$ be (column) orthogonal matrices, whose sizes are $n \times n$. Let $H$ be an arbitrary matrix in $\R^{n \times n}$ satisfying
    \begin{align}\label{eq:h}
        H = A \Sigma B^\top,
    \end{align}
    where $A \in \R^{n \times k}$, $B \in \R^{n \times k}$, and $\Sigma \in \R^{k \times k}$. Note that $A$ and $B$ might not be orthogonal, but $\Sigma$ is diagonal. The matrix $W \in \R^{n \times n}$ is an entry wise non-negative matrix, which has an artificial spectral gap, satisfying
    \begin{align*}
        W = \mathbf{1}_n \mathbf{1}_n^\top + \gamma n J \Sigma_W K^\top,
    \end{align*}
    where  
    \begin{align*}
        \|\Sigma_W\| = 1.
    \end{align*}
    Let $\sigma_1 := \max_{r\in [k]} \sigma_r(\Sigma)$. Then, we have
    \begin{align*}
        \|(W - \mathbf{1}_n \mathbf{1}_n^\top) \circ H\| \leq \gamma k \sigma_1 \sqrt{\rho(A) \rho(B)}.
    \end{align*}
\end{lemma}
\begin{proof}

For each $i \in [n]$, let $A_{i,:}$ denote the $i$-th row of matrix $A \in \R^{n \times k}$. For each $r \in [k]$, let $A_r$ denote the $r$-th column of matrix $A \in \R^{n \times k}$. Then we have
\begin{align}\label{eq:bound_A_r_circ_x}
\sum_{r=1}^k \| A_r \circ x \|_2^2 
= & ~
\sum_{r=1}^k \sum_{i=1}^n A_{i,r}^2 x_i^2 \notag \\
= & ~ \sum_{i=1}^n x_i^2 \sum_{r=1}^k A_{i,r}^2 \notag \\
= & ~ \sum_{i=1}^n x_i^2 \| A_{i,:} \|_2^2 \notag \\
\leq & ~ \sum_{i=1}^n x_i^2 \frac{k}{n} \rho(A) \notag \\
\leq & ~ \frac{k}{n} \rho(A)
\end{align}
where the first step follows from the definition of $\|\cdot\|_2^2$, the second step follows from simple algebra, the third step follows from the definition of $\|\cdot\|_2^2$, the fourth step follows from the definition of $\rho$ (Definition~\ref{def:rho}), and the last step follows from $\sum_{i=1}^n x_i^2 \leq 1$.

Let $x,y \in \R^n$ be two arbitrary unit vectors. Then, we have
\begin{align*}
    x^\top ((W - \mathbf{1}_n \mathbf{1}_n^\top) \circ H) y
    = & ~ x^\top ((W - \mathbf{1}_n \mathbf{1}_n^\top) \circ (A \Sigma B^\top)) y\\
    = & ~ \sum_{r = 1}^k \sigma_r x^\top ((W - \mathbf{1}_n \mathbf{1}_n^\top) \circ A_rB_r^\top) y\\
    = & ~ \gamma n \sum_{r = 1}^k \sigma_r (A_r \circ x)^\top J \Sigma_W K^\top (B_r \circ y)\\
    \leq & ~ \gamma n \sum_{r = 1}^k \sigma_r \|A_r \circ x\|_2 \cdot \|J \Sigma_W K^\top\| \cdot \|B_r \circ y\|_2 \\
    \leq & ~ \gamma n \sum_{r = 1}^k \sigma_r \|A_r \circ x\|_2 \cdot \|B_r \circ y\|_2 \\
    \leq & ~ \gamma n \sigma_1 \sum_{r = 1}^k \|A_r \circ x\|_2 \cdot \|B_r \circ y\|_2 \\
    \leq & ~ \gamma n \sigma_1 ( \sum_{r = 1}^k \|A_r \circ x\|_2^2 )^{1/2} ( \sum_{r = 1}^k\|B_r \circ y\|_2^2 )^{1/2} \\
    \leq & ~ \gamma n \sigma_1 \sqrt{ \frac{k}{n} \rho(A) } \sqrt{ \frac{k}{n} \rho(B) } \\
    \leq & ~ \gamma \sigma_1 \cdot k \cdot ( \rho(A) \rho(B) )^{1/2},
\end{align*}
where the first step follows from the definition of $H$ (see Eq.~\eqref{eq:h}), the second step follows from $\Sigma$ is a diagonal matrix, the third step follows from the definition of $W$, the fourth step follows from $\| A x \|_2 \leq \| A \| \cdot \| x \|_2$, the fifth step follows from $\| J \Sigma_W K^\top \| \leq \| J \| \cdot \| \Sigma_W \| \cdot \| W \| \leq 1$, the sixth step follows from $\sigma_1 = \max_{r \in [k]} \sigma_r$, the seventh step follows from Cauchy-Schwarz inequality, the eighth step follows from Eq.~\eqref{eq:bound_A_r_circ_x}, and the last step follows from simple algebra. 

The lemma follows from the definition of the operator norm.
\end{proof}

\begin{lemma}[Wedin’s Theorem, Lemma 6 in \cite{llr16}]\label{lem:wedin_thm}

$M^*$ is a matrix, and $\sigma_1, \ldots, \sigma_n$ are the singular values of $M^*$. ${\wt M}$ is a matrix, and ${\wt \sigma}_1, \ldots, {\wt \sigma}_n$ are the singular values of ${\wt M}$. Suppose that $X, Y$ and $U, V$ are the first $k$ singular vectors (left and right) of ${\wt M}, M^*$ respectively. If there exists $a$ which is greater than $0$ and satisfies 
\begin{align*}
    \max_{r \in \{k + 1, \cdots, n\}} {\wt \sigma}_r \leq \min_{i \in \{1, \cdots, k\}} \sigma_i - a,
\end{align*}
then
\begin{align*}
    \frac{\|M^* - {\wt M}\|}{a} \geq \max\{\sin{\theta(V, Y)}, \sin{\theta(U, X)}\}.
\end{align*}
\end{lemma}

\subsection{Property of Rank-\texorpdfstring{$k$}{} SVD}
\label{sub:app_init_svd:property}

Now, we first define the parameter $\Delta_1$, and then analyze the properties of $\rank$-$k$ SVD.

\begin{definition}\label{def:Delta_1}
We define $\Delta_1$ as follows
\begin{align*}
\Delta_1 : = \frac{10 ( \gamma \mu k + \delta)}{ \sigma_{\min} (M^*) }.
\end{align*}
\end{definition}

\begin{lemma}[Lemma 7 in \cite{llr16}]\label{lem:max_tangent}
    Assume that $W$ and $M^*$ satisfy every assumption. We define $(X, \Sigma, Y) := \rank$-$k$ $\mathrm{SVD} (W \circ M)$. Let $\Delta_1$ be defined as Definition~\ref{def:Delta_1} and assume that $\Delta_1 \leq 0.01$. Then, we have
    \begin{align*}
        \max \{\tan{\theta (X, U)}, \tan{\theta (Y, V)}\} \leq 0.5 \Delta_1.
    \end{align*}
\end{lemma}
\begin{proof}

We know that
\begin{align}\label{eq:W_circ_M_minus_M_*}
    \|W \circ M - M^*\|
    = & ~ \|W \circ (M^* + N) - M^*\|\notag \\
    \leq & ~ \|W \circ M^* - M^*\| + \|W \circ N\| \notag \\
    = & ~ \|W \circ M^* - M^*\| + \delta \notag \\
    \leq & ~ \gamma \mu k \sigma_{\max} (M^*) + \delta \notag \\
    \leq & ~ \gamma \mu k + \delta
\end{align}
where the first step follows from the definition of $M$, the second step follows from triangle inequality and the third step follows from Definition~\ref{def:delta}, the fourth step follows from Lemma~\ref{lem:lemma_5_in_llr16}, the last step follows from $\sigma_{\max}(M^*)=1$.

Therefore, 
\begin{align*}
    \max_{r\in [k+1,n]} \sigma_r (W \circ M) 
    \leq & ~ \max_{r\in [k+1,n]}\sigma_r (W \circ M - M^*) + \max_{r\in [k+1,n]} \sigma_r (M^*) \\  
    \leq & ~ \max_{r\in [k+1,n]}\sigma_r (W \circ M - M^*) + 0 \\
    \leq & ~ \| W \circ M - M^* \| \\
    \leq & ~ \gamma \mu k + \delta \\
    \leq & ~ \frac{1}{4} \sigma_{\min}(M^*) + \delta \\
    \leq & ~ \frac{1}{2} \sigma_{\min} (M^*),
\end{align*}
where the first step follows from triangle inequality, the second step follows from the fact that $M^*$ has rank-$k$, the third step follows from Eq.~\eqref{eq:W_circ_M_minus_M_*}, the fourth step follows from $\gamma \mu k < 0.1 \sigma_{\min}(M^*)$, and the last step follows from $\delta \leq 0.1 \sigma_{\min}(M^*)$.

Now, by Wedin’s theorem (see Lemma~\ref{lem:wedin_thm}) with 
\begin{align*}
    a = \frac{1}{2} \sigma_{\min} (M^*),
\end{align*}
for 
\begin{align*}
    (X, \Sigma, Y) = \text{rank-} k \, \mathrm{SVD}(W \circ M),
\end{align*}
we have 
\begin{align*}
    \max \{\sin{\theta (U, X)}, \sin{\theta (V, Y)}\} \leq \frac{2 (\gamma \mu k + \delta)}{\sigma_{\min} (M^*)}.
\end{align*}
By our choice of parameters
\begin{align*}
    \sin{\theta} \leq 1/2.
\end{align*}
Using Lemma~\ref{lem:more_properties_angle_distance}, we have 
\begin{align*}
    \tan{\theta} \leq 2\sin{\theta}.
\end{align*}
Then the lemma follows.
\end{proof}

\subsection{Initial Properties of Distance and \texorpdfstring{$\rho$}{}}
\label{sub:app_init_svd:init_property}

We analyze the properties of distance and $\rho$ during initialization. We show that as long as $\Delta_1$ is chosen properly, the distance and $\rho$ can be bounded.

\begin{lemma}[SVD initialization, a variation of Lemma 8 in \cite{llr16}]\label{lem:svd_init}
    Assume that $W$ and $M^*$ satisfy every assumption. Let $\Delta_1$ be defined as Definition~\ref{def:Delta_1}. Let $\wt Y_1$ hold the top-$k$ right singular vectors of $W\circ M$, let $\ov Y_1 := \textsc{Clip}(\wt Y_1)$ and let $Y_1 := \textsc{QR}(\ov Y_1)$. (The exact SVD initialization uses this $Y_1$ as the initial matrix $Y_0$ of Algorithm~\ref{alg:main}.) Assume that $\Delta_1 \leq \frac{\alpha}{100\sqrt{\mu}\,k^{1.5}}(n/\|W\|_{\infty,1})^{1/2}$ (this implies $\Delta_1 \leq 0.01/k$, because $\| W \|_{\infty,1} \geq \alpha n/(\mu k)$). Then, we have
    \begin{itemize}
        \item Part 1. $ \dist(V, Y_1) \leq 5\sqrt{k}\,\Delta_1 \leq 1/2$. 
        \item Part 2. Let $\rho(\cdot)$ be defined as Definition~\ref{def:rho}. We have $\rho(Y_1) \leq 16 \mu$. 
    \end{itemize}
\end{lemma}
\begin{proof}

{\bf Proof of Part 1.} First, we consider ${\wt Y}_1 \in \R^{n \times k}$. By Lemma~\ref{lem:max_tangent}, we get that
\begin{align*}
    \dist({\wt Y}_1, V) \leq \Delta_1,
\end{align*}
which means that there exists $Q \in O^{k \times k}$, such that 
\begin{align}\label{eq:spectral_wt_Y_1_Q_minus_V}
    \|{\wt Y}_1 Q - V\| \leq \Delta_1.
\end{align}
Hence, 
\begin{align}\label{eq:F_wt_Y_1_Q_minus_V}
    \|{\wt Y}_1 Q - V\|_F
    \leq & ~ \sqrt{k} \cdot \| {\wt Y}_1 Q - V \| \notag \\
    \leq & ~ \sqrt{k} \cdot \Delta_1 \notag \\
    \leq & ~ \frac{1}{10},
\end{align}
where the first step follows from Fact~\ref{fac:norm}, the second step follows from Eq.~\eqref{eq:spectral_wt_Y_1_Q_minus_V}, and the last step follows from $\Delta_1\leq 0.01 /k$ .

Next, we consider $\ov{Y_1} \in \R^{n \times k}$. In the clipping step, there are two cases for all $i\in [n]$.

{\bf Case 1.} $\| \wt{Y}_{1,i} \|_2^2 > 4\xi$. 
We know
\begin{align}\label{eq:wtY_1i_Q}
    \|\wt{Y}_{1, i} Q\|_2
    = & ~ \| \wt{Y}_{1, i}\|_2  \notag\\
    > & ~ 2\sqrt{\xi},
\end{align}
where the first step follows from Fact~\ref{fac:norm}, and the second step follows from our Case 1 assumption.

We have 
\begin{align}\label{eq:lower_bound_wt_Y_1_i_minus_V_i}
    \|\wt{Y}_{1, i} Q - V_i\|_2 
    \geq & ~ \|  \wt{Y}_{1,i} Q \|_2 - \| V_i \|_2 \notag \\
    \geq & ~ 2\sqrt{\xi} - \| V_i \|_2 \notag \\
    \geq & ~ 2\sqrt{\xi} - \sqrt{\xi} \notag  \\
    = & ~ \sqrt{\xi},
\end{align}
where the first step follows from the triangle inequality, the second step follows from Eq.~\eqref{eq:wtY_1i_Q}, the third step follows from $\| V_i \|_2^2 \leq \mu k/n = \xi$ (incoherence and Definition~\ref{def:xi}), and the last step follows from simple algebra.

By the definition of clipping, in this time, $\ov{Y}_{1,i} = 0$.
\begin{align*}
    \| \ov{Y}_{1, i} Q - V_i\|_2 
    = & ~ \| 0 - V_i \|_2 \\
    = & ~ \|V_i\|_2 \\
    \leq & ~ \sqrt{\xi} \\
    \leq & ~ \| \wt{Y}_{1,i} Q - V_i \|_2
\end{align*}
where the first step follows from $\ov{Y}_{1,i} = 0$ and the second step follows from simple algebra, the third step follows from the property of $V_i$, and the last step follows from Eq.~\eqref{eq:lower_bound_wt_Y_1_i_minus_V_i}. 

{\bf Case 2.} $\| \wt{Y}_{1,i} \|_2^2 \leq 4\xi$. In this case, we know
\begin{align*}
    \ov{Y}_{1,i} = \wt{Y}_{1,i}.
\end{align*}
Thus,
\begin{align*}
\| \ov{Y}_{1,i} Q - V_i \|_2 = \|  \wt{Y}_{1,i} Q - V_i \|_2.
\end{align*}

Combining Case 1 and Case 2, we know that for all $i \in [n]$,
\begin{align*}
\| \ov{Y}_{1,i} - V_i \|_2 \leq \| \wt{Y}_{1,i} - V_i \|_2.
\end{align*}
Taking the summation of squares, we have 
\begin{align}\label{eq:F_ov_Y_1_Q_minus_Y}
\| \ov{Y}_{1} Q - V \|_F^2 \leq & ~ \| \wt{Y}_{1} Q - V \|_F^2 \notag \\
\leq & ~ k \Delta_1^2 \leq \frac{1}{100}
\end{align}
the last step follows from Eq.~\eqref{eq:F_wt_Y_1_Q_minus_V}.

Eventually, we would like to show $V$ is close to $Y_1$. Suppose
\begin{align*}
    Y_1 = \ov{Y_1}R^{-1},
\end{align*}
where $R$ is an upper-triangular matrix.

Then, 
\begin{align*}
    \sin{\theta(V, Y_1)}
    = & ~ \|V_\bot^\top Y_1\|\\
    = & ~ \| V_\bot^\top (\ov{Y}_1 - VQ^{-1})R^{-1} \|\\
    \leq & ~ \|\ov{Y}_1 Q - V\| \cdot \|R^{-1} \|\\
    \leq & ~ \|\ov{Y}_1 Q - V\| \cdot \frac{1}{ \sigma_{\min} ( \ov{Y}_1 ) } \\
    \leq & ~ \|\ov{Y}_1 Q - V\|_F \cdot \frac{1}{\sigma_{\min} (\ov{Y}_1)} ,
\end{align*}
where the first step follows from definition of $\sin$, the second step follows from definition of $Y_1$ and $V_\bot^\top V = 0$, the third step follows from $\| A B \| \leq \| A \| \cdot \| B \|$ and Fact~\ref{fac:norm}, the fourth step follows from singular values of $R$ and those of $\ov{Y}_1$ are identical, and the last step follows from $\| \cdot \| \leq \| \cdot \|_F$.

Note that
\begin{align}\label{eq:lower_bound_sigma_min_ov_Y_1}
    \sigma_{\min} (\ov{Y}_1)
    = & ~ \sigma_{\min} ( \ov{Y}_1 Q ) \notag \\
    \geq & ~ \sigma_{\min} (V) - \| \ov{Y}_1 Q - V \| \notag \\
    \geq & ~ \sigma_{\min} (V) - \|\ov{Y}_1 Q - V\|_F\notag \\
    \geq & ~ \sigma_{\min} (V) - 1/10 \notag \\
    \geq & ~ \frac{1}{2},
\end{align}
where the first step follows from Fact~\ref{fac:norm}, the second step follows from Fact~\ref{fac:norm}, the third step follows from $\| \cdot \| \leq \| \cdot \|_F$, the fourth step follows from Eq.~\eqref{eq:F_ov_Y_1_Q_minus_Y}, and the last step follows from $\sigma_{\min}(V)= 1$.

Thus, 
\begin{align}\label{eq:sin_theta_V_Y_1}
    \sin{\theta(V, Y_1)}
    \leq & ~ 2 \cdot \|\ov{Y}_1 Q - V\|_F \notag \\
    \leq & ~ 2 \sqrt{k}\, \Delta_1 \notag \\
    \leq & ~ 0.02,
\end{align}
where the first step follows from Eq.~\eqref{eq:lower_bound_sigma_min_ov_Y_1}, and the second step follows from Eq.~\eqref{eq:F_ov_Y_1_Q_minus_Y}, the last step follows from $\Delta_1 \leq 0.01/k$. In particular $\cos\theta(V,Y_1) \geq 0.99$ by Lemma~\ref{lem:sin^2_and_cos^2_is_1}.

Therefore, 
\begin{align*}
    \dist(V, Y_1)
    \leq & ~ 2\tan{\theta(V, Y_1)}\\
    = & ~ 2\sin{\theta(V, Y_1)}/\cos{\theta(V, Y_1)}\\
    \leq & ~ 4.1 \sqrt{k}\, \Delta_1 \\
    \leq & ~ 5 \sqrt{k}\, \Delta_1 \leq 1/2,
\end{align*}
where the first step follows from Lemma~\ref{lem:more_properties_angle_distance}, the second step follows from Lemma~\ref{lem:tan_is_sin_cos}, the third step follows from Eq.~\eqref{eq:sin_theta_V_Y_1} and $\cos\theta(V,Y_1) \geq 0.99$, and the last step follows from $\Delta_1 \leq 0.01/k$.

{\bf Proof of Part 2.} For $\rho(Y_1)$, we observe that 
\begin{align*}
    Y_{1, i} = \ov{Y}_i R^{-1},
\end{align*}
We have
\begin{align}\label{eq:upper_bound_Y_1_i}
    \|Y_{1, i}\|_2^2
    \leq & ~ \|\ov{Y}_{1,i}\|_2^2 \cdot \|R^{-1}\|^2 \notag \\
    \leq & ~ 4\xi \cdot \|R^{-1}\|^2 \notag \\
    \leq & ~ 4\xi \cdot \sigma_{\min}(\ov{Y}_1)^{-2} \notag \\
    \leq & ~ 4\xi \cdot 4
\end{align}
where the first step follows from $\| A x \|_2 \leq \| A \| \cdot \| x \|_2$, the second step follows from $\| \ov{Y}_{1,i} \|_2^2 \leq 4\xi$ (by clipping), and the third step follows from $\| R^{-1} \| = \sigma_{\min}(\ov{Y})^{-1}$, and last step follows from $\sigma_{\min}(\ov{Y}_1)^{-1} \leq 2$ (see Eq.~\eqref{eq:lower_bound_sigma_min_ov_Y_1}).

Note that
\begin{align*}
\rho(Y_{1,i}) 
= & ~  \frac{n}{k} \cdot \max_{i\in [n]} \| Y_{1,i} \|_2^2 \\
\leq & ~  \frac{n}{k} \cdot 16 \xi \\
= & ~ 16 \mu
\end{align*}
where the first step follows from Definition~\ref{def:rho}, the second step follows from Eq.~\eqref{eq:upper_bound_Y_1_i}, last step follows from $\xi = \mu k /n$. This leads to the bound, which completes the proof. 
\end{proof}

\subsection{Main Result}
\label{sub:app_init_svd:main_result}

Finally, in this section, we present our main results.

\begin{table}[!ht]
\caption[Summary of results]{Summary of results. Every row states the error bound $O(\alpha^{-1}k\tau)\|W\circ N\|+\epsilon$. $^{\dagger}$The random-initialization guarantee of \cite{llr16} fails on the example of Remark~\ref{rem:random_init_gap}.}
\begin{center}
\begin{tabular}{ |l|l|l| } \hline
{\bf References} & {\bf Init} & {\bf Time} \\  \hline
\citep{llr16} & Random$^{\dagger}$ & $\wt{O}((\|W\|_0k^2+nk^3) \log(1/\epsilon) )  $ \\   \hline
 \citep{llr16} & SVD & $O(n^3) + \wt{O}( (\|W\|_0k^2+nk^3) \log(1/\epsilon) )$\\ \hline
 Theorem~\ref{thm:main_SVD_formal} & SVD &  $O(n^3) + \wt{O}( ( \|W\|_0k + n k^3 )\log(1/\epsilon) )$\\ \hline
Theorem~\ref{thm:main} & Fast SVD (Algorithm~\ref{alg:fast_svd_init}) & $\wt{O}( ( \|W\|_0k + n k^3 ) \log(1/\epsilon) )$ \\  \hline
\end{tabular}
\end{center}
\end{table}

\begin{theorem}[Main result, exact SVD initialization]\label{thm:main_SVD_formal}
Suppose that the hypotheses of Theorem~\ref{thm:main} hold. Algorithm~\ref{alg:main}, started from the exact SVD initialization ($Y_0$ is the matrix $Y_1$ of Lemma~\ref{lem:svd_init}), runs in $O(\log(1/\epsilon))$ iterations and, with probability at least $1-O(1/n^2)$, generates a matrix $\wt{M} \in \R^{n \times n}$ of rank at most $k$ such that
\begin{align*}
\| \wt{M} - M^* \| \leq O(\alpha^{-1} k \tau) \| W \circ N \| + \epsilon.
\end{align*}
The total running time is
\begin{align*}
O(n^3) + \wt{O}( ( \|W\|_0k + n k^3 ) \log(1/\epsilon) ).
\end{align*}
\end{theorem}
\begin{proof}
The error bound and the probability follow from Lemma~\ref{lem:main_correct}, which covers this initialization. The iterations take $\wt{O}( ( \|W\|_0k + n k^3 ) \log(1/\epsilon) )$ time by the proof of Lemma~\ref{lem:main_time}, and the rank-$k$ SVD of $W \circ M$ used for the initialization takes $O(n^3)$ time.
\end{proof}
\section{Fast SVD Initialization}\label{sec:app_init_fast}

In Section~\ref{sub:app_init_fast:alg}, we present the initialization used by Algorithm~\ref{alg:main}, which runs a few rounds of subspace iteration on $W\circ M$ from a Gaussian start. We then prove that this initialization meets the hypotheses of the induction (Lemma~\ref{lem:fast_svd_init}). In Section~\ref{sub:app_init_fast:random}, we record the random initialization of \cite{llr16} and explain why we do not use it (Remark~\ref{rem:random_init_gap}).

\subsection{Subspace Iteration from a Gaussian Start}
\label{sub:app_init_fast:alg}

The exact SVD initialization (Lemma~\ref{lem:svd_init}) uses the top-$k$ right singular vectors of $W\circ M$, which take $O(n^3)$ time to compute. The proof of Lemma~\ref{lem:max_tangent} shows that $W\circ M$ is close to $M^*$ in spectral norm, so the $k$-th singular value of $W\circ M$ is much larger than its $(k+1)$-th. With such a gap, a few rounds of subspace iteration give a matrix that serves as well as the exact singular vectors.

\begin{algorithm}[!ht]
\caption{Fast SVD initialization.}
\label{alg:fast_svd_init}
\begin{algorithmic}[1]
\Procedure{FastSVDInit}{$M\in\R^{n\times n}$, $W\in\R^{n\times n}$, $k$} \Comment{Lemma~\ref{lem:fast_svd_init}}
\State $A\gets W\circ M$ \Comment{Only the $\|W\|_0$ entries on the support of $W$ are stored.}
\State $q\gets\lceil c_q\ln n\rceil$ \Comment{$c_q \geq (c_1+6)/12$, where $\|W\circ N\| \geq n^{-c_1}$ (Lemma~\ref{lem:fast_svd_init}).}
\State Draw $G\in\R^{n\times k}$ with i.i.d.\ $\N(0,1)$ entries
\State $Z\gets\textsc{QR}(G)$
\For{$s=1$ to $q$}
\State $Z\gets\textsc{QR}(A^\top(AZ))$
\EndFor
\State \Return $\textsc{QR}(\textsc{Clip}(Z))$ \Comment{Algorithm~\ref{alg:clip}}
\EndProcedure
\end{algorithmic}
\end{algorithm}

The analysis has three steps. Lemma~\ref{lem:gaussian_start} shows that a Gaussian start is not nearly orthogonal to a fixed $k$-dimensional subspace. Lemma~\ref{lem:subspace_iteration} is the standard convergence bound for subspace iteration. Lemma~\ref{lem:clip_qr} shows that clipping and QR keep a matrix close to $V$ and make it incoherent.

\begin{lemma}[Gaussian start]\label{lem:gaussian_start}
Let $n \geq 2$ and $1 \leq k \leq n$, let $G \in \R^{n\times k}$ have i.i.d.\ $\N(0,1)$ entries, and let $B \in \R^{n\times k}$ be a fixed matrix with orthonormal columns. Then, with probability at least $1-2/n^2$, we have $\|G\| \leq n^2$ and $\sigma_{\min}(B^\top G) \geq n^{-3.5}$.
\end{lemma}
\begin{proof}
Since $\E[\|G\|_F^2] = nk$, Markov's inequality gives $\Pr[\|G\|_F^2 > n^3 k] \leq n^{-2}$. Outside this event, $\|G\| \leq \|G\|_F \leq n^{1.5}k^{1/2} \leq n^2$.

Let $H := B^\top G \in \R^{k\times k}$, with columns $h_1,\ldots,h_k$. The columns $g_1,\ldots,g_k$ of $G$ are independent $\N(0,I_n)$ vectors and $B^\top B = I_k$, so $h_j = B^\top g_j$ are independent $\N(0,I_k)$ vectors. Fix $j\in[k]$. The span of $\{h_i\}_{i\neq j}$ has dimension at most $k-1$, so there is a unit vector $\nu_j \in \R^k$ orthogonal to it that depends only on $\{h_i\}_{i\neq j}$. Conditioned on $\{h_i\}_{i\neq j}$, the variable $\langle \nu_j, h_j\rangle$ is $\N(0,1)$, so $\Pr[|\langle \nu_j, h_j\rangle| \leq t] \leq \sqrt{2/\pi}\, t \leq t$ for every $t > 0$. By a union bound over $j$, with probability at least $1-kt$ we have $|\langle \nu_j, h_j\rangle| > t$ for all $j \in [k]$. On this event, let $x\in\R^k$ be a unit vector and pick $j$ with $|x_j| \geq k^{-1/2}$. Since $\nu_j$ is orthogonal to $h_i$ for every $i\neq j$,
\begin{align*}
\|Hx\|_2 \geq |\langle \nu_j, Hx\rangle| = |x_j| \cdot |\langle \nu_j, h_j\rangle| > t/\sqrt{k}.
\end{align*}
So $\sigma_{\min}(H) > t/\sqrt{k}$. With $t = n^{-3}$ this holds with probability at least $1-kn^{-3} \geq 1-n^{-2}$, and then $\sigma_{\min}(B^\top G) > n^{-3}k^{-1/2} \geq n^{-3.5}$. A union bound over the two events completes the proof.
\end{proof}

\begin{lemma}[Subspace iteration]\label{lem:subspace_iteration}
Let $A\in\R^{n\times n}$ have singular values $s_1 \geq \cdots \geq s_n \geq 0$ with $s_k > s_{k+1}$, and let the columns of $\wt Y \in \R^{n\times k}$ be orthonormal right singular vectors of $A$ for $s_1,\ldots,s_k$. Let $G \in \R^{n\times k}$ be such that $\wt Y^\top G$ is invertible, let $Z_0 := \textsc{QR}(G)$ and, for $s = 1, \ldots, q$, let $Z_s := \textsc{QR}(A^\top (A Z_{s-1}))$. Then every QR step is applied to a matrix with full column rank, $\wt Y^\top Z_q$ is invertible, and
\begin{align*}
\tan\theta(\wt Y, Z_q) \leq \Big(\frac{s_{k+1}}{s_k}\Big)^{2q} \cdot \frac{\|G\|}{\sigma_{\min}(\wt Y^\top G)}.
\end{align*}
\end{lemma}
\begin{proof}
Let the columns of $\wt Y_\bot \in \R^{n\times (n-k)}$ be orthonormal right singular vectors of $A$ for $s_{k+1},\ldots,s_n$. Then
\begin{align*}
A^\top A = \wt Y L_1 \wt Y^\top + \wt Y_\bot L_2 \wt Y_\bot^\top,
\end{align*}
where $L_1 := \diag(s_1^2,\ldots,s_k^2)$ and $L_2 := \diag(s_{k+1}^2,\ldots,s_n^2)$. Since $s_k > s_{k+1} \geq 0$, $L_1$ is invertible, $\|L_1^{-1}\| = s_k^{-2}$ and $\|L_2\| = s_{k+1}^2$. Write $G_1 := \wt Y^\top G$ and $G_2 := \wt Y_\bot^\top G$. For every integer $s \geq 0$,
\begin{align*}
(A^\top A)^s G = \wt Y L_1^s G_1 + \wt Y_\bot L_2^s G_2 .
\end{align*}
We show by induction on $s$ that $Z_s = (A^\top A)^s G\, S_s$ for an invertible $S_s \in \R^{k\times k}$. For $s=0$, $\wt Y^\top G = G_1$ is invertible, so $G$ has full column rank, $G = Z_0 R_0$ with $R_0$ invertible, and $S_0 := R_0^{-1}$. For $s\geq 1$, $A^\top A Z_{s-1} = (A^\top A)^s G\, S_{s-1}$, and $\wt Y^\top (A^\top A)^s G\, S_{s-1} = L_1^s G_1 S_{s-1}$ is invertible. So $A^\top A Z_{s-1}$ has full column rank, its QR decomposition $A^\top A Z_{s-1} = Z_s R_s$ has $R_s$ invertible, and $S_s := S_{s-1} R_s^{-1}$.

For $s = q$ this gives $\wt Y^\top Z_q = L_1^q G_1 S_q$, which is invertible, and $\wt Y_\bot^\top Z_q = L_2^q G_2 S_q$. By Definition~\ref{def:angle_and_distance},
\begin{align*}
\tan\theta(\wt Y, Z_q) = \| \wt Y_\bot^\top Z_q (\wt Y^\top Z_q)^{-1} \| = \| L_2^q G_2 G_1^{-1} L_1^{-q} \| \leq \|L_2\|^q \cdot \|G_2\| \cdot \|G_1^{-1}\| \cdot \|L_1^{-1}\|^q ,
\end{align*}
and $\|G_2\| \leq \|G\|$ and $\|G_1^{-1}\| = 1/\sigma_{\min}(G_1)$.
\end{proof}

\begin{lemma}[Clipping and QR]\label{lem:clip_qr}
Suppose that $M^* = U\Sigma V^\top$ is $\mu$-incoherent (Assumption~\ref{def:mu_incoherent}). Let $Z \in \R^{n\times k}$ have orthonormal columns and satisfy $\dist(Z, V) \leq \Delta$ with $\sqrt{k}\,\Delta \leq 1/10$, and let $\ov Z := \textsc{Clip}(Z)$ (Algorithm~\ref{alg:clip}). Then $\ov Z$ has full column rank, and $Y := \textsc{QR}(\ov Z)$ satisfies
\begin{align*}
\dist(Y, V) \leq 2.5\sqrt{k}\,\Delta \qquad \text{and} \qquad \rho(Y) \leq 5\mu .
\end{align*}
\end{lemma}
\begin{proof}
Let $Q \in O_k$ attain $\|ZQ - V\| = \dist(Z,V) \leq \Delta$. For $i\in[n]$ let $z_i$, $\ov z_i$ and $v_i$ denote the $i$-th rows of $Z$, $\ov Z$ and $V$, viewed as row vectors, and recall that $\|v_i\|_2 \leq \sqrt{\xi}$ with $\xi = \mu k/n$ (Definition~\ref{def:xi}). If $\|z_i\|_2^2 \leq 4\xi$, then $\ov z_i = z_i$. Otherwise $\ov z_i = 0$, and since $\|z_i Q\|_2 = \|z_i\|_2 > 2\sqrt{\xi}$,
\begin{align*}
\|\ov z_i Q - v_i\|_2 = \|v_i\|_2 \leq \sqrt{\xi} < \|z_i Q\|_2 - \|v_i\|_2 \leq \|z_i Q - v_i\|_2 .
\end{align*}
In both cases $\|\ov z_i Q - v_i\|_2 \leq \|z_i Q - v_i\|_2$. Summing the squares over $i$ and using Part 4 of Fact~\ref{fac:norm},
\begin{align*}
\|\ov Z Q - V\| \leq \|\ov Z Q - V\|_F \leq \|Z Q - V\|_F \leq \sqrt{k}\, \|ZQ - V\| \leq \sqrt{k}\,\Delta \leq \frac{1}{10}.
\end{align*}
By Parts 5 and 6 of Fact~\ref{fac:norm}, $\sigma_{\min}(\ov Z) = \sigma_{\min}(\ov Z Q) \geq \sigma_{\min}(V) - \|\ov Z Q - V\| \geq 9/10$. So $\ov Z$ has full column rank, and its QR decomposition $\ov Z = Y R$ has $\|R^{-1}\| = 1/\sigma_{\min}(\ov Z) \leq 10/9$.

Since $(I - VV^\top)V = 0$, we have $(I - VV^\top) Y = (I - VV^\top)\ov Z R^{-1} = (I-VV^\top)(\ov Z Q - V) Q^\top R^{-1}$, so
\begin{align*}
\sin\theta(V, Y) = \|(I-VV^\top)Y\| \leq \|\ov Z Q - V\| \cdot \|R^{-1}\| \leq \frac{10}{9}\sqrt{k}\,\Delta \leq \frac{1}{9}.
\end{align*}
By Lemma~\ref{lem:sin^2_and_cos^2_is_1}, $\cos\theta(V,Y) \geq (1 - 1/81)^{1/2} \geq 0.99$. By Part 5 of Lemma~\ref{lem:more_properties_angle_distance} and Lemma~\ref{lem:tan_is_sin_cos},
\begin{align*}
\dist(Y, V) = \dist(V, Y) \leq 2\tan\theta(V,Y) = \frac{2\sin\theta(V,Y)}{\cos\theta(V,Y)} \leq \frac{2}{0.99}\cdot\frac{10}{9}\sqrt{k}\,\Delta \leq 2.5\sqrt{k}\,\Delta,
\end{align*}
where $\dist(Y,V) = \dist(V,Y)$ because $\|YQ' - V\| = \|Y - VQ'^\top\|$ for every $Q' \in O_k$. Finally, the $i$-th row of $Y = \ov Z R^{-1}$ is $\ov z_i R^{-1}$, and $\|\ov z_i\|_2^2 \leq 4\xi$ by clipping. So $\|\ov z_i R^{-1}\|_2^2 \leq 4\xi \cdot (10/9)^2 \leq 5\xi$ and $\rho(Y) \leq \frac{n}{k}\cdot 5\xi = 5\mu$.
\end{proof}

\begin{lemma}[Fast SVD initialization]\label{lem:fast_svd_init}
Suppose that $M^*$ and $W$ satisfy Assumption~\ref{assump:weighted_low_rank} with $\sigma_{\max}(M^*)=1$, that $\Delta_1$ (Definition~\ref{def:Delta_1}, with $\delta = \|W\circ N\|$) satisfies the assumption of Lemma~\ref{lem:svd_init}, and that $\|W\circ N\| \geq n^{-c_1}$ for a constant $c_1 \geq 0$, where $n \geq 2$. Let $D_*$ be as in Lemma~\ref{lem:induction}. If the number of rounds satisfies $q \geq (c_1+6)\ln n/12$, then, with probability at least $1-2/n^2$, the output $Y_0$ of Algorithm~\ref{alg:fast_svd_init} satisfies
\begin{align*}
\dist(Y_0, V) \leq 5\sqrt{k}\,\Delta_1 \leq D_* \qquad \text{and} \qquad \rho(Y_0) \leq 5\mu .
\end{align*}
Algorithm~\ref{alg:fast_svd_init} runs in $O((\|W\|_0 k + nk^2)\, q)$ time.
\end{lemma}
\begin{proof}
Write $A := W\circ M$, let $s_1\geq\cdots\geq s_n$ be its singular values, and write $\sigma := \sigma_{\min}(M^*)$. As noted in Lemma~\ref{lem:svd_init}, the assumption on $\Delta_1$ implies $\Delta_1 \leq 0.01/k \leq 0.01$. By Eq.~\eqref{eq:W_circ_M_minus_M_*} and Definition~\ref{def:Delta_1}, $\|A - M^*\| \leq \gamma\mu k + \delta = \sigma\Delta_1/10 \leq 0.001\,\sigma$. Since $M^*$ has rank $k$ and $\sigma_k(M^*) = \sigma$, Weyl's inequality for singular values, $|s_j - \sigma_j(M^*)| \leq \|A - M^*\|$ for all $j$, gives $s_{k+1} \leq 0.001\,\sigma$ and $s_k \geq 0.999\,\sigma$. Hence $s_k > s_{k+1}$ and
\begin{align*}
r := s_{k+1}/s_k \leq 0.001/0.999 \leq e^{-6}.
\end{align*}
Let the columns of $\wt Y\in\R^{n\times k}$ be top-$k$ right singular vectors of $A$, as in Lemma~\ref{lem:max_tangent}. This matrix depends only on the input, not on the Gaussian matrix $G$ drawn by the algorithm.

\textbf{Subspace iteration.} By Lemma~\ref{lem:gaussian_start} with $B = \wt Y$, with probability at least $1 - 2/n^2$ we have $\|G\| \leq n^2$ and $\sigma_{\min}(\wt Y^\top G) \geq n^{-3.5}$. Assume this event, and let $Z$ be the matrix after the $q$ rounds of the loop, before clipping. By Lemma~\ref{lem:subspace_iteration},
\begin{align*}
\tan\theta(\wt Y, Z) \leq r^{2q}\, n^{5.5} \leq e^{-12q}\, n^{5.5} \leq n^{-c_1-6}\, n^{5.5} \leq n^{-c_1} \leq \frac{\Delta_1}{2},
\end{align*}
where the third step uses $12 q \geq (c_1+6)\ln n$, and the last step uses $\Delta_1 \geq 10\delta/\sigma \geq 10\delta \geq 10 n^{-c_1}$ (recall that $\sigma \leq \sigma_{\max}(M^*) = 1$).

\textbf{Distance to $V$.} By Part 5 of Lemma~\ref{lem:more_properties_angle_distance}, $\dist(\wt Y, Z) \leq 2\tan\theta(\wt Y, Z) \leq \Delta_1$, and by Lemma~\ref{lem:max_tangent}, $\dist(\wt Y, V) \leq 2\tan\theta(\wt Y,V) \leq \Delta_1$. The distance obeys the triangle inequality: if $Q_1, Q_2 \in O_k$ attain $\|Z Q_1 - \wt Y\| = \dist(Z,\wt Y)$ and $\|\wt Y Q_2 - V\| = \dist(\wt Y, V)$, then
\begin{align*}
\dist(Z, V) \leq \|Z Q_1 Q_2 - V\| \leq \|(Z Q_1 - \wt Y) Q_2\| + \|\wt Y Q_2 - V\| = \dist(Z,\wt Y) + \dist(\wt Y, V),
\end{align*}
and $\dist(Z,\wt Y) = \dist(\wt Y, Z)$ as in the proof of Lemma~\ref{lem:clip_qr}. Hence $\dist(Z, V) \leq 2\Delta_1$.

\textbf{Clipping and QR.} The matrix $Z$ has orthonormal columns, and the output is $Y_0 = \textsc{QR}(\textsc{Clip}(Z))$. Since $2\sqrt{k}\,\Delta_1 \leq 0.02/\sqrt{k} \leq 1/10$, Lemma~\ref{lem:clip_qr} applies with $\Delta = 2\Delta_1$ and gives $\dist(Y_0, V) \leq 5\sqrt{k}\,\Delta_1$ and $\rho(Y_0) \leq 5\mu$. Finally, the assumption of Lemma~\ref{lem:svd_init} gives
\begin{align*}
5\sqrt{k}\,\Delta_1 \leq 5\sqrt{k}\cdot \frac{\alpha}{100\sqrt{\mu}\, k^{1.5}} \Big(\frac{n}{\|W\|_{\infty,1}}\Big)^{1/2} = \frac{\alpha}{20\sqrt{\mu}\,k}\Big(\frac{n}{\|W\|_{\infty,1}}\Big)^{1/2} = D_* .
\end{align*}

\textbf{Running time.} The matrix $A$ has at most $\|W\|_0$ nonzero entries, all on the support of $W$, and forming them takes $O(\|W\|_0)$ time. Drawing $G$ takes $O(nk)$ time. In each round, $AZ$ and $A^\top(AZ)$ take $O(\|W\|_0 k)$ time, and the QR decomposition of an $n\times k$ matrix takes $O(nk^2)$ time. Clipping takes $O(nk)$ time, and the first and the last QR decompositions take $O(nk^2)$ time. The total is $O((\|W\|_0 k + nk^2)\, q)$.
\end{proof}

\subsection{Random Initialization}
\label{sub:app_init_fast:random}

For comparison, we record the random initialization of \cite{llr16} (Algorithm~\ref{alg:random_init}) and the property of it that \cite{llr16} proves (Lemma~\ref{lem:random_init}). Algorithm~\ref{alg:main} does not use this initialization, for the reason given in Remark~\ref{rem:random_init_gap}.

\begin{algorithm}[!ht]
    \caption{Random Initialization}
    \label{alg:random_init}
    \begin{algorithmic}[1]
    \Procedure{RandomInit}{$n,k$}
        \State Let $Y \in \R^{n \times k}$ generated with $Y_{i, j} \gets \frac{1}{\sqrt{n}} b_{i,j}$, where $b_{i,j}$ is drawn uniformly from $\{-1, +1\}$
    \State \Return $Y$ 
    \EndProcedure
    \end{algorithmic}
\end{algorithm}

The following lemma shows that the minimum singular value of the matrix $Y^\top D_{W^i} Y$ can be lower bounded with high probability over all $i\in [n]$.

\begin{lemma}[Random initialization, Lemma 9 in \cite{llr16}]\label{lem:random_init}
    Let $Y \in \R^{n \times k}$ be a random matrix with $Y_{i, j} = \frac{1}{\sqrt{n}} b_{i, j}$, where $b_{i, j}$ are independent and uniform variables from $\{-1, 1\}$. Let $\mu \geq 1$. Let $k \geq 1$. Let $\alpha > 0$. Let $n$ satisfy $\ln n \geq 10$. We assume that  $\| W \|_\infty \leq \frac{\alpha }{k^2 \mu \log^2 {n}} \cdot n$. Then, we have
    \begin{align*}
       \Pr \Big[  \sigma_{\min} ( Y^\top D_{W^i} Y ) \geq \frac{\alpha}{4 \mu k},\ \forall i \in [n] \Big ] \geq 1-1/n^2.
    \end{align*}
\end{lemma}
\begin{proof}

Notice that  
\begin{align*}
    Y^\top D_{W^i} Y
    = & ~ \sum_{j \in [n]} (Y_j)^\top (D_{W^i})_j Y_j.
\end{align*}

For every $j \in [n]$,
\begin{align*}
    (Y_j)^\top (D_{W^i})_j Y_j
\end{align*}
is independent, and
\begin{align*}
    \E[(Y_j)^\top (D_{W^i})_j (Y_j)] = \frac{1}{n} (D_{W^i})_j.
\end{align*}

Using linearity of expectation, we have
\begin{align*}
    \E[\sum_{j \in [n]} (Y_j)^\top (D_{W^i})_j (Y_j)] = \frac{1}{n} \sum_{j \in [n]} (D_{W^i})_j.
\end{align*}

Then, we can get that the following equation holds.
\begin{align}\label{eq:dwi_geq_alphan}
    \sum_{j \in [n]} (D_{W^i})_j \geq \frac{ \alpha n}{k \mu}.
\end{align}

Indeed, by the assumption that the weight is not degenerate, we can get that for every unit vector $a$ in $\R^k$,
\begin{align}\label{eq:V_j_a^2_leq_mukn}
    \alpha \leq a^\top V^\top D_{W^i} V a
    = & ~ \sum_{j \in [n]} (D_{W^i})_j \langle V_j, a \rangle^2 \notag \\
    \leq & ~ \max_{j \in [n]}\{\langle V_j, a \rangle^2\} \sum_{j \in [n]} (D_{W^i})_j \notag \\
    \leq & ~ \frac{\mu k}{n} \sum_{j \in [n]} (D_{W^i})_j,
\end{align}
where the first step follows from Assumption~\ref{def:alpha_beta_bounded}, the third step follows from Fact~\ref{fac:inequality}, and the last step follows from $\langle V_j, a \rangle^2 \leq \| V_j \|_2^2 \leq \mu k/n$ (incoherence of $V$). Rearranging gives Eq.~\eqref{eq:dwi_geq_alphan}. In particular, $\| W \|_{\infty,1} \geq \alpha n/(\mu k)$.

Combining everything together, we get
\begin{align*}
    \E[\sum_{j \in [n]} (Y_j)^\top (D_{W^i})_j Y_j] \geq \frac{\alpha}{k \mu}.
\end{align*}

Define
\begin{align*}
    B := \|(Y_j)^\top (D_{W^i})_j Y_j\| 
    \leq & ~ \frac{k}{n} (D_{W^i})_j \\
    \leq & ~ \frac{\alpha}{k \mu \log^2{n}},
\end{align*}
where the first step follows from our sampling procedure and the second step follows from the assumption that $\|W\|_\infty \leq \frac{\alpha n}{k^2 \mu \log^2{n}}$.

Since all the random variables 
\begin{align*}
    (Y_j)^\top (D_{W^i})_j Y_j
\end{align*}
are independent, applying Matrix Chernoff we get that
\begin{align*}
    \Pr[\sum_{j \in [n]} (Y_j)^\top (D_{W^i})_j Y_j \leq (1 - \delta) \frac{\alpha}{k \mu}] 
    \leq & ~ n(\frac{e^{-\delta}}{(1-\delta)^{(1 - \delta)}})^{\frac{\alpha}{k\mu B}}\\
    \leq & ~ n(\frac{e^{-\delta}}{(1-\delta)^{(1 - \delta)}})^{\log^2{n}}.
\end{align*}
Picking $\delta = \frac{3}{4}$, and union bounding over all $i$, with probability at least $1 - \frac{1}{n^2}$ (here we use $\ln n \geq 10$), for all $i$,
\begin{align*}
    \sigma_{\min}(Y^\top D_{W^i} Y) \geq \frac{\alpha}{4 k \mu}
\end{align*}
as needed.

\end{proof}

\begin{remark}[Why Algorithm~\ref{alg:main} does not use a random start]\label{rem:random_init_gap}
For a random start $Y$, Lemma~\ref{lem:random_init} bounds the smallest eigenvalue of every matrix $Y^\top D_{W^i} Y$ from below. It does not show that the first alternating step brings the iterate close to $V$. Part 2 of Lemma~\ref{lem:robust_update_lemma} uses $\cos\theta(Y,V) \geq 1/2$, whereas a random $Y$ is nearly orthogonal to $V$. For $k=1$, for instance, $\cos\theta(Y,V) = |\langle Y, V\rangle|$ is of order $n^{-1/2}$. The random-start guarantee of \cite{llr16} is its Theorem~3 (numbered as in its arXiv version 1602.02262v2), and the proof of that theorem applies Part~(2) of its Lemma~12 to a random $Y$ in the same way. The following example shows that no argument that uses only the accuracy hypothesis on the solver can close this gap. With exact updates, which meet that hypothesis, the random start fails with probability much larger than $1/n^2$.

Let $n \geq 16$ be even, $k = 1$, $W = {\bf 1}_n{\bf 1}_n^\top$, $M^* = \frac{1}{n}{\bf 1}_n{\bf 1}_n^\top$ and $N = s\, vv^\top$, where $v \perp {\bf 1}_n$ is the unit vector whose entries are $1/\sqrt{n}$ and $-1/\sqrt{n}$ in alternating order, and $0 < s \leq 1$. Then $\gamma = 0$, $\alpha = \beta = 1$, $\mu = \tau = 1$, $\|W\|_\infty = 1 \leq n/\log^2 n$ and $\|W\circ N\| = s$. So, for small $s$ and large $n$, this instance meets the hypotheses that \cite{llr16} and the earlier version of this paper state for their random-start results. Since $W = {\bf 1}_n{\bf 1}_n^\top$, an exact update maps a unit vector $y$ to $My$, and $M{\bf 1}_n = {\bf 1}_n$ while $Mv = sv$. The Rademacher start $y = b/\sqrt{n}$ satisfies ${\bf 1}_n^\top b = 0$ with probability $\binom{n}{n/2}2^{-n} = \Theta(n^{-1/2})$. The events ${\bf 1}_n^\top b = 0$ and $v^\top b = 0$ hold together only if the odd-indexed and the even-indexed entries of $b$ each sum to zero, which has probability $O(1/n)$. So, with probability $\Theta(n^{-1/2})$, we have ${\bf 1}_n^\top b = 0 \neq v^\top b$. In this case the first update $My = (s\, v^\top b/\sqrt{n})\, v$ is a nonzero multiple of $v$. Each such update has entries of squared magnitude at most $s^2/n \leq \mu k/n$, so neither the clipping of Algorithm~\ref{alg:clip} nor that of \cite{llr16} removes anything, and hence every later update is again such a multiple. Then $Y_{T-1}$ is $y$ or $\pm v$, both orthogonal to ${\bf 1}_n$, so the output satisfies $\wt M{\bf 1}_n = \wh X_T Y_{T-1}^\top{\bf 1}_n = 0$, while $M^*{\bf 1}_n = {\bf 1}_n$. Hence $\|\wt M - M^*\| \geq 1$, whereas the claimed bound $O(\alpha^{-1}k\tau)\|W\circ N\| + \epsilon = O(s) + \epsilon$ is less than $1$ for small $s$ and $\epsilon$. Exact updates meet the accuracy hypothesis on $\vec X$ in Lemma~\ref{lem:robust_update_lemma}, so no analysis that uses only that hypothesis on the solver can give a random-start version of Theorem~\ref{thm:main}.

Algorithm~\ref{alg:fast_svd_init} avoids this problem. Like the Rademacher start, a Gaussian start has only a small component in the top-$k$ right singular subspace of $W\circ M$ (of order $n^{-1/2}$ for $k=1$). With probability at least $1-2/n^2$ this component is at least $1/\poly(n)$ (Lemma~\ref{lem:gaussian_start}). The spectral gap of $W\circ M$ lets subspace iteration amplify it by a large constant factor in each round, so after $O(\log n)$ rounds the iterate is as close to $V$ as the exact SVD (Lemma~\ref{lem:fast_svd_init}).
\end{remark}


\section{Bounding the Final Error}\label{sec:app_final}
In Section~\ref{sub:app_final:rewrite}, we express $\wt{M} - M^*$ as a simpler form that is easier for further analysis. In Section~\ref{sub:app_final:bound}, we prove that $\|\wt{M} - M^*\|$ is bounded. Both of these are used to support the proof of our main theorem.

\subsection{Rewrite \texorpdfstring{$\wt{M} - M^*$}{}}
\label{sub:app_final:rewrite}

In this section, we start to simplify/rewrite $\wt{M} - M^*$.

\begin{claim}\label{cla:rewrite_wt_M_minus_M*}

Let $M^* \in \R^{n \times n}$ is $(\mu,\tau)$-incoherent (see Definition~\ref{def:mu_incoherent}).
Let $\wt{M}$ be defined as Theorem~\ref{thm:main}.
    
Then, we have
\begin{align*}
\wt{M} - M^* = U \Sigma Q \Delta_y^\top + U \Sigma V^\top \Delta_y Y^\top +R Y^\top
\end{align*}
\end{claim}
\begin{proof} 
We start expanding the difference by definition:
\begin{align*}
    & ~ \wt{M} - M^* \\
    = & ~ \wh{X} Y^\top - M^* \\
    = & ~ ( U \Sigma V^\top (VQ + \Delta_y) + R) (VQ + \Delta_y )^\top - U \Sigma V^\top \\
    = & ~ U \Sigma V^\top (VQ + \Delta_y) (VQ + \Delta_y)^\top + R (VQ + \Delta_y)^\top \\
    - & ~ U \Sigma V^\top \\
    = & ~ U \Sigma V^\top V Q (VQ + \Delta_y)^\top + U \Sigma V^\top \Delta_y (VQ + \Delta_y)^\top  + R(VQ + \Delta_y)^\top - U \Sigma V^\top \\
    = & ~ U \Sigma V^\top V Q Q^\top V^\top - U \Sigma V^\top + U \Sigma V^\top V Q \Delta_y^\top + U \Sigma V^\top \Delta_y (VQ + \Delta_y)^\top + R(VQ + \Delta_y)^\top \\
    = & ~ U \Sigma V^\top V Q \Delta_y^\top + U \Sigma V^\top \Delta_y (VQ + \Delta_y)^\top + R(VQ + \Delta_y)^\top \\
    = & ~ U \Sigma Q \Delta_y^\top + U \Sigma V^\top \Delta_y Y^\top + R Y^\top, 
\end{align*}
where the first step follows from $\wt{M}=\wh{X}Y^\top$, the second step follows from $\wh{X} = U \Sigma V^\top Y + R$, $Y = V Q + \Delta_y$, and $M^* = U \Sigma V$, the third step follows from simple algebra, the fourth step follows from the simple algebra, the fifth step follows from that for all matrices $A,B$, $(A + B)^\top = A^\top + B^\top$, the sixth step follows from $V \in \R^{n \times k}$ is an orthogonal matrix and $Q \in \R^{k \times k}$ is a rotation matrix, and the last step follows from $Y = V Q + \Delta_y$. 
\end{proof}

\subsection{Bounding \texorpdfstring{$\| \wt{M} - M^* \|$}{}}
\label{sub:app_final:bound}

In this section, we bound $\| \wt{M} - M^* \|$.

\begin{claim}\label{cla:bound_wt_M_minus_M*}
Let $Y := Y_{T-1}$ and $\wh X := \wh X_T$, so that $\wt M = \wh X Y^\top$, and let $Q$, $\Delta_y$ and $R$ be as in the proof of Lemma~\ref{lem:main_correct}: $Y = VQ + \Delta_y$ with $Q \in O_k$ and $\|\Delta_y\| = \dist(Y,V)$, and $R = \wh X - U\Sigma V^\top Y$. Suppose that
\begin{align*}
\|R\| \leq \Delta_d \dist(Y,V) + \Delta_f \|W\circ N\| \qquad \text{and} \qquad \dist(Y,V) \leq 2^{-(T-1)} + \Delta_F \|W\circ N\|,
\end{align*}
where $\Delta_d \leq 1$ and $\Delta_f \leq \Delta_F$, and that $T \geq \log_2(6/\epsilon) + 1$. Then
\begin{align*}
\| \wt{M} - M^* \| \leq 4\Delta_F \| W \circ N \| + \epsilon .
\end{align*}
\end{claim}
\begin{proof}
By Claim~\ref{cla:rewrite_wt_M_minus_M*},
\begin{align*}
    \|\wt{M} - M^*\|
    = & ~ \| U \Sigma Q \Delta_y^\top + U \Sigma V^\top \Delta_y Y^\top + R Y^\top \| \\
    \leq & ~ \|U \Sigma Q\| \|\Delta_y\| + \|U \Sigma V^\top\| \|\Delta_y\| \|Y\| + \|R\| \|Y\| \\
    = & ~ 2 \dist(Y, V) + \|R\| \\
    \leq & ~ (2 + \Delta_d) \dist(Y,V) + \Delta_f \| W \circ N \| \\
    \leq & ~ 3 \cdot 2^{-(T-1)} + (3\Delta_F + \Delta_f) \| W \circ N \| \\
    \leq & ~ 4 \Delta_F \| W \circ N \| + \epsilon ,
\end{align*}
where the second step uses the triangle inequality and $\|AB\| \leq \|A\| \|B\|$, the third step uses $\|U\Sigma Q\| = \|U\Sigma V^\top\| = \|\Sigma\| = 1$ and $\|Y\| = 1$ (the matrices $U$, $V$ and $Y$ have orthonormal columns and $Q$ is orthogonal), the fourth step uses the bound on $\|R\|$, the fifth step uses $\Delta_d \leq 1$ and the bound on $\dist(Y,V)$, and the last step uses $\Delta_f \leq \Delta_F$ and $3\cdot 2^{-(T-1)} \leq \epsilon/2$.
\end{proof}

\section{Proof of Main Result}
\label{sec:proof_thm_main}

We dedicate this section to the proof of Theorem~\ref{thm:main}. There are two parts of Theorem~\ref{thm:main}: the correctness part and the running time part. In Section~\ref{sub:app_final:proof}, we present the proof of the correctness part of Theorem~\ref{thm:main}. In Section~\ref{sub:our_results:running_time_proof}, we display the proof of the running time part of Theorem~\ref{thm:main}.

\subsection{Correctness Part of Theorem~\ref{thm:main}}
\label{sub:app_final:proof}

Now, we start proving the correctness part of Theorem~\ref{thm:main}.

\begin{lemma}[Correctness part of Theorem~\ref{thm:main}]\label{lem:main_correct}
Assume the hypotheses of Theorem~\ref{thm:main}, with $\|W\circ N\| \geq n^{-c_1}$ for a constant $c_1 \geq 0$. Run Algorithm~\ref{alg:main} with $T \geq \log_2(6/\epsilon) + 1$ iterations, with $\delta_{\mathrm{sk}} = 1/\poly(n,T)$ small enough and $\epsilon_{\mathrm{sk}} \leq 10^{-4}\Delta_f^2\|W\circ N\|^2$, and with the initial matrix $Y_0$ computed either by Algorithm~\ref{alg:fast_svd_init} with $q \geq (c_1+6)\ln n/12$ rounds or by the exact SVD initialization. Then, with probability at least $1-O(1/n^2)$, the output $\wt M = \wh X_T Y_{T-1}^\top$ has rank at most $k$ and satisfies
\begin{align*}
\|\wt{M} - M^*\| \leq 4\Delta_F \|W \circ N\| + \eps = O( \alpha^{-1}  k \tau  ) \cdot \|W \circ N\| + \eps,
\end{align*}
where $\Delta_F = 100\,\tau\,\Delta_f = 100\, C\alpha^{-1}k\tau$ (with $\eta = 1$).
\end{lemma}

\begin{proof}
Write $\sigma := \sigma_{\min}(M^*) = 1/\tau$ and $\delta := \|W\circ N\|$. The choice $\epsilon_{\mathrm{sk}} = 10^{-4} n^{-2c_1}$ is allowed, since $\Delta_f = C\alpha^{-1}k \geq 1$, and it is $1/\poly(n)$, as in Algorithm~\ref{alg:main}.

\textbf{Hypotheses of Lemma~\ref{lem:induction}.} First, $\Delta_d \leq \sigma/100$. We take the polynomial $p := \poly(k,\tau,\mu)$ in Definition~\ref{def:gamma_upper_bound} to satisfy $p \geq 3C\mu^{1.5}k^2\tau^2$, so that $\gamma \leq \alpha^{1.5}n^{-c_0}/(100p)$. By Part 2 of Lemma~\ref{lem:W_infty_norm}, $(\|W\|_{\infty,1}/n)^{1/2} \leq (n^{1/2}\gamma + 1)^{1/2} \leq n^{1/4}\gamma^{1/2} + 1$, and $n^{1/4}\gamma^{1.5} \leq \alpha^{2.25}/(1000p^{1.5})$ because $c_0 \geq 1/6$. So the first term of $\Delta_d$ (Definition~\ref{def:Delta_d_Delta_f}) is at most
\begin{align*}
C\alpha^{-1.5}\mu^{1.5}k^2 (n^{1/4}\gamma^{1.5} + \gamma) \leq C\mu^{1.5}k^2 \Big(\frac{1}{1000p} + \frac{1}{100p}\Big) \leq \frac{\sigma}{200},
\end{align*}
and the second term, with $\eta = 1$, is $C\alpha^{-1}\mu k^2\tau\gamma \leq C\mu k^2\tau/(100p) \leq \sigma/200$. Second, $\Delta_F\delta \leq D_*$, because the hypothesis on $\|W\circ N\|$ gives
\begin{align*}
\Delta_F \delta = 100\,C\alpha^{-1}k\tau\,\delta \leq 100\,C c\,\frac{\alpha}{\sqrt{\mu}\,k}\Big(\frac{n}{\|W\|_{\infty,1}}\Big)^{1/2} = 2000\,C c\, D_* \leq D_*
\end{align*}
for $c \leq 1/(2000C)$. Third, for $c \leq 1/100$ the hypothesis on $\Delta_1$ implies the assumption of Lemma~\ref{lem:svd_init}. So, by Lemma~\ref{lem:base}, the initial matrix satisfies $\dist(Y_0, V) \leq D_*$ and $\rho(Y_0) \leq 25\mu\tau^2$, with probability at least $1-2/n^2$ for the fast SVD initialization.

\textbf{Accuracy of the regressions.} Consider a step that computes $\vec X_t$ from $Y = Y_{t-1}$, and a row $i$ with $\sigma_{\min}(Y^\top D_{W^i}Y) > 0.25\alpha$. The $i$-th regression has the design matrix $A_i := (D_{\sqrt{W^i}} Y)_{S_i,:}$, where $S_i$ is the support of $W_{i,:}$, and $\sigma_{\min}(A_i)^2 = \sigma_{\min}(Y^\top D_{W^i} Y) > 0.25\alpha$. With probability at least $1-\delta_{\mathrm{sk}}$, Lemma~\ref{lem:weighted_lr_meta} and Lemma~\ref{lem:alg_takes_time} return a $(1+\epsilon_0)$-approximate solution with $\epsilon_0 = \epsilon_{\mathrm{sk}}/\poly(n)$, and then Lemma~\ref{lem:distance_app_opt} gives
\begin{align*}
\| (\vec X_t)_{i,:} - (\wt X_t)_{i,:} \|_2^2 \leq O(\epsilon_0)\cdot \frac{4}{\alpha} \cdot \sum_{j \in [n]} W_{i,j} M_{i,j}^2 \leq \frac{\epsilon_{\mathrm{sk}}}{n}.
\end{align*}
Here the optimal cost is at most $(\sum_{j} W_{i,j}M_{i,j}^2)^{1/2}$, and $\sum_{j} W_{i,j}M_{i,j}^2 \leq n\|W\|_\infty (1 + \|N\|_\infty)^2 \leq \poly(n)$ by Lemma~\ref{lem:W_infty_norm} and $|M^*_{i,j}| \leq \|U_{i,:}\|_2\|V_{j,:}\|_2 \leq 1$, while $\alpha \geq 1/\poly(n)$. The same holds for the steps that compute $\vec Y_t$. The sketches of each step are drawn independently of the earlier steps, so a union bound over the $2nT$ regressions shows that the accuracy hypothesis of Lemma~\ref{lem:induction} holds with probability at least $1-2nT\delta_{\mathrm{sk}} \geq 1 - 1/n^2$.

\textbf{Final error.} On the intersection of these events, which has probability at least $1-3/n^2$, Lemma~\ref{lem:induction} applies. Let $Y := Y_{T-1}$ and $\wh X := \wh X_T$, the clipped approximate solution returned by Algorithm~\ref{alg:main}. Then $\dist(Y,V) \leq 2^{-(T-1)} + \Delta_F\delta$, $\dist(Y,V) \leq D_*$ and $\rho(Y) \leq 25\mu\tau^2$. So Part 1 of Lemma~\ref{lem:robust_update_lemma} applies to $Y$ and gives, with $R := \wh X - U\Sigma V^\top Y$,
\begin{align*}
\|R\| \leq \|R\|_F \leq \Delta_u = \Delta_d \dist(Y,V) + \Delta_f\delta .
\end{align*}
Let $P := \arg\min_{Q'\in O_k}\|YQ' - V\|$, $Q := P^\top$ and $\Delta_y := Y - VQ$. Then $Y = VQ + \Delta_y$ and $\|\Delta_y\| = \|(YP - V)P^\top\| = \dist(Y,V)$. By Claim~\ref{cla:rewrite_wt_M_minus_M*},
\begin{align*}
\wt{M} - M^* = U \Sigma Q \Delta_y^\top + U \Sigma V^\top \Delta_y Y^\top +R Y^\top,
\end{align*}
and Claim~\ref{cla:bound_wt_M_minus_M*}, with $\Delta_d \leq \sigma/100 \leq 1$ and $\Delta_f \leq 100\tau\Delta_f = \Delta_F$, gives $\|\wt M - M^*\| \leq 4\Delta_F\delta + \epsilon$. Finally, $\wt M = \wh X Y^\top$ with $\wh X \in \R^{n\times k}$ has rank at most $k$.
\end{proof}

\subsection{Running Time Part of Theorem~\ref{thm:main}}
\label{sub:our_results:running_time_proof}

Now, we start proving the running time part of Theorem~\ref{thm:main}.

\begin{lemma}[Running Time Part of Theorem~\ref{thm:main}]\label{lem:main_time}
The running time of Algorithm~\ref{alg:main} is $\wt{O}( ( \|W\|_0 k + n k^3) \log(1/\epsilon) )$ with the fast SVD initialization of Algorithm~\ref{alg:fast_svd_init} and $q = O(\log n)$ rounds.
\end{lemma}

\begin{proof}
Now we analyze the running time. 
  We first compute the initialization time.  By Lemma~\ref{lem:fast_svd_init} with $q = O(\log n)$, Algorithm~\ref{alg:fast_svd_init} takes $O((\|W\|_0 k + nk^2)\log n)$ time, which is within the claimed bound.  There are $T$ iterations.  For each iteration, there are three major steps, solving regression, Clip and QR.
      The dominating step is to solve regression. We choose $\epsilon_{\mathrm{sk}}$ as Claim~\ref{cla:epsilon_sk_condition}. 
      Using Lemma~\ref{lem:weighted_lr_meta} and Lemma~\ref{lem:distance_app_opt}, we know that we should choose $\epsilon_0 =\epsilon_{\mathrm{sk}}/\poly(n)$ 
      and $\delta_0 = 1/\poly(n, \log(1/\epsilon))$, this step takes $\wt{O}(\|W\|_0 k + n k^3)$ time, because each regression is first restricted to the $\|W_i\|_0$ rows of $D_{\sqrt{W_i}} B$ with nonzero weight (Algorithm~\ref{alg:multiple_regression}) and $\alpha \geq 1/\poly(n)$.  The \textsc{Clip} and \textsc{QR} algorithms take time $O(n k)$ and $O(n k^2)$ respectively.   Hence, the $T$ iterations take time $\wt{O} ( (\|W\|_0 k + n k^3 ) \log(1/\epsilon))$. 
\end{proof}

\section{Experimental Results}

We conducted two experiments showing the performance of our main algorithm (Algorithm~\ref{alg:main}) and one experiment, particularly for our novel high-precision regression algorithm (Algorithm~\ref{alg:iter_regression}). We first present the first two experiments for our main algorithm. In both experiments, we set 
\begin{align*}
M = M^* + N \in \mathbb{R}^{n \times n}
\end{align*}
where $ M^* $ is the rank-$ k $ ground truth and $ N $ is a higher-rank noise matrix. We set $ n = 800 $ and $ k = 100 $. We generate the noise matrix $ N $ as an $ n \times n $ random matrix with i.i.d. Gaussian entries of zero mean and variance $ \frac{1}{k} $. We apply the sketching matrix $ S \in \mathbb{R}^{m \times n} $ with $ m = 150 $ (we choose the CountSketch matrix \citep{ccf02} when solving for the regression problems (see Lines~\ref{line:fast_mul_reg_y} and~\ref{line:fast_mul_reg_x} from our Algorithm~\ref{alg:main}). We iterate the alternating minimization steps (see Line~\ref{line:for_loop}) for $ T = 20 $ times. To show the performance of our algorithm, we compare the running time and error between our Algorithm~\ref{alg:main} and the exact solver from \cite{llr16}. These experiments are those of the earlier version of this paper. They were run before the initialization of Algorithm~\ref{alg:main} was changed to Algorithm~\ref{alg:fast_svd_init} and were not rerun, so they do not test the new initialization.

\paragraph{Experiment 1}
 
The first experiment is the matrix completion problem. For each row of the weight matrix $ W \in \mathbb{R}^{n \times n}_{\geq 0} $, we randomly select 400 entries to be equal to 1 and the remaining 400 entries to be 0. The second experiment is the general weighted low rank approximation where the weight matrix $ W $ is constructed via $ {\bf 1}_n {\bf 1}_n^\top + G $ for $ G $ being a random matrix with standard Gaussian entries.

Below, we present our experimental results of the matrix completion problem. We generate the ground truth $ M^* = XY^\top $ by a pair $ X, Y \in \mathbb{R}^{n \times k} $ with random i.i.d. entries scaled by $ \frac{1}{\sqrt{k}} $, and the distributions are Laplace, Gaussian, and uniform, respectively. Our time is measured in seconds.

\begin{table}[!ht]
    \centering
    \caption{Experimental results of Algorithm~\ref{alg:main} for the matrix completion problem.}
    \begin{tabular}{lp{2cm}p{2cm}lp{2cm}p{2cm}}
        \toprule
        Distribution & Time of the exact solver & Time of the approximate solver & \% speedup & Exact solver error & Approximate solver error \\
        \midrule
        Laplace  & 234  & \textbf{205} & 12.39\% & $1.54 \times 10^{-4}$ & $3.56 \times 10^{-3}$ \\
        Gaussian & 247  & \textbf{216} & 12.55\% & $3.85 \times 10^{-4}$ & $1.33 \times 10^{-3}$ \\
        Uniform  & 248  & \textbf{207} & 16.53\% & $1.34 \times 10^{-5}$ & $4.99 \times 10^{-4}$ \\
        \bottomrule
    \end{tabular}
\end{table}

\paragraph{Experiment 2}

We present our second experimental results as follows (recall the weight matrix is an all-$1$'s matrix plus a noise matrix with i.i.d. standard Gaussian entries):

\begin{table}[!ht]
    \centering
    \caption{Experimental results of Algorithm~\ref{alg:main} for the weighted low-rank approximation.}
    \begin{tabular}{lp{2cm}p{2cm}lp{2cm}p{2cm}}
        \toprule
        Distribution & Time of the exact solver & Time of the approximate solver & \% speedup & Exact solver error & Approximate solver error \\
        \midrule
        Laplace  & 249  & \textbf{222} & 10.84\% & $2.11 \times 10^{-4}$ & $4.71 \times 10^{-4}$ \\
        Gaussian & 223  & \textbf{204} & 8.52\%  & $6.29 \times 10^{-5}$ & $2.15 \times 10^{-4}$ \\
        Uniform  & 221  & \textbf{215} & 2.71\%  & $3.00 \times 10^{-5}$ & $7.13 \times 10^{-5}$ \\
        \bottomrule
    \end{tabular}
\end{table}

We note that in these two settings, our algorithm achieves a speedup compared to the algorithm of \cite{llr16}. For the matrix completion setting, our speedup is in the range of 12\%-16\%, while for the dense weights regime, our speedup ranges from 2.7\% to 10.8\%. In 20 iterations, our approximate solver obtains errors within a factor of about 40 of those of the exact solver. Since a standard Gaussian entry is below $-1$ with probability about $0.16$, the weight matrix of Experiment 2 has negative entries and hence lies outside the setting $W \in \R^{n\times n}_{\geq 0}$ of Problem~\ref{prob:weighted_low_rank}. The experiments also use a CountSketch-based solver, which differs from the {\sf SRHT}-based \textsc{HighPrecisionReg} (Algorithm~\ref{alg:iter_regression}) that we analyze.

\paragraph{Experiment 3}

We expect the speedup will be more significant once the discrepancy between $ n $ and $ k $ is larger, as sketching is known to work well in the regime where $ n \gg k $, as evidenced by the following experiment on using sketching to solve the regression:

\begin{align*}
\min_{x \in \mathbb{R}^{k}} \|Ax - b\|_2^2 \quad \text{for } A \in \mathbb{R}^{n \times k} \text{ and } b \in \mathbb{R}^{n}.
\end{align*}

We test the performance of regression solvers for $ n = 10^6 $, $ k = 500 $ with a sketch size $ m = 5500 $, and run our solver for 5 iterations. The results are as follows:

\begin{table}[!ht]
    \centering
    \caption{Experimental results of Algorithm~\ref{alg:iter_regression}.}
    \begin{tabular}{lcccc}
        \toprule
        Distribution & Time for exact solve & Time for approx solve & Percentage speedup & Error \\
        \midrule
        Gaussian  & 5.149 & \textbf{3.695} & 28.24\% & $8.24 \times 10^{-3}$ \\
        Laplace   & 5.480 & \textbf{3.866} & 29.45\% & $8.41 \times 10^{-3}$ \\
        Power Law & 5.505 & \textbf{4.444} & 19.27\% & $8.77 \times 10^{-3}$ \\
        \bottomrule
    \end{tabular}
\end{table}

Our data matrices $ A $ and response vectors $ b $ are generated according to standard Gaussian, Laplace, and power law distribution with $ p = 5 $, and we measure the $ \ell_{\infty} $ error of the solution, i.e., let $ \hat{x} $ denote the vector outputted by the solver, we measure 
\begin{align*}
\|\hat{x} - x^*\|_{\infty}
\end{align*}
where 
\begin{align*}
x^* = \arg \min_{x \in \mathbb{R}^k} \|Ax - b\|_2^2.
\end{align*}
The speedup obtained here ranges from 20\% to 30\%, so we have strong grounds to believe that the acceleration will be even more evident when $ n $ is large. However, performing weighted low-rank approximation on $ 10^6 \times 10^6 $ size matrices is currently out of the scope of our computational power, so we leave this as a future direction.

\ifdefined\isarxiv
\bibliographystyle{alpha}
\bibliography{ref}
\else
\fi



\end{document}